\documentclass[10pt,twocolumn,letterpaper]{article}

\usepackage{wacv}
\usepackage{times}
\usepackage{epsfig}
\usepackage{graphicx}
\usepackage{amsmath}
\usepackage{amssymb}

\usepackage{comment}
\usepackage[table,dvipsnames]{xcolor}
\usepackage{array,multirow}
\usepackage{soul}
\usepackage{ulem}
\usepackage{subcaption}
\usepackage[vlined,ruled]{algorithm2e}
\SetKwInOut{Input}{input}
\SetKwInOut{Output}{output}
\SetKwComment{Comment}{}{}

\SetCommentSty{mycmtsty}

\definecolor{Cyan}{rgb}{0,.68,.94} 
\usepackage[abs]{overpic}  
\newcommand{\overimg}[3][]{%
    \begin{overpic}[#1]{#2}%
      \put (0, 2) {%
        \setlength{\fboxsep}{2pt}%
        \colorbox{Cyan!0!white}{%
          \scriptsize\sffamily\vphantom{y}%
          #3%
        }%
      }%
    \end{overpic}%
}
		
\usepackage{pgfpages}
\usepackage{pgf}
\usepackage{tikz}

\usetikzlibrary{positioning,fit,calc,shapes.geometric}
\tikzset{block/.style={thick,minimum height=.6cm,align=center},
         line/.style={-latex,thick}
}

\newcommand{\B}[1]{\bf{#1}}

\newcommand{\pa}[2]{{}{#2}}
\newcommand{\vd}[2]{{}{#2}}

\newcommand{\nada}[1]{}
\newcommand{\gfsqueeze}{\vspace{-.5em}}

\DeclareMathOperator*{\argmin}{arg\,min}

\newcommand{\myparagraph}[1]{\medskip\noindent\textbf{#1}}

\def\wacvPaperID{????} 

\wacvfinalcopy 
\ifwacvfinal
\def\assignedStartPage{9876} 
\fi

\ifwacvfinal
\usepackage[breaklinks=true,bookmarks=false]{hyperref}
\else
\usepackage[pagebackref=true,breaklinks=true,colorlinks,bookmarks=false]{hyperref}
\fi

\ifwacvfinal
\else
\fi

\begin{document}
\normalem

\title{Self-supervised training for blind multi-frame video denoising}

\author{
Valéry Dewil\hspace{1em} \hfill Jérémy Anger\hspace{1em} \hfill Axel Davy\hspace{1em} \hfill  Thibaud Ehret\hspace{1em} \hfill Gabriele Facciolo\hspace{1em} \hfill Pablo Arias\\
Université Paris-Saclay, CNRS, ENS Paris-Saclay, Centre Borelli, 
91190, Gif-sur-Yvette, France\\
{\tt\normalsize \color{purple} \url{https://cmla.github.io/mf2f/}}\\
}

\maketitle

\begin{abstract}
We propose a self-supervised approach for training multi-frame video denoising networks. These networks predict  each frame from a  stack of frames around it. Our self-supervised approach benefits from the  temporal consistency in the video by minimizing a loss that penalizes the difference between the predicted frame  and a neighboring one, after aligning them using an optical flow. 
We use the proposed strategy 
{to denoise a video contaminated with an unknown noise type, by fine-tuning a pre-trained denoising network on the noisy video}. The proposed fine-tuning reaches and sometimes surpasses the performance of state-of-the-art networks trained with supervision. We demonstrate this by showing extensive results on video blind denoising of different synthetic and real noises. 
{In addition, the proposed fine-tuning can be applied to any parameter that controls the denoising performance of the network. We show how this can be expoited to perform joint denoising and noise level estimation for heteroscedastic noise.}

\end{abstract}

\section{Introduction}


Denoising has been a fundamental problem of image and video processing since the early days of these disciplines. It continues to be an active research area due to the continuous need for reducing the size of imaging sensors and the desire of imaging in increasingly challenging conditions (such as low light and short exposure times).

\begin{figure*}
	\begin{center}
		\def\imagesize{0.235\textwidth}
		\setlength{\tabcolsep}{2pt}
		\begin{tabular}{lcccc}
		\rotatebox{90}{\small\hspace{4.5mm}Poisson noise}~
		&
		\includegraphics[width=\imagesize]{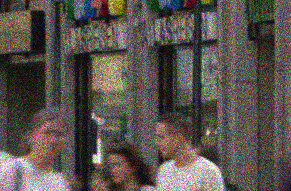}\hspace{0.8mm}%
		&
		\overimg[width=\imagesize]{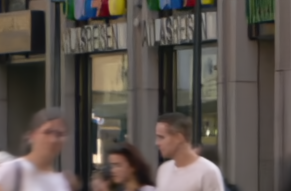}{37.37dB}\hspace{0.8mm}
		&
		\overimg[width=\imagesize]{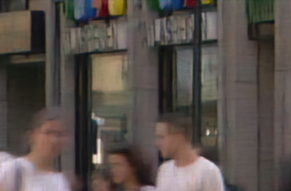}{30.87dB}\hspace{0.8mm}
        &
		\overimg[width=\imagesize]{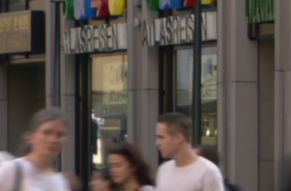}{36.66dB}
        \\[0.3mm]
		\rotatebox{90}{\small\hspace{6mm}Box noise}~%
		&
		\includegraphics[width=\imagesize]{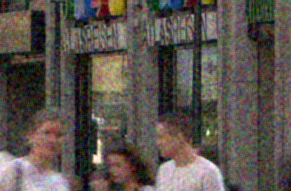}\hspace{0.8mm}%
		&
		\overimg[width=\imagesize]{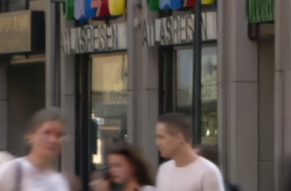}{36.71dB}\hspace{0.8mm}
		&
		\overimg[width=\imagesize]{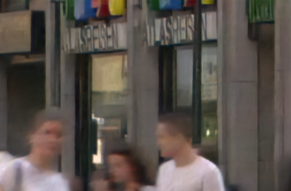}{32.76dB}\hspace{0.8mm}
		&
		\overimg[width=\imagesize]{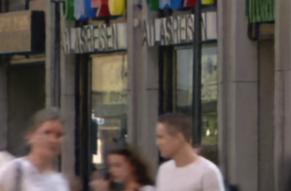}{36.36dB}
        \\[0.3mm]
		\rotatebox{90}{\small\hspace{1mm}Demosaicked noise}~%
		&
		\includegraphics[width=\imagesize]{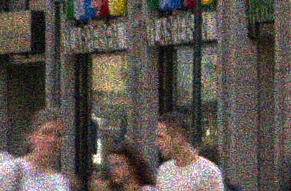}\hspace{0.8mm}%
		&
		\overimg[width=\imagesize]{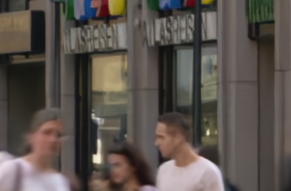}{36.18dB}\hspace{0.8mm}
		&
		\overimg[width=\imagesize]{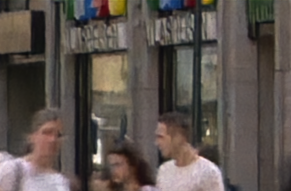}{33.23dB}\hspace{0.8mm}
		&
		\overimg[width=\imagesize]{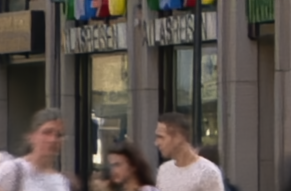}{35.73dB}\\
		& \small Noisy frame & \small FastDVDnet (supervised)  & \small DnCNN (F2F, noise blind) &  \small FastDVDnet (MF2F, noise blind) 
		\end{tabular}
\gfsqueeze	
	\caption{Denoising results on Poisson noise ($p=8$), Box noise ($\sigma=40$, $3\times{}3$) and demosaicked Poisson noise ($p=4$). From left to right: noisy frame; FastDVDnet \protect\cite{tassano2019fastdvdnet} trained for each noise  (supervised); DnCNN \protect\cite{Zhang2017BeyondDenoising} fine-tuned with frame-to-frame (F2F) \protect\cite{ehret2019model} (self-supervised); FastDVDnet fine-tuned with the offline version of the proposed multi-frame-to-frame (MF2F) framework (self-supervised). 
	}\label{fig:teaser}
	\end{center}
\end{figure*}

The current state of the art in image and video denoising is dominated by convolutional neural networks (CNNs)~\cite{Zhang2017BeyondDenoising,16-santhanam-rbdn,Liu2018-nl-rnns,vnlnet_jmiv2020,tassano2019fastdvdnet}.
In addition to their superior performance, CNNs offer a greater flexibility as they can be trained to denoise potentially any type of noise~\cite{18-chen-see-in-the-dark,wang2017sar,kang2017deep,chang2019-blind-medical-denoising}. 
In contrast, traditional model-based approaches typically require 
a tractable model of the noise, and specific algorithms for each type of noise (\emph{e.g.}~\cite{lebrun2015noise,18-gonzales-denoising-decompression,maggioni2014joint,salmon2014poisson,coupe2009nonlocal,boulanger2009patch,zhao2019ratio}).
This flexibility however, comes at a price, as it has been observed that CNNs are very sensitive to mismatches between the data and noise distributions at training and testing~\cite{plotz2017benchmarking}.
This has fueled the interest in training CNNs for real noisy raw images, with the publication of several datasets and benchmarks~\cite{Nam_2016_CVPR,plotz2017benchmarking,18-chen-see-in-the-dark,SIDD_2018_CVPR,Brummer2019-natural-image-noise}, as well as methods~\cite{guo2018toward,brooks2019unprocessing,Plotz2018-NNN,kim2020-transfer-synth-to-real-noise,kim2020-transfer-synth-to-real-noise,zamir2020cycleisp}. Most of this research focuses almost exclusively on still image denoising.

Producing datasets of realistic noisy-clean pairs for supervised training is a challenging task. Some works contaminate clean images with synthesized realistic noise~\cite{guo2018toward,brooks2019unprocessing,kim2020-transfer-synth-to-real-noise,zamir2020cycleisp}, 
but the results depend on the fit between the synthetic and real data. {Realistic data can not be always be generated. For example the noise distribution might be unkown.}  %
Generative Adversarial Networks have been proposed to generate samples from a unknown noise distribution~\cite{chen2018-GAN-noise-modelling}.
Other works have proposed datasets {of real images with ground truth.}
For still image denoising, it is possible to acquire pairs of images of exactly the same scene, either altering the exposure time so that one of them is approximately noiseless~\cite{plotz2017benchmarking,18-chen-see-in-the-dark,chen2019seeing-motion}, or by taking a second noisy shot with an independent noise realization as proposed by \emph{noise-to-noise}~\cite{Lehtinen2018}.
Acquiring such pairs with real noise can be cumbersome and prone to dataset biases as the scenes need to be static. For video denoising the situation is even worse as it would require independent acquisitions of the exact same action \cite{yue2020supervised}.

A more ambitious goal is that of \emph{self-supervised} training, where the network learns exclusively from noisy images/videos $x_i$ with a loss that uses them both as input and target, for instance $\sum_i \|\mathcal F(x_i) - x_i\|^2$. 
To prevent the network from learning the identity function, 
restrictions are incorporated in the architecture. Denoising autoencoders~\cite{stacked-denoising-autoencoders-2010} use a bottleneck forcing the network to filter out information.
\emph{Blind-spot} networks~\cite{noise2void,batson19aN2S} do not have access to the input pixel at $j$
for computing the output pixel $j$ (a blind spot at the center of the receptive field). 
This has a significant penalty on the performance, as the noisy value of a pixel is a valuable piece of information for denoising it.
Some works re-introduce the blind spot in a second Bayesian estimation step
\cite{BSS,krull2019probabilisticN2V}, but this requires knowing the noise distribution. 
A related approach is proposed in~\cite{Quan2020self2self}, where a fraction of input pixels is masked at random, and the network then learns to do joint denoising and inpainting. Averaging predictions obtained with different random masks leads to results comparable to supervised training.
Blind-spot networks fail if the noise is spatially correlated. 
Spatially correlated noise is handled in~\cite{Moran2020noisier2noise}, but it requires knowing the parameters (\emph{e.g.} variance) of the noise distribution. 
Other works~\cite{metzler2018sure,soltanayev2018sure} approximate the MSE risk with an unsupervised loss using Stein's unbiased risk estimator (SURE)~\cite{stein1981}. 
Unfortunately, these approaches require the noise distribution to be known and cannot be applied to other risks.

The treatment of real noise in videos is beginning to attract more attention. Patch-based approaches have been proposed for handling correlated noise in compressed~\cite{Liu2010} and infrared videos~\cite{maggioni2014joint}. 
In~\cite{mildenhall2018kpn-bursts,marinc2019mkpn-bursts,claus2019videnn} CNNs are trained by synthesizing signal dependent noise. In~\cite{chen2019seeing-motion} a still image denoising network is trained on low-light static sequences using a long exposure image as ground truth. The authors add a temporal consistency {term to the} loss to improve generalization to dynamic scenes.
In~\cite{yue2020supervised} the authors train a video denoising network for raw video using a complex combination of simulated noise, datasets of long and short exposure raw images and a dataset of stop motion raw videos. In the latter each video is actually a sequence of static scenes. In this way several images can be captured for each video frame, and averaged to reduce the noise.
Recently, Ehret et al.~\cite{ehret2019model} proposed \emph{frame-to-frame} (F2F), a method to fine-tune an image denoising network (or joint denoising and demosaicking~\cite{ehret2019join}) to an unknown noise type from a single noisy video. The fine-tuning is based on a loss that penalizes the motion compensated error between the predicted frame and the previous noisy frame as \emph{target}. The fine-tuned network achieved (and even surpassed) the performance of the same network trained with supervision for that specific noise. 
An important limitation of F2F is that its single-frame denoising network leads to sub-optimal video denoising results and lacks of temporal consistency.

\myparagraph{Contributions.} We introduce \emph{multi-frame-to-frame} (MF2F), a self-supervised fine-tuning framework for video denoising networks that take a \emph{stack} of several frames as input. The proposed fine-tuning allows to adapt a multi-frame  network to an unknown noise type using a single noisy sequence.
This extends the single-image F2F approach of~\cite{ehret2019model} to multi-frame 
networks, resulting in a model blind video denoising method that achieves, \emph{for the first time}, results on par with those of non-blind state-of-the-art methods.

{Naively applying F2F to a multi-frame network leads to unwanted trivial solutions, as the target frame in the loss is part of the input stack. This applies to any loss in which the target is a function of the input to the network.
}
We evaluate different configurations of non-overlapping input stacks and target frames and identify the ones yielding the best performance. 
We also found that the fine-tuned network leads to even better results if we switch back to the standard input stack at inference time.

{We call MF2F the fine-tuning method resulting from applying F2F with the proposed training stack.}
We demonstrate the effectiveness and flexibility of the proposed MF2F by fine-tuning a network pre-trained for additive white Gaussian noise (AWGN) to different noise types (AWGN, Poisson, colored Gaussian and demosaicked Poisson) and levels. The results are comparable to those of supervised noise-specific training
(see Fig.~\ref{fig:teaser}). Evaluations on videos with real and realistic~\cite{brooks2019unprocessing} camera noise show that MF2F outperforms state-of-the-art raw video denoising networks.

The proposed fine-tuning can also be applied to any network parameter which influences the denoising performance. We illustrate this by working with an AWGN denoising network that receives as input a noise variance map. Fine-tuning the variance map allows to jointly estimate the variance at each input pixel and denoise the video. We apply this to heteroscedastic Gaussian and Poisson noises. {The proposed fine-tuning is able to recover complex variance maps with remarkable spatial resolution.}

In Section~\ref{sec:method} we describe the proposed framework. 
We validate our approach on synthetic data in Section~\ref{sec:experiments} and on real noisy sequences in Section~\ref{sec:real}. Concluding remarks are given in Section~\ref{sec:conclusions}.

\section{Self-supervised Video Denoising}
\label{sec:method}

\begin{figure}[t]
    \centering
    \includegraphics[width=.65\linewidth]{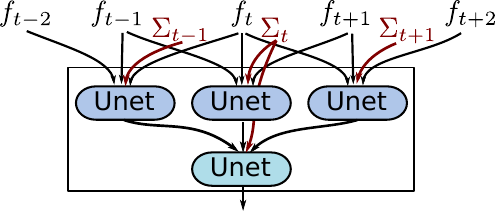}
\gfsqueeze
    \caption{The FastDVDnet \protect\cite{tassano2019fastdvdnet} architecture consists of two cascaded U-nets \protect\cite{Ronneberger2015U-Net:Segmentation}, each of which takes as input three frames (without alignment), plus a variance map $\Sigma_i$ of the same size as the input frames. The first U-net is applied three times to produce initial estimates of the frames {$t-1$, $t$ and $t+1$}.
    These estimates are then fed into the second network which predicts the central frame $\hat u_t$. 
    }
    \label{fig:fastdvdnet}
\end{figure}

We consider a video $f$ with frames $f_t$, that is a noisy version of a video $u$. The distribution of the noise is unknown. We assume that the noise at each frame is independent and  median preserving in the noise-to-noise sense~\cite{Lehtinen2018,ehret2019model}.

Our self-supervised loss for video denoising extends the F2F loss introduced in~\cite{ehret2019model}, which penalizes the error between the output of the network at frame $t$ with {the noisy} frame $t-1$ (the \emph{target} frame).
The authors of~\cite{ehret2019model} consider a denoising network $\mathcal{F}_\theta$ which takes a single image as input, and train it via the following loss:
\begin{equation}
    \ell_1^{\text{F2F}} \left( \mathcal F_\theta(f_t), f_{t-1} \right) = \Vert \kappa_t\circ( W_{t,t-1} \mathcal F_\theta(f_t) - f_{t-1}) \Vert_1.
    \label{eq:f2f-loss}
\end{equation}
Here $\circ$ denotes the element-wise product, $W_{t,t-1}$ the warping operator from frame $t$ to the target frame $t-1$, and $\kappa_t$ is an occlusion mask removing mismatches from the loss. 
{Given the optical flow $v_{t-1,t}$, the warping operator from frame $t-1$ to $t$ is defined as
\begin{equation}
    (W_{t,t-1}u_t)(x) = u_t(x + v_{t-1,t}(x)), 
    \label{eq:warping}
\end{equation}
where bicubic interpolation is used to resample the $u_t$.}

\begin{figure}[t]
    \centering
    \includegraphics[width=.8\linewidth]{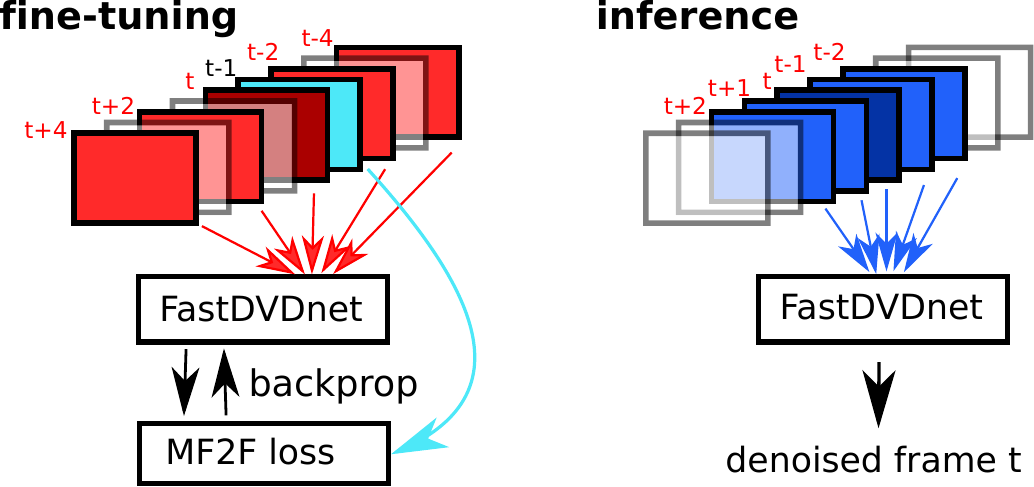}
\gfsqueeze
    \caption{Proposed multiframe-to-frame blind fine-tuning for a video denoising network taking as input a stack of frames.
    During fine-tuning we use a dilated input stack (in \textcolor{red}{red}) so that the target frame is hidden from the network. At inference time we use the natural stack (in \textcolor{blue}{blue}). }
    \label{fig:arch-training}
\end{figure}

In~\cite{ehret2019model} the temporal consistency of the video is used to train the network but the network itself takes as input a single image. Better results can be obtained by network architectures that take into account temporal information.
This can be done with frame recurrent networks~\cite{chen2016deep} or by providing multiple frames as the input of the network~\cite{xue2019,vnlnet_icip2019,tassano2019fastdvdnet}. We focus on the latter type of networks as they {currently produce} state-of-the-art results in video denoising. 
In particular, we will adopt the recent FastDVDnet~\cite{tassano2019fastdvdnet}  (see Fig.~\ref{fig:fastdvdnet}), but the proposed fine-tuning can be applied to other multi-frame networks as well.
The FastDVDnet architecture is  well suited as it takes five frames as input as well as variance maps. Furthermore, it can be trained end-to-end without requiring an external motion estimation stage. 

We denote the input stack of frames as
$\mathcal S_t = [f_{t-n},...,f_{t+n}].$ 
The $t^\text{th}$ denoised frame is produced as $\hat u_t = \mathcal F_\theta(\mathcal S_t)$ (to simplify the notation we will omit the input variance maps $\Sigma_{r}$ with $r = t-1,t,t+1$).

\subsection{Frame stacks for self-supervised training}

The F2F loss \eqref{eq:f2f-loss} cannot be directly applied to $\mathcal F_\theta(\mathcal S_t)$, as 
it can be minimized simply by warping $f_{t-1}$ (which is in the input stack) to $W^{-1}_{t,t-1}f_{t-1}$, \emph{i.e.} by aligning the noisy frame $f_{t-1}$ to $f_t$ without removing the noise. 
In the following we show that any loss that depends only on the network input and its output leads to these unwanted solutions.

Suppose we want to predict $y$ from $z$. In our case, $z$ is a stack of noisy frames and $y$ the clean version of the central frame of the stack, however the following arguments also apply to other regression problems. In a supervised training setting we minimize an approximation of the expected value of a loss penalizing the difference between the network output and the desired output $y$. The optimal estimators $\hat y = \mathcal F^*(z)$ capture some aspects of the data distribution, which is the point of a data-driven approach. For instance, for the MSE loss we have $\mathcal F^*(z) = \mathbb E\{y|z\}$~\cite{kay1993fundamentals}.

We would like to configure the input stack to our network so that the fine-tuning can learn such data-driven estimators. The following observation restricts the number of options.

\smallskip\noindent\textbf{Observation 1.} \textit{Let $(z,y)$ distributed according to $p(z,y)$. An estimator $\hat y(z) = \mathcal F^*(z)$ minimizing the expected value of a loss $\mathbb E\{\ell(\mathcal F(z),z)\}$ that depends only on $z$ and $\mathcal F(z)$ is independent from the data distribution $p(z,y)$.}

\smallskip

As the loss depends only on the input $z$ and $\mathcal F(z)$, the minimization of $\mathbb E\{\ell(\mathcal F(z),z)\}$ can be done for each input $z$ independently, \emph{i.e.} 
$ 
\mathcal F^*(z) = \argmin_{\hat y} \ell(\hat y, z).
$
Given $z$, the optimal estimator is the minimizer of the loss for that specific $z$. As a consequence it is independent of the data distribution and only depends on the chosen loss. 

Observation 1 implies that the reference frame
{cannot be a function of the input stack} if we want to have a data-driven estimator. 
Therefore, for our input stack we will adopt a solution similar to that of blind spot networks \cite{noise2void,batson19aN2S}: remove the target frame from the input stack. 
Denoting by $\mathcal S'_t$ the fine-tuning stack, we then minimize the following \emph{multi-frame to frame} (MF2F) loss:
\begin{equation}
    \ell_1^{\text{MF2F}} \left( \mathcal F_\theta(\mathcal S'_t), f_{t-1} \right) = \Vert \kappa_t\circ( W_{t,t-1} \mathcal F_\theta(\mathcal S'_t) - f_{t-1}) \Vert_1,
    \label{eq:mf2f-loss}
\end{equation}
where the warping operator $W_{t,t-1} $ is defined in \eqref{eq:warping}.

\begin{table}[t]
    \centering
    {\small
    \begin{tabular}{l| c | c | c | c }
		 
		 Training stack $\mathcal S'_t$   & ref. & Box & AWGN  & Poiss.\\\hline
         $f_{t-2}, f_{t-1}, f_{t}, f_{t+1}, f_{t+2}$ & $f_{t-3}$ &    28.93  & 28.78 & 28.13 \\
         $f_{t-3}, f_{t-1}, f_{t}, f_{t+1}, f_{t+2}$ & $f_{t-2}$ &    32.02 & 32.25 & 31.18 \\
         $f_{t-3}, f_{t-2}, f_{t}, f_{t+1}, f_{t+2}$ & $f_{t-1}$ &    \B{36.23}  & 37.25 & \B{35.20}\\
         $f_{t-4}, f_{t-2}, f_{t}, f_{t+2}, f_{t+4}$ & $f_{t-1}$ & \B{36.22} & \B{37.32} & \B{35.21}\\\hline
       FastDVDnet superv. & n/a & 36.58  & 37.29 & 35.82\\
    \end{tabular}}
\gfsqueeze
    \caption{Results for different reference frames and training stacks $\mathcal S'_t$. During inference after training is done, we denoise the video \linebreak using the natural stack $\mathcal S_t$.
    We report the average PSNR of the online fine-tuning over 17 sequences from the  Derf \protect\cite{montgomeryxiph} and Vid3oC-10 \protect\cite{kim2019vid3oc} datasets, excluding the first 10 frames of each video (to allow for the adaptation time). We considered 3$\times$3 box noise with $\sigma$=40, AWGN $\sigma$=20 and scaled Poisson noise with  $p=8$.}
    \label{tab:stacks}
\end{table}

We evaluated the fine-tuning with different configurations of disjoint stack and reference frame: introducing the blind spot at different distances from the central frame or evenly spacing the frames in the stack. 
Table~\ref{tab:stacks} summarizes the results. 
We found that: (1) The target frame has to be as close as possible to the denoised frame. Otherwise, the quality of the alignment degrades, negatively impacting the fine-tuning.
(2) The results can be slightly improved using the \emph{dilated stack} $[f_{t-4},f_{t-2},f_t,f_{t+2},f_{t+4}]$. 

We also observed that, regardless of the fine-tuning stack, the best way to perform the inference is by using the natural stack. Frames in the natural stack are more temporally correlated and this helps improving the denoising. 
The results reported in Table~\ref{tab:stacks} were obtained using the natural stack for testing. The results obtained using the training stack at testing time (which can be found in the supplementary material) are on average 0.3dB below the ones shown in the table.   
In our remaining experiments we will always use the dilated stack for fine-tuning and the natural stack for final inference as illustrated in Fig.~\ref{fig:arch-training}.

\begin{figure}[t]
    \centering
    \def\s{0.24\linewidth}
       \overimg[width=\s]{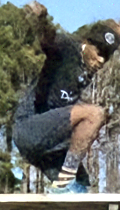}{\rotatebox{90}{F2F without mask - 29.82dB}}
       \overimg[width=\s]{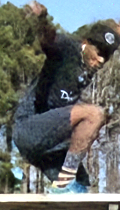}{\rotatebox{90}{F2F with  mask - 29.97dB}}
       \overimg[width=\s]{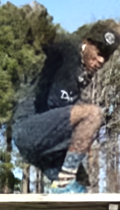}{\rotatebox{90}{MF2F without mask - 32.76dB}}
       \overimg[width=\s]{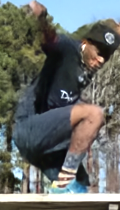}{\rotatebox{90}{MF2F with  mask - 33.04dB}}
\gfsqueeze    
    \caption{Effect of the mask. Errors in the optical flow  create mismatches that negatively influence the fine-tuning (\emph{e.g.}~transparency of right arm, hallucinated texture). We detect occlusions and warping errors and remove them from the loss via a binary mask. 
    }\label{fig:effect-of-mask}
\end{figure}

\begin{table*}[t]
\centering
{\footnotesize
\setlength{\tabcolsep}{0.68em}
\begin{tabular}{|c|l|c|c|c||c|c||c|c|c|} 
\hline
\multicolumn{2}{|c|}{}  & \multicolumn{3}{c||}{Non-blind}  & \multicolumn{2}{c||}{AWGN $\sigma$ blind} & \multicolumn{3}{c|}{Model blind}   \\\hline
\multicolumn{2}{|c|}{}  &  \multicolumn{2}{c|}{FastDVDnet supervised} &                    & online MF2F   & online MF2F  & online F2F & online MF2F & offline MF2F      \\
\multicolumn{2}{|c|}{}  &  noise specif. & multi task & VBM3D              & scalar sigma & 8 sigmas                   & weights & weights & weights   \\
\hline

\multirow{7}{*}{\rotatebox{90}{Derf}} 
& Gaussian 20  & 36.96 / .946 & 36.49 / .942 & 36.21 / .933  & 36.92 / .946 & 36.90 / .945 & 34.53 / .909 & 37.32 / .951 & \B{\ul{37.48}} / \B{\ul{.952}}  \\
& Gaussian 40  &  34.00 / .907 & 33.70 / .903 & 32.63 / .871 & 33.95 / .906 & 33.91 / .905 & 32.04 / .847 & 34.24 / .913 & \B{\ul{34.27}} / \B{\ul{.914}} \\
& Poisson 1   & 40.45 / \ul{.974} & 39.71 / .970  & 38.99 / .959 & 39.50 / .960 &   40.15 / .972  & 36.56 / .953 & 40.39 / \B{\ul{.974}} & \B{\ul{40.51}} / \B{\ul{.974}} \\
& Poisson 8   &   \ul{36.00} / .939 &  35.58 / .935 & 34.18 / .897 & 34.15 / .890 & \B{35.73} / .934 & 31.98 / .890 & 35.57 / .941 & 35.68 / \B{\ul{.942}}  \\
& Box $3\times 3$,  40   &  35.42 / \ul{.932} &  34.86 / .924 & 29.94 / .757 & 34.13 / .900 & 34.34 / .902  & 32.55 / .891 & 35.51 / \B{.927} & \B{\ul{35.60}} / \B{.927} \\
& Box $5\times 5$,  65  & \ul{34.78} / \ul{.932} & 33.98 / .919 & 28.37  / .736 & 32.14 / .873 & 32.60 / .888  & 31.78 / .886 & 34.29 / \B{.928} & \B{34.35} / \B{.928} \\
& Demosaicked 4    & \ul{34.85} / .926 & 25.53 / .533 & 33.16 / .890 & 33.23 / .877  & 34.30 / .916 & 32.61 / .885 & 34.75 / .926 &       \B{34.81} / \B{\ul{.927}}  \\\hline

\multirow{7}{*}{\rotatebox{90}{Vid3oC-10}} 
& Gaussian 20   & 37.49 / .964 &  36.84 / .959 & 36.55 / .951  & 37.43 / .963 &    37.40 / .963  & 31.62 / .868  & 37.32 / .964  & \B{\ul{37.55}} / \B{\ul{.966}} \\
& Gaussian 40   & \ul{34.27} / .937 & 33.70 / .929 & 32.74  / .901 & 34.23 / .935 &  34.21 / .935  & 29.10 / .799  & 34.17 / .937 & \B{34.26} / \B{\ul {.938}}  \\
& Poisson 1    & \ul{40.63} / \ul{.980} & 39.71 / .975 & 39.32 / .967   & 39.30 / .966 & \textbf{40.29} / .978 & 33.54 / .905  & 40.01 / .978 & 40.16 / \B{.979}   \\
& Poisson 8    & \ul{35.72} / \ul{.951} & 35.09 / .944 &  32.16 / .844  & 33.26 / .887 & \textbf{35.42} / .946 & 29.92 / .823  & 34.99 / \B{.947}  & 35.00 / \B{.947}   \\
& Box $3\times 3$,  40    & \ul{37.28} / \ul{.964} & 36.54 / .957 & 30.19 / .770  & 34.70 / .936 &    34.90 / .938  & 31.27 / .871  & 36.65 / \B{.963}  &       \B{36.76} / \B{.963}  \\
& Box $5\times 5$,  65   & \ul{36.81} / \ul{.965} &  35.84 / .955 & 28.53 / .746  & 32.65 / .909 &    33.11  / .917 & 30.75 / .869 & 35.65 / .956  & \B{35.79}  / .957 \\
& Demosaicked 4    & \ul{34.50} / \ul{.941}  & 23.96 / .508 & 32.31 / .882  & 32.24 / .876 &    33.86 / .931  & 31.59 / .890  & 33.95 / .933  & \B{33.98} / .934  \\
\hline
\end{tabular}}
\gfsqueeze
\caption{Average PSNR and SSIM over all the sequences for a given dataset and type of noise. The MF2F fine-tuning is applied to a FastDVDnet network \protect\cite{tassano2019fastdvdnet}, either on the weights (model blind)  or the input variance map ($\sigma$ blind). F2F fine-tuning is applied to the weights of a single frame DnCNN network \protect\cite{Zhang2017BeyondDenoising}. The best PSNR in each case is \ul{underlined}. The best blind method is in {\bf bold}. }
\label{tab:allresults}
\end{table*}

\subsection{Handling warping errors}

Following~\cite{ehret2019model}, {the warping transformations are estimated using the TV-L1 optical flow method}~\cite{zach2007duality,sanchez2013tv}, as it gives consistent results across noise types and intensities. Moreover, it is based on minimizing the photometric distance between pixels, which is precisely what we need for our loss. 

In Equation \eqref{eq:mf2f-loss} the mask $\kappa$ is zero for regions where a misalignement is likely, and one otherwise. 
Alignment errors are defined as the union of occlusions computed from the optical flow and regions with a large warping residual
\begin{equation}
r_{t,t-1} = g \ast \|W_{t,t-1}(g \ast f_t) - g \ast f_{t-1}\|_1,
\end{equation}
where $g$ is Gaussian filter of standard deviation $\sigma=2$ used to obtain a rough estimate of the clean video. The warping residual is then thresholded with a robust adaptive threshold \vd{}{\pa{}{(see supp. material)}}. {This mask differs from the one used in~\cite{ehret2019model}, which is based on the divergence of the optical flow.}

The mask $\kappa$ is particularly important for fine-tuning a multi-frame denoising network. Indeed, since the network has access to multiple input frames, it is likely that the mismatched target can be found in the stack.  The result in Fig.~\ref{fig:effect-of-mask} confirms this. We see that the result of F2F barely changes with or without mask. However, applying MF2F without a mask leads to a degradation of the result. 

All the experiments in the following sections were computed using the same parameters for the optical flow and misalignment mask.

\subsection{Fine-tuning and inference}

Similarly to~\cite{ehret2019model}, the proposed fine-tuning can be done online or offline.
In the offline setting, the network is first  fine-tuned on the entire video, which is considered as a dataset of frames. We form batches by randomly sampling frame stacks from the video and update the network parameters by performing one optimizer step per batch. This is repeated a fixed number of epochs. 
Afterwards, the fine-tuned parameters are  used to denoise the video by applying the network to each frame using the natural input stack.

The online setting defines a time-varying sequence of network parameters, and can thus adapt to temporal changes in the distribution of the noise or the signal. The video is processed sequentially, applying the following two steps on each frame.
First the network parameters are updated by performing a fixed number $N$ of optimizer steps of the loss \eqref{eq:mf2f-loss}.
Then the denoised output is produced by applying the updated network on the natural input stack.


The proposed fine-tuning can be applied to the network weights, or any other parameter that has an influence on the denoising performance, such as the variance map $\Sigma_t$ that FastDVDnet takes as input (see Fig.~\ref{fig:fastdvdnet}).
In FastDVDnet~\cite{tassano2019dvdnet} the authors consider only homoscedastic AWGN, and use therefore a constant image $\Sigma_t(x) = \sigma^2$ as variance map.
The input variance map controls the denoising strength and the correct value has to be provided at inference time.
In Section \ref{sec:experiments_sigmap_map} we apply the MF2F fine-tuning to simultaneously estimate the variance map and denoise the video. This is particularly relevant when the noise is AWGN, but its variance map unknown.

In all the experiments in upcoming sections we use the same hyper-parameters when the fine-tuning is done with respect to the weights. In the online setting we use a learning rate of $10^{-5}$ and $N=20$ iterations of the Adam optimizer on mini-batches consisting of pairs of frames (\emph{i.e.} the weights are updated each two frames). 
In the offline setting we use the same learning rate and perform $200$ Adam iterations with mini-batches of $20$ frames (no improvement was observed with more iterations). When the fine-tuning is applied to the variance map, we adapt the learning rate and keep the same number of iterations and batch sizes.

\section{Experiments on synthetic noise}\label{sec:experiments}

\pa{In this section we present experimental results discussing several aspects of the proposed framework using a wide range of synthetic noise types and use cases.}
{We now present results of the proposed framework for a wide range of synthetic noise types and use cases.}
The  evaluation is performed on videos from two datasets. 
One is a set of seven Full HD videos of $100$ frames each extracted from the Derf’s Test Media collection~\cite{montgomeryxiph}. 
The second, 
more challenging, dataset consists of ten videos of $120$ frames 
extracted
from the training {split of the Vid3oC dataset \cite{kim2019vid3oc}. We refer to this dataset as Vid3oC-10.}
We downscaled all videos by a factor of two. 

In our experiments we consider four noise types (1) AWGN noise, (2) scaled Poisson noise with scaling parameter $p$ (the mean of the noisy pixel $f_i$ is the clean pixel $u_i$, and the variance is $pu_i$), (3) correlated noise (denoted \emph{box noise}) obtained by filtering AWGN with an $s\times s$ box filter  and (4) demosaicked Poisson noise, obtained by mosaicking the image, adding scaled Poisson noise and then applying the demosaicking algorithm of~\cite{kiku2014mlri-demosaicking}. For the first three types we consider two noise levels, indicated in the first columns of Table~\ref{tab:allresults}.
The demosaicked Poisson noise simulates the correlation introduced by a demosaicking algorithm applied on the noisy data. 
We evaluate the average PSNR and SSIM for the given sequence using the ground-truth, but excluding the first $10$ frames of each sequence. 
\vd{}{\pa{}{Additional qualitative comparisons of the methods can be found in the supplementary material.}}

We remark that all results are obtained by fine-tuning a network independently on each video. In all cases the fine-tuning starts from weights pre-trained for AWGN with $\sigma = 25$. The proposed fine-tuning scheme can be applied to any denoising network and any pre-trained weights can be used as starting point.
In the supplementary material we studied the impact of the starting point on the performance of the fine-tuned network for different noise types. Unsurprinsingly, the 
the speed of the adaptation depends on the similarity between the pre-trained and the target noise distributions. We observed that it is easier for the AWGN weights to adapt to other noise types.

\subsection{Fine-tuning the network weights}

In the following experiments we apply the proposed fine-tuning to the network weights of FastDVDnet, while keeping constant the input variance map. We compare with F2F~\cite{ehret2019model} which is, to the best of our knowledge, the only other blind video denoising method in the literature.
As reference, we compare with three non-blind algorithms. (1) A \emph{noise-specific} FastDVDnet network trained in a supervised setting for each kind of noise. For AWGN we used the weights provided by the authors of FastDVDnet \cite{tassano2019fastdvdnet}, which work in conjunction with the input variance map. {For other noise types we retrained one FastDVDnet per noise type and intensity, using the same dataset~\cite{davis} and hyperparameters as in~\cite{tassano2019fastdvdnet}.} 
(2) A {\em multi-task} FastDVDnet trained to handle multiple noise types: Gaussian  ($\sigma=20$ and $\sigma=40$), Poisson  ($p=1$ and $p=8$), and our box noise ($3\times3, \sigma=40$ and $5\times5, \sigma=65$). 
(3) The VBM3D~\cite{Dabov2007v} path-based method in which the noise parameter is set to yield the best result.

In Table~\ref{tab:allresults}, we report quantitative results obtained with the proposed {MF2F fine-tuning applied on the network weights both in the online and the offline settings}. 
From the results we can see that the performance of the offline MF2F method is slightly superior to the online one.
Compared against the multi-task network the proposed MF2F fine-tuning always attains better results. Furthermore, the multi-task network cannot handle demosaicking noise  (which was unseen during training) while our self-supervised networks compete with the network specifically trained with supervision for that  noise type.

For most noise types, the results of our self-supervised MF2F fine-tuning are close to those obtained with the noise-specific FastDVDnet network trained with supervision. 
This is confirmed in Fig.~\ref{fig:teaser}, where we compare some results of the offline method with the noise-specific FastDVDnet (supervised), and F2F.

In all experiments we observe a consistent PSNR gain of about 3dB with respect to F2F. This corresponds to a noise reduction of a factor 2, which is expected from a network that exploits the redundancy of  5 frames compared against a single-frame method.

\begin{figure}[t]
    \centering
    \begin{subfigure}{0.48\linewidth}
        \vspace{0.25cm}\includegraphics[clip,trim={10 0 0 0},width=\linewidth]{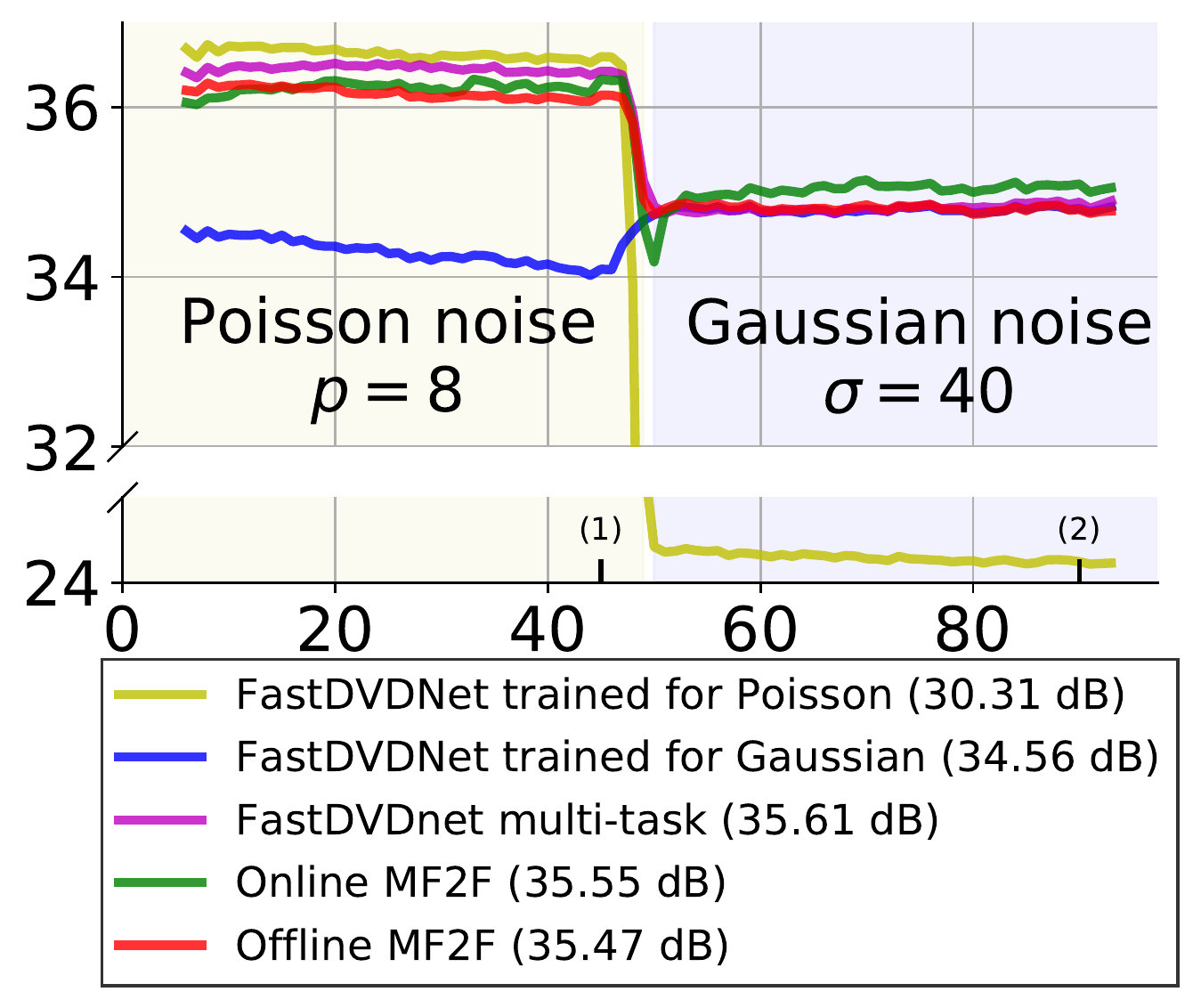} 
    \end{subfigure}
    \begin{subfigure}{0.48\linewidth}
        \centering
            \def\s{0.5\linewidth}
            \fboxsep=0.0mm
            \fboxrule=1.25pt
            \scriptsize{(1): Poisson noise~~~~~(2): Gaussian noise} 
            \\[0.1em]
            \color{Goldenrod}
            \fbox{\includegraphics[width=\s]{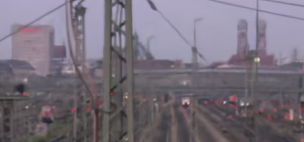}}%
            \fbox{\includegraphics[width=\s]{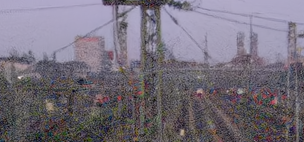}}
            \color{blue}
            \fbox{\includegraphics[width=\s]{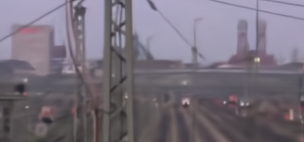}}%
            \fbox{\includegraphics[width=\s]{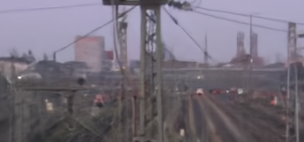}}
            \color{Green}
            \fbox{\includegraphics[width=\s]{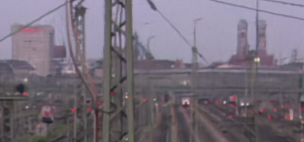}}%
            \fbox{\includegraphics[width=\s]{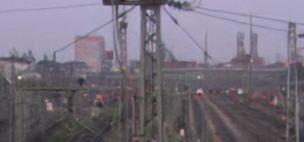}}
    \end{subfigure}
\gfsqueeze
    \caption{Adaptation to changes in the noise properties. We simulate a sequence with Poisson noise for the first half and Gaussian noise for the second half. The frames (1) and (2) corresponds to Poisson and Gaussian noise respectively. The pretrained methods for the specific noise types perform poorly on the other half (yellow and blue), while the proposed methods (online and offline) are able to cope with the abrupt change.}
    \label{fig:plot_noise_change}
\end{figure}

\myparagraph{Time varying noise.} The online fine-tuning of the weights permits to quickly adapt to changes in the noise properties. The PSNR plot in Fig.~\ref{fig:plot_noise_change} shows the per-frame PSNR computed on a video in which the noise switches from Poisson ($p=8$) to Gaussian ($\sigma=40$) at frame 50.
While the two noise-specific FastDVDnet networks perform well for their respective noise types, their performance strongly degrades for the other type.  
On the other hand, the proposed MF2F approaches are able to cope with the abrupt change of noise, and the online version even outperforms the network trained for Gaussian on the Gaussian section. 
The offline method is able to handle both noise types  by learning to denoise them with the same network.
The results obtained with MF2F are on par to those of a multi-task FastDVDnet trained with supervision for these two noise distributions.
The crops in Fig.~\ref{fig:plot_noise_change} show that the noise-specific networks failed for the other noise types (as expected), while the online method is able to restore fine details for both noise types.

\begin{figure}
    \centering

\begin{tabular}{@{}c@{\,\,}c}
          \includegraphics[clip,trim=15px 28px 30px 3px, width=0.47\linewidth]{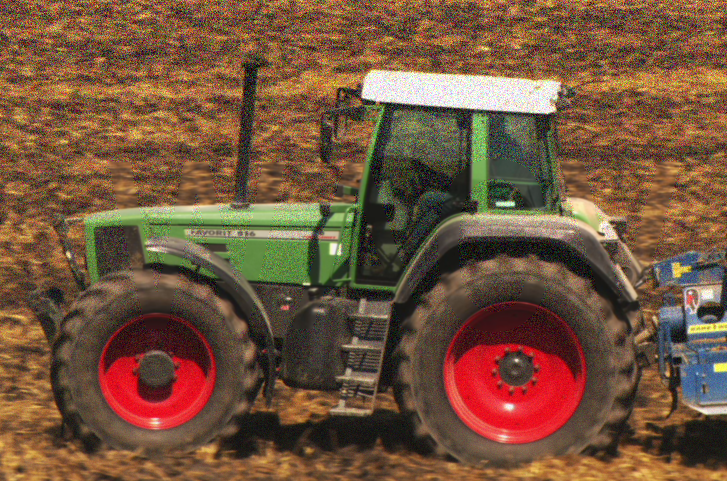}   & 
          \includegraphics[clip,trim=15px 28px 30px 3px, width=0.47\linewidth]{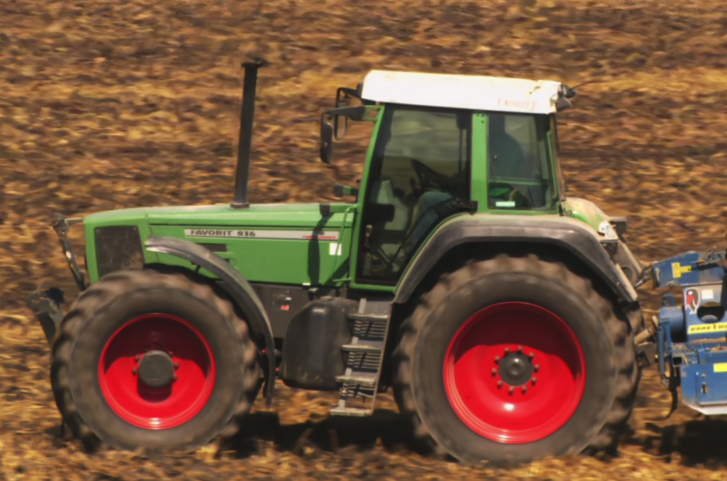} \\
        \includegraphics[clip,trim=15px 28px 30px 3px, width=0.47\linewidth]{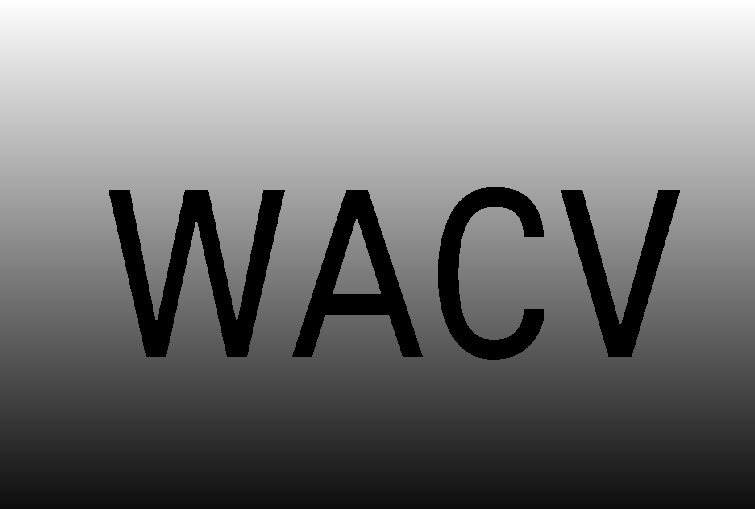} &
        \includegraphics[clip,trim=15px 28px 30px 1px, width=0.47\linewidth]{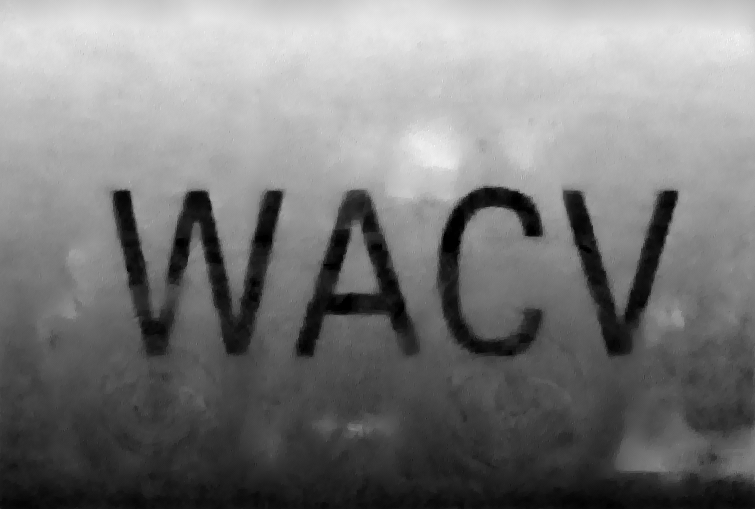} 
\end{tabular}

\gfsqueeze
	\caption{Spatially variant variance map used in a synthetic experiment. From left to right and top to bottom: a noisy frame; the corresponding denoised by the proposed online MF2F; variance map of the added Gaussian noise; and the variance map obtained  by using the proposed  fine-tuning of the input variance map. The noise level of the letters is $\sigma=0$ and varies linearly in the background from $\sigma=0$ to $40$; }

	\label{fig:exp_spatial_noise_map_WACV}
\end{figure}

\subsection{Fine-tuning only the noise map}
\label{sec:experiments_sigmap_map}

The proposed fine-tuning can be used to estimate the input variance map of the network, while keeping the network weights fixed.  This is done by minimizing the MF2F loss~\eqref{eq:mf2f-loss} with respect to the variance map $\Sigma_t(x)$. To highlight the flexibility of the framework, we examine three parameterizations of the input variance map. All experiments in these section were performed with the online fine-tuning. Thus, a different input variance map is estimated for each frame, and used as the starting condition for the next frame.

\myparagraph{Homoscedastic AWGN.} If a sequence has homoscedastic AWGN of unknown variance $\sigma^2$, we set a constant variance map  $\Sigma_t(x) = \sigma^2$. By fine-tuning only with respect to $\sigma$, we obtain a  $\sigma$-blind denoiser (\emph{scalar sigma} in Table~\ref{tab:allresults}). 
The convergence is very fast and after a few frames the estimated $\sigma$ stabilizes around the real noise variance. If the noise is not homoscedastic AWGN, a compromise value for $\sigma$ is found.

\myparagraph{Heteroscedastic AWGN.} We now consider AWGN with a space varying variance map which is constant in time, \emph{i.e.} $\Sigma_t(x) = \Sigma_0(x)$, and fine-tune with respect to the entire input variance map. \pa{In this case  w}
{W}e add a TV regularization term to the loss (\emph{e.g.} \cite{rudin1992nonlinear}) enforcing smoothness on the variance map. We tested this on a sequence contaminated with AWGN with variance map shown in Fig.~\ref{fig:exp_spatial_noise_map_WACV}. Our method is able to recover the message hidden in the noise variance, with remarkable spatial resolution. The convergence took around 40 frames, which is quite slower than for the scalar $\sigma$. 
\pa{To the best of our knowledge, there are no other methods in the literature}
{We are not aware of any other method} that can estimate such a noise map from a video. This could be useful in practical cases where the noise distribution is spatially variant,
such as optical images with vignetting or MRI imaging \cite{landman2009spatially-variable-mri}. In these cases the variance map is much smoother than the artificial one we used for our experiment.
Note that the ultimate goal here is not to accurately recover the noise map, but to maximize the denoising performance. For example, in a flat region, the fine-tuning might lead to an over-estimation of the variance so as to increase the amount of smoothing. This explains the differences in Fig.~\ref{fig:exp_spatial_noise_map_WACV} between the actual noise map and the one found.

\myparagraph{Signal dependent AWGN.}
Poisson noise can be approximated as AWGN with signal dependent variance map given by $\Sigma_t(x) \propto u_t(x)$.
To cope with this noise, we use a spatially-variant noise map based on the brightness of the noisy video. 
We parametrize the variance map with $K$ trainable parameters $\sigma_1, \dots \sigma_K$. We split the intensity range of the noisy video in $K$ equal intervals, each with a corresponding $\sigma_i$. The input variance map at pixel $(x,t)$ is the $\sigma_i$ corresponding to the pixel brightness. At each frame, the $\sigma_i$ are automatically determined with the proposed online MF2F fine-tuning. We set $K=8$, since it results a good trade-off between efficiency and denoising performance. 
Quantitative results are reported in Table~\ref{tab:allresults}. For Poisson and AWGN noise, the obtained results are close to the ones obtained by the noise-specific FastDVDnet trained with supervision.
Visual results for Poisson noise can be found in the supplementary material.

\section{Results on real noisy videos} \label{sec:real}

In this section we show results on real noisy videos. In these examples the network adapts not only to different types of noise, but also to signals on different domains.

\begin{figure*}
    \centering
    \def\s{0.24\linewidth}
    	\overimg[width=\s]{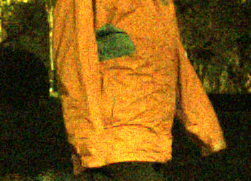}{noisy raw (demosaicked)}
    	\overimg[width=\s]{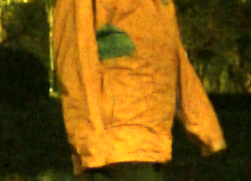}{online F2F}
    	\overimg[width=\s]{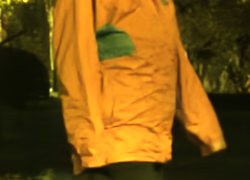}{offline MF2F}
    	\overimg[width=\s]{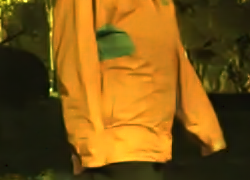}{RViDeNet}
    	 
    	\overimg[width=\s]{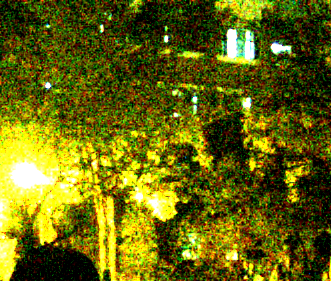}{noisy raw (demosaicked)}
    	\overimg[width=\s]{fig/crvd/enhanced-f2f-trees-037.png}{online F2F}
    	\overimg[width=\s]{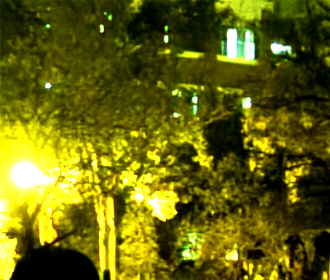}{offline MF2F}
    	\overimg[width=\s]{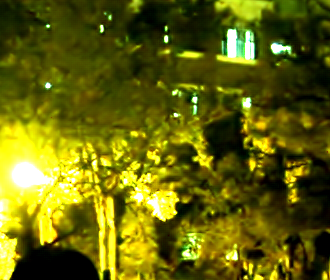}{RViDeNet}
\gfsqueeze    		
    \caption{Details from frame of a denoised raw video (ISO 12800) processed by F2F, offline MF2F, and RViDeNet. All images are demosaicked and gamma corrected.}
    \label{fig:crvd}
\end{figure*}

Fig.~\ref{fig:crvd} shows results obtained on the outdoor CRVD dataset~\cite{yue2020supervised}. This dataset consists in raw sequences of 50 frames with real noise from a surveillance camera with the sensor IMX385, at five ISO levels. Since our fine-tuning does not handle mosaicked videos, we first applied a simple tone curve ($\gamma= 2.5$) and a demosaicking algorithm~\cite{kiku2014mlri-demosaicking}. We then applied our offline blind denoising on the demosaicked noisy frames.
Although the conventional approach is to perform first denoising and then demosaicking (or even better, perform both jointly), recent work in~\cite{jin2020review} suggested that good results can also be obtained by applying the denoising after the demosaicking. 
We compare with the multi-frame denoiser of RViDeNet~\cite{yue2020supervised} which was fine-tuned on the indoor CRVD dataset, consisting on short raw stop motion sequences with ground truth. This network takes raw noisy videos as input. 
In order to obtain comparable results, the denoised raw frames are then tone-mapped and demosaicked~\cite{kiku2014mlri-demosaicking} as described above. 
From Fig.~\ref{fig:crvd} we can see that the results of MF2F 
are sharper and contain more details than those of RViDeNet, and has less residual noise than F2F (more visual results for the five ISO levels can be found in the supplementary material).

To obtain a quantitative evaluation on realistic noisy videos we follow~\cite{yue2020supervised} and use the \emph{unprocessing} network~\cite{brooks2019unprocessing} to simulate raw videos from clean RGB ones. 
These simulated videos are also used in~\cite{yue2020supervised} to pre-train (in a supervised setting) RViDeNet.
We consider the Poisson-Gaussian noise (to model the shot and read noises), using the parameters estimated in~\cite{yue2020supervised} for the ISO levels 1600, 3200, 6400, 12800 and 25600. To avoid any influence from the demosaicking step we evaluate the performance on the raw denoised videos. 
For MF2F the fine-tuning is performed independently on each video. We apply the same raw process as for the outdoor CRVD dataset: tone curve and demosaicking~\cite{kiku2014mlri-demosaicking} before denoising. Since the results are in sRGB, we re-mosaick them and invert the tone curve for evaluating the PSNR in the raw domain.
Table~\ref{tab:PSNR_per_iso_synthetic_raw} presents the average PSNRs on all the sequences for each ISO level. We can see that even though RViDeNet was pre-trained on the same noise type, MF2F performs much better.

\begin{figure}
    \centering
    \def\s{0.48\linewidth}

        \overimg[width=\s]{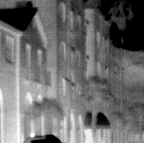}{Noisy}
        \overimg[width=\s]{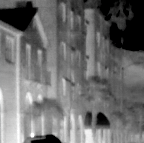}{online MF2F}
        \overimg[width=\s]{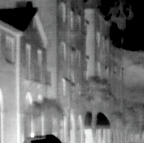}{offline MF2F}
        \overimg[width=\s]{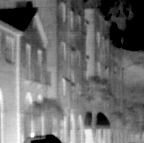}{online F2F}                    
\gfsqueeze
    \caption{Example frame of a restoration of infra-red real video. }\label{fig:flir_IR}
\end{figure}

Lastly, we also tested MF2F on thermal infra-red video from the FLIR ADAS~\cite{flir2018adas} dataset, which consists of videos taken with a Tau2-640 sensor. 
Fig.~\ref{fig:flir_IR} shows a frame from the result obtained with the offline MF2F method on a sequence of 4224 frames.

\begin{table}
\centering
\small
\begin{tabular}{|c|c|c|c|c|c|} 
\hline
 & 1600 & 3200 & 6400 & 12800 & 25600 \\
\hline
RViDeNet & 45.79 &  44.26 & 42.64 & 40.83 & 38.93  \\
Online MF2F & \B{46.51} & \B{45.13} & \B{43.64} & \B{41.91}  & \B{39.68} \\

\hline
\end{tabular}
\gfsqueeze
\caption{Synthetic raw denoising: average PSNR per ISO levels over the dataset Derf. The best PSNR in each case is in {\bf bold}. }
\label{tab:PSNR_per_iso_synthetic_raw}
\end{table}

\color{black}

\section{Conclusions}
\label{sec:conclusions}

In this work we address the problem of blind video denoising. To that aim, we extend the self-supervised fine-tuning approach introduced in~\cite{ehret2019model} to multi-frame denoising networks.
This is achieved by fine-tuning the network using a dilated input frame stack and switching back to the natural input stack at inference time.

The proposed approach demonstrates that by exploiting the temporal consistency in videos it is possible to fine-tune a video denoising network using only a few frames of a single noisy sequence and attain the performance of a network trained with supervision on a large dataset.
This also allows to handle time-varying noise, which could be useful for vision systems exposed to varying conditions (for instance a surveillance camera at day and night).

There are some interesting future research perspectives. For instance, \emph{meta-learning} could be used to improve the adaptation speed of the pre-trained network \cite{tonioni2019-learnint-to-adapt}. Also, the optical flow and the warping mask used in the MF2F loss could be computed by a network and trained at the same time, similar to \cite{yu2020joint,deudon2020highresnet}.

{\bf Acknowledgments.}
Work partly financed by IDEX Paris-Saclay  IDI  2016,  ANR-11-IDEX-0003-02,  Office  of  Naval  research grant N00014-17-1-2552,  DGA  Astrid  project  ”filmer  la  Terre”  no  ANR-17-ASTR-0013-01, MENRT. 
We acknowledge the support of NVIDIA Corporation with the donation of one Titan V GPU. 
This work was also performed using HPC resources 
from GENCI–IDRIS (grant 2020-AD011011801) and  from the “Mésocentre” computing center of CentraleSupélec and ENS 
Paris-Saclay supported by CNRS and Région Île-de-France \vd{}{(http://mesocentre.centralesupelec.fr/)}.

{\small

\begin{thebibliography}{10}\itemsep=-1pt

\bibitem{SIDD_2018_CVPR}
Abdelrahman Abdelhamed, Stephen Lin, and Michael~S. Brown.
\newblock A high-quality denoising dataset for smartphone cameras.
\newblock In {\em The IEEE Conference on Computer Vision and Pattern
  Recognition (CVPR)}, June 2018.

\bibitem{batson19aN2S}
Joshua Batson and Loic Royer.
\newblock {N}oise2{S}elf: Blind denoising by self-supervision.
\newblock In Kamalika Chaudhuri and Ruslan Salakhutdinov, editors, {\em
  Proceedings of the 36th International Conference on Machine Learning},
  volume~97 of {\em Journal of Machine Learning Research}, pages 524--533, Long
  Beach, California, USA, 09--15 Jun 2019. PMLR.

\bibitem{boulanger2009patch}
J{\'e}r{\^o}me Boulanger, Charles Kervrann, Patrick Bouthemy, Peter Elbau,
  Jean-Baptiste Sibarita, and Jean Salamero.
\newblock Patch-based nonlocal functional for denoising fluorescence microscopy
  image sequences.
\newblock {\em IEEE Transactions on Medical Imaging}, 29(2):442--454, 2009.

\bibitem{brooks2019unprocessing}
Tim Brooks, Ben Mildenhall, Tianfan Xue, Jiawen Chen, Dillon Sharlet, and
  Jonathan~T Barron.
\newblock Unprocessing images for learned raw denoising.
\newblock In {\em The IEEE Conference on Computer Vision and Pattern
  Recognition (CVPR)}, pages 11036--11045, June 2019.

\bibitem{Brummer2019-natural-image-noise}
Benoit Brummer and Christophe De~Vleeschouwer.
\newblock Natural image noise dataset.
\newblock In {\em The IEEE Conference on Computer Vision and Pattern
  Recognition Workshops (CVPRW)}, June 2019.

\bibitem{chang2019-blind-medical-denoising}
Yi Chang, Luxin Yan, Meiya Chen, Houzhang Fang, and Sheng Zhong.
\newblock Two-stage convolutional neural network for medical noise removal via
  image decomposition.
\newblock {\em IEEE Transactions on Instrumentation and Measurement}, pages
  1--1, 2019.

\bibitem{chen2019seeing-motion}
Chen Chen, Qifeng Chen, Minh~N. Do, and Vladlen Koltun.
\newblock Seeing motion in the dark.
\newblock In {\em The IEEE International Conference on Computer Vision (ICCV)},
  October 2019.

\bibitem{18-chen-see-in-the-dark}
Chen Chen, Qifeng Chen, Jia Xu, and Vladlen Koltun.
\newblock Learning to see in the dark.
\newblock In {\em The IEEE Conference on Computer Vision and Pattern
  Recognition (CVPR)}, June 2018.

\bibitem{chen2018-GAN-noise-modelling}
Jingwen Chen, Jiawei Chen, Hongyang Chao, and Ming Yang.
\newblock Image blind denoising with generative adversarial network based noise
  modeling.
\newblock In {\em The IEEE Conference on Computer Vision and Pattern
  Recognition (CVPR)}, pages 3155--3164, June 2018.

\bibitem{chen2016deep}
Xinyuan Chen, Li Song, and Xiaokang Yang.
\newblock Deep rnns for video denoising.
\newblock In {\em Applications of Digital Image Processing}, 2016.

\bibitem{claus2019videnn}
Michele Claus and Jan van Gemert.
\newblock Videnn: Deep blind video denoising.
\newblock In {\em The IEEE Conference on Computer Vision and Pattern
  Recognition Workshop (CVPRW)}, June 2019.

\bibitem{coupe2009nonlocal}
Pierrick Coup{\'e}, Pierre Hellier, Charles Kervrann, and Christian Barillot.
\newblock Nonlocal means-based speckle filtering for ultrasound images.
\newblock {\em IEEE Transactions on Image Processing}, 18(10):2221--2229, 2009.

\bibitem{Dabov2007v}
Kostadin Dabov, Alessandro Foi, and Karen Egiazarian.
\newblock {Video denoising by sparse 3D transform-domain collaborative
  filtering}.
\newblock In {\em EUSIPCO}, 2007.

\bibitem{vnlnet_icip2019}
Axel Davy, Thibaud Ehret, Jean-Michel Morel, Pablo Arias, and Gabriele
  Facciolo.
\newblock {A Non-Local CNN for Video Denoising}.
\newblock In {\em The IEEE International Conference on Image Processing
  (ICIP)}, volume 2019-Septe, pages 2409--2413. IEEE, sep 2019.

\bibitem{vnlnet_jmiv2020}
Axel Davy, Thibaud Ehret, Jean-Michel Morel, Pablo Arias, and Gabriele
  Facciolo.
\newblock {Video Denoising by Combining Patch Search and CNNs}.
\newblock {\em Journal of Mathematical Imaging and Vision}, pages 1--16, oct
  2020.

\bibitem{deudon2020highresnet}
Michel Deudon, Alfredo Kalaitzis, Md~Rifat Arefin, Israel Goytom, Zhichao Lin,
  Kris Sankaran, Vincent Michalski, Samira~E Kahou, Julien Cornebise, and
  Yoshua Bengio.
\newblock Highres-net: Multi-frame super-resolution by recursive fusion.
\newblock Technical report, 2020.

\bibitem{ehret2019join}
Thibaud Ehret, Axel Davy, Pablo Arias, and Gabriele Facciolo.
\newblock Joint demosaicing and denoising by overfitting of bursts of raw
  images.
\newblock In {\em The IEEE International Conference on Computer Vision (ICCV)},
  2019.

\bibitem{ehret2019model}
Thibaud Ehret, Axel Davy, Jean-Michel Morel, Gabriele Facciolo, and Pablo
  Arias.
\newblock Model-blind video denoising via frame-to-frame training.
\newblock In {\em The IEEE Conference on Computer Vision and Pattern
  Recognition (CVPR)}, June 2019.

\bibitem{18-gonzales-denoising-decompression}
Mario Gonzalez, Javier Preciozzi, Pablo Muse, and Andres Almansa.
\newblock Joint denoising and decompression using cnn regularization.
\newblock In {\em The IEEE Conference on Computer Vision and Pattern
  Recognition Workshops (CVPRW)}, June 2018.

\bibitem{guo2018toward}
Shi Guo, Zifei Yan, Kai Zhang, Wangmeng Zuo, and Lei Zhang.
\newblock Toward convolutional blind denoising of real photographs.
\newblock {\em arXiv preprint arXiv:1807.04686}, 2018.

\bibitem{flir2018adas}
FLIR~Systems Inc.
\newblock Flir thermal dataset for algorithm training.
\newblock {\em Online, \url{https://www.flir.in/oem/adas/adas-dataset-form/}},
  2018.

\bibitem{jin2020review}
Qiyu Jin, Gabriele Facciolo, and Jean-Michel Morel.
\newblock A review of an old dilemma: Demosaicking first, or denoising first?
\newblock In {\em The IEEE Conference on Computer Vision and Pattern
  Recognition Workshops (CVPRW)}, pages 514--515, June 2020.

\bibitem{kang2017deep}
Eunhee Kang, Junhong Min, and Jong~Chul Ye.
\newblock A deep convolutional neural network using directional wavelets for
  low-dose x-ray ct reconstruction.
\newblock {\em Medical physics}, 44(10):e360--e375, 2017.

\bibitem{kay1993fundamentals}
Steven~M Kay.
\newblock Fundamentals of statistical processing, volume i: Estimation theory:
  Estimation theory v. 1, 1993.

\bibitem{kiku2014mlri-demosaicking}
Daisuke Kiku, Yusuke Monno, Masayuki Tanaka, and Masatoshi Okutomi.
\newblock {Minimized-Laplacian residual interpolation for color image
  demosaicking}.
\newblock In Nitin Sampat, Radka Tezaur, Sebastiano Battiato, and Boyd~A.
  Fowler, editors, {\em Digital Photography X}, volume 9023, pages 197 -- 204.
  International Society for Optics and Photonics, SPIE, 2014.

\bibitem{kim2019vid3oc}
Sohyeong Kim, Guanju Li, Dario Fuoli, Martin Danelljan, Zhiwu Huang, Shuhang
  Gu, and Radu Timofte.
\newblock The vid3oc and intvid datasets for video super resolution and quality
  mapping.
\newblock In {\em The International Conference on Computer Vision Workshop
  (ICCVW)}, pages 3609--3616. IEEE, 2019.

\bibitem{kim2020-transfer-synth-to-real-noise}
Yoonsik Kim, Jae~Woong Soh, Gu~Yong Park, and Nam~Ik Cho.
\newblock {Transfer Learning From Synthetic to Real-Noise Denoising With
  Adaptive Instance Normalization}.
\newblock In {\em The IEEE Conference on Computer Vision and Pattern
  Recognition (CVPR)}, pages 3479--3489. IEEE, jun 2020.

\bibitem{noise2void}
Alexander Krull, Tim-Oliver Buchholz, and Florian Jug.
\newblock {Noise2Void - Learning Denoising From Single Noisy Images}.
\newblock In {\em The IEEE Conference on Computer Vision and Pattern
  Recognition (CVPR)}, pages 2124--2132. IEEE, jun 2019.

\bibitem{krull2019probabilisticN2V}
Alexander Krull, Tomas Vicar, and Florian Jug.
\newblock Probabilistic noise2void: Unsupervised content-aware denoising.
\newblock Technical report, 2019.

\bibitem{BSS}
Samuli Laine, Tero Karras, Jaakko Lehtinen, and Timo Aila.
\newblock High-quality self-supervised deep image denoising.
\newblock In {\em Advances in Neural Information Processing Systems 32 (NeurIPS
  2019)}. 2019.

\bibitem{landman2009spatially-variable-mri}
Bennett~A Landman, Pierre-Louis Bazin, Seth~A Smith, and Jerry~L Prince.
\newblock Robust estimation of spatially variable noise fields.
\newblock {\em Magnetic Resonance in Medicine: An Official Journal of the
  International Society for Magnetic Resonance in Medicine}, 62(2):500--509,
  2009.

\bibitem{lebrun2015noise}
Marc Lebrun, Miguel Colom, and Jean-Michel Morel.
\newblock The noise clinic: a blind image denoising algorithm.
\newblock {\em Image Processing On Line}, 5:1--54, 2015.

\bibitem{Lehtinen2018}
Jaakko Lehtinen, Jacob Munkberg, Jon Hasselgren, Samuli Laine, Tero Karras,
  Miika Aittala, and Timo Aila.
\newblock {N}oise2{N}oise: Learning image restoration without clean data.
\newblock In {\em Proceedings of the 35th International Conference on Machine
  Learning}, volume~80 of {\em Journal of Machine Learning Research}, pages
  2965--2974. PMLR, 10--15 Jul 2018.

\bibitem{Liu2010}
Ce Liu and William~T. Freeman.
\newblock {A high-quality video denoising algorithm based on reliable motion
  estimation}.
\newblock In {\em European Conference on Computer Vision (ECCV)}, pages
  706--719, 2010.

\bibitem{Liu2018-nl-rnns}
Ding Liu, Bihan Wen, Yuchen Fan, Chen~Change Loy, and Thomas~S Huang.
\newblock Non-local recurrent network for image restoration.
\newblock In {\em Advances in Neural Information Processing Systems (NIPS)},
  2018.

\bibitem{maggioni2014joint}
Matteo Maggioni, Enrique S{\'a}nchez-Monge, and Alessandro Foi.
\newblock Joint removal of random and fixed-pattern noise through
  spatiotemporal video filtering.
\newblock {\em IEEE Transactions on Image Processing}, 23(10):4282--4296, 2014.

\bibitem{marinc2019mkpn-bursts}
Talmaj Marin{\v{c}}, Vignesh Srinivasan, Serhan G{\"u}l, Cornelius Hellge, and
  Wojciech Samek.
\newblock Multi-kernel prediction networks for denoising of burst images.
\newblock In {\em The IEEE International Conference on Image Processing
  (ICIP)}, pages 2404--2408, September 2019.

\bibitem{metzler2018sure}
Christopher~A. Metzler, Ali Mousavi, Reinhard Heckel, and Richard~G. Baraniuk.
\newblock Unsupervised learning with stein's unbiased risk estimator, 2018.

\bibitem{mildenhall2018kpn-bursts}
Ben Mildenhall, Jonathan~T Barron, Jiawen Chen, Dillon Sharlet, Ren Ng, and
  Robert Carroll.
\newblock Burst denoising with kernel prediction networks.
\newblock In {\em The IEEE Conference on Computer Vision and Pattern
  Recognition (CVPR)}, June 2018.

\bibitem{montgomeryxiph}
Chris Montgomery et~al.
\newblock Xiph. org video test media (derf's collection), the xiph open source
  community, 1994.
\newblock {\em Online, \url{https://media. xiph. org/video/derf}}.

\bibitem{Moran2020noisier2noise}
Nick Moran, Dan Schmidt, Yu Zhong, and Patrick Coady.
\newblock {Noisier2Noise: Learning to Denoise from Unpaired Noisy Data}.
\newblock In {\em The IEEE Conference on Computer Vision and Pattern
  Recognition (CVPR)}, pages 12064--12072, June 2020.

\bibitem{Nam_2016_CVPR}
Seonghyeon Nam, Youngbae Hwang, Yasuyuki Matsushita, and Seon Joo~Kim.
\newblock A holistic approach to cross-channel image noise modeling and its
  application to image denoising.
\newblock In {\em The IEEE Conference on Computer Vision and Pattern
  Recognition (CVPR)}, June 2016.

\bibitem{plotz2017benchmarking}
Tobias Plotz and Stefan Roth.
\newblock Benchmarking denoising algorithms with real photographs.
\newblock In {\em The IEEE Conference on Computer Vision and Pattern
  Recognition (CVPR)}, pages 1586--1595, July 2017.

\bibitem{Plotz2018-NNN}
Tobias Pl\"{o}tz and Stefan Roth.
\newblock Neural nearest neighbors networks.
\newblock In {\em Advances in Neural Information Processing Systems (NIPS)},
  2018.

\bibitem{davis}
Jordi Pont-Tuset, Federico Perazzi, Sergi Caelles, Pablo Arbel\'aez, Alexander
  Sorkine-Hornung, and Luc {Van Gool}.
\newblock The 2017 davis challenge on video object segmentation.
\newblock {\em arXiv:1704.00675}, 2017.

\bibitem{Quan2020self2self}
Yuhui Quan, Mingqin Chen, Tongyao Pang, and Hui Ji.
\newblock {Self2Self With Dropout: Learning Self-Supervised Denoising From
  Single Image}.
\newblock In {\em The IEEE Conference on Computer Vision and Pattern
  Recognition (CVPR)}, pages 1890--1898, June 2020.

\bibitem{Ronneberger2015U-Net:Segmentation}
Olaf Ronneberger, Philipp Fischer, and Thomas Brox.
\newblock {U-Net: Convolutional Networks for Biomedical Image Segmentation}.
\newblock {\em Miccai}, pages 234--241, 2015.

\bibitem{rudin1992nonlinear}
Leonid~I Rudin, Stanley Osher, and Emad Fatemi.
\newblock Nonlinear total variation based noise removal algorithms.
\newblock {\em Physica D: nonlinear phenomena}, 60(1-4):259--268, 1992.

\bibitem{salmon2014poisson}
Joseph Salmon, Zachary Harmany, Charles-Alban Deledalle, and Rebecca Willett.
\newblock Poisson noise reduction with non-local pca.
\newblock {\em Journal of mathematical imaging and vision}, 48(2):279--294,
  2014.

\bibitem{sanchez2013tv}
Javier S{\'a}nchez~P{\'e}rez, Enric Meinhardt-Llopis, and Gabriele Facciolo.
\newblock Tv-l1 optical flow estimation.
\newblock {\em Image Processing On Line}, 2013:137--150, 2013.

\bibitem{16-santhanam-rbdn}
Venkataraman Santhanam, Vlad~I. Morariu, and Larry~S. Davis.
\newblock Generalized deep image to image regression.
\newblock {\em CoRR}, abs/1612.03268, 2016.

\bibitem{soltanayev2018sure}
Shakarim Soltanayev and Se~Young Chun.
\newblock Training deep learning based denoisers without ground truth data.
\newblock In S. Bengio, H. Wallach, H. Larochelle, K. Grauman, N. Cesa-Bianchi,
  and R. Garnett, editors, {\em Advances in Neural Information Processing
  Systems 31}, pages 3257--3267. Curran Associates, Inc., 2018.

\bibitem{stein1981}
Charles~M. Stein.
\newblock Estimation of the mean of a multivariate normal distribution.
\newblock {\em Ann. Statist.}, 9(6):1135--1151, 11 1981.

\bibitem{tassano2019dvdnet}
Matias Tassano, Julie Delon, and Thomas Veit.
\newblock Dvdnet: A fast network for deep video denoising.
\newblock In {\em The IEEE International Conference on Image Processing
  (ICIP)}, September 2019.

\bibitem{tassano2019fastdvdnet}
Matias Tassano, Julie Delon, and Thomas Veit.
\newblock Fastdvdnet: Towards real-time deep video denoising without flow
  estimation.
\newblock In {\em The IEEE International Conference on Computer Vision and
  Pattern Recognition (CVPR)}, pages 1354--1363, June 2020.

\bibitem{tonioni2019-learnint-to-adapt}
Alessio Tonioni, Oscar Rahnama, Tom Joy, Luigi Di~Stefano, Ajanthan
  Thalaiyasingam, and Philip Torr.
\newblock Learning to adapt for stereo.
\newblock In {\em The IEEE Conference on Computer Vision and Pattern
  Recognition (CVPR)}, June 2019.

\bibitem{stacked-denoising-autoencoders-2010}
Pascal Vincent, Hugo Larochelle, Isabelle Lajoie, Yoshua Bengio, and
  Pierre-Antoine Manzagol.
\newblock Stacked denoising autoencoders: Learning useful representations in a
  deep network with a local denoising criterion.
\newblock {\em J. Mach. Learn. Res.}, 11:3371--3408, Dec. 2010.

\bibitem{wang2017sar}
Puyang Wang, He Zhang, and Vishal~M Patel.
\newblock Sar image despeckling using a convolutional neural network.
\newblock {\em IEEE Signal Processing Letters}, 24(12):1763--1767, 2017.

\bibitem{xue2019}
Tianfan Xue, Baian Chen, Jiajun Wu, Donglai Wei, and William~T. Freeman.
\newblock Video enhancement with task-oriented flow.
\newblock {\em International Journal of Computer Vision (IJCV)},
  127(8):1106--1125, Aug 2019.

\bibitem{yu2020joint}
Songhyun Yu, Bumjun Park, Junwoo Park, and Jechang Jeong.
\newblock Joint learning of blind video denoising and optical flow estimation.
\newblock In {\em The IEEE Conference on Computer Vision and Pattern
  Recognition Workshops (CVPRW)}, pages 500--501, June 2020.

\bibitem{yue2020supervised}
Huanjing Yue, Cong Cao, Lei Liao, Ronghe Chu, and Jingyu Yang.
\newblock Supervised raw video denoising with a benchmark dataset on dynamic
  scenes.
\newblock In {\em The IEEE Conference on Computer Vision and Pattern
  Recognition (CVPR)}, pages 2301--2310, June 2020.

\bibitem{zach2007duality}
Christopher Zach, Thomas Pock, and Horst Bischof.
\newblock A duality based approach for realtime tv-l 1 optical flow.
\newblock In {\em Joint Pattern Recognition Symposium}. Springer, 2007.

\bibitem{zamir2020cycleisp}
Syed~Waqas Zamir, Aditya Arora, Salman Khan, Munawar Hayat, Fahad~Shahbaz Khan,
  Ming-Hsuan Yang, and Ling Shao.
\newblock Cycleisp: Real image restoration via improved data synthesis.
\newblock In {\em The IEEE Conference on Computer Vision and Pattern
  Recognition (CVPR)}, pages 2696--2705, June 2020.

\bibitem{Zhang2017BeyondDenoising}
Kai Zhang, Wangmeng Zuo, Yunjin Chen, Deyu Meng, and Lei Zhang.
\newblock {Beyond a Gaussian Denoiser: Residual Learning of Deep CNN for Image
  Denoising}.
\newblock {\em IEEE Transactions on Image Processing}, 26(7):3142--3155, 7
  2017.

\bibitem{zhao2019ratio}
Weiying Zhao, Charles-Alban Deledalle, Lo{\"\i}c Denis, Henri Ma{\^\i}tre,
  Jean-Marie Nicolas, and Florence Tupin.
\newblock Ratio-based multitemporal sar images denoising: Rabasar.
\newblock {\em IEEE Transactions on Geoscience and Remote Sensing},
  57(6):3552--3565, 2019.

\end{thebibliography}

}

\end{document}


\title{Self-supervised training for blind multi-frame video denoising\\
Supplementary material}

\maketitle

\appendix

This document contains supplementary material for our paper ``Self-supervised training for blind multi-frame video denoising''. It is not intended to be self-contained. It follows the notation introduced in the main paper.

\section{{Pseudocodes}}\label{sec:pseudocodes}

The pseudocodes of the online and offline versions of our method are shown in Algorithms \ref{alg:online} and \ref{alg:offline}. In both cases, the input video is made of frames $\{f_t\}_{t \in \{1, \dots, T\}}$. For the online method, the weights are updated by $N$ iterations of Adam optimization. 

The offline method uses mini-batches of $N_b$ frames. The weights are updated for $N_A$ steps of Adam optimizer.

{The source code of the proposed method will be made available.}

\begin{algorithm*}
\SetInd{.5em}{2em}
\Input{Noisy video $\{f_t\}_{t \in \{1, \dots, T\}}$, initial weights $\theta_0$, number of Adam step $N^S$}
\Output{Denoised video $\hat u$}
\DontPrintSemicolon

\For{$t = 2,\ldots, T$ } {
$\ma v_{t-1,t} \gets  \texttt{optical-flow}(f_{t-1},f_t)$\;
$W_{t,t-1} \gets  \texttt{warping-operator}(\ma v_{t-1,t})$\;
$\kappa_t \gets  \texttt{alignment-error-mask}(\ma v_{t-1,t}, W_{t,t-1}f_t, f_{t-1})$\;
$\mathcal S'_{t} \gets [f_{t-4}, f_{t-2}, f_{t}, f_{t+2}, f_{t+4}] $ // \textit{training input stack} \\
$\mathcal S_{t} \gets [f_{t-2}, f_{t-1}, f_{t}, f_{t+1}, f_{t+2}] $ // \textit{inference input stack}

// \textit{update the network} \\
\For{$i = 1,\ldots, N^{\text{S}}$ } {
    $\theta_t \gets \texttt{adam-step}(\ell^{\text{MF2F}}_1( \mathcal F_{\theta}(\mathcal S'_{t}), f_{t-1}, W_{t,t-1}, \kappa_t))$
}

// \textit{denoise the frame $t$} \\
$\hat u_t  \gets \mathcal F_{\theta_t}(\mathcal S_t)$
}
\caption{Online fine-tuning}
\label{alg:online}
\end{algorithm*}

\begin{algorithm*}
\SetInd{.5em}{2em}
\Input{Noisy video $f$, initial weights $\theta_0$, number of Adam updates $N_A$, mini-batch size $N_b$}
\Output{Denoised video $\hat u$}
\DontPrintSemicolon
\For{$i = 1,\ldots, N_A$ } {
$loss \gets 0$\\
\For{$j = 1, \ldots, N_b$}{
$t  \gets \texttt{randint}(1, T)$ // \textit{choose a random frame}\;
$\ma v_{t-1,t} \gets  \texttt{optical-flow}(f_{t-1},f_t)$\;
$W_{t,t-1} \gets  \texttt{warping-operator}(\ma v_{t-1,t})$\;
$\kappa_t \gets  \texttt{alignment-error-mask}(\ma v_{t-1,t}, W_{t,t-1}f_t, f_{t-1})$\;
$\mathcal S'_{t} \gets [f_{t-4}, f_{t-2}, f_{t}, f_{t+2}, f_{t+4}] $ // \textit{ training input stack}\;
$\mathcal S_{t} \gets [f_{t-2}, f_{t-1}, f_{t}, f_{t+1}, f_{t+2}] $ // \textit{inference input stack}\;
// \textit{accumulate gradients}\;
$loss \gets loss + (\ell^{\text{MF2F}}_{1}( \mathcal F_{\theta}(\mathcal S'_{t}), f_{t-1}, W_{t,t-1}, \kappa_t))$
}
// \textit{update student network with Adam step}\;
 $\theta_t \gets \texttt{adam-step}($loss$)$ 
}

// \textit{Process the denoising of the entire video}\;
\For{$t = 1,\ldots, T$ } {
$\hat u_t  \gets \mathcal F_{\theta_t}(\mathcal S_t)$ // \textit{denoise the frame $t$}
}

\caption{Offline fine-tuning}
\label{alg:offline}
\end{algorithm*}

\section{Convergence of the offline fine-tuning}

In the offline fine-tuning we estimate the gradient using mini-batches corresponding to 20 frames randomly sampled throughout the video. For each sampled frame the denoised frame is computed using the corresponding training stack. The weights are updated using the Adam update. This is repeated for $N_A = 200$ iterations.

In \citefig{fig:plot_psnr_average_offline} we show the evolution of the average PSNR over the complete sequence with respect to the number of weight updates. The PSNR grows fast during the first 100 iterations. After that it continues to grow at a slower rate or plateaus. Based on this evolution, we set $N_A = 200$ iterations which is a reasonable trade-off between fine-tuning time and denoising performance.


\begin{figure*}
    \centering
    
        \subfloat[Gaussian $\sigma = 20$]{    \includegraphics[width=0.49\textwidth]{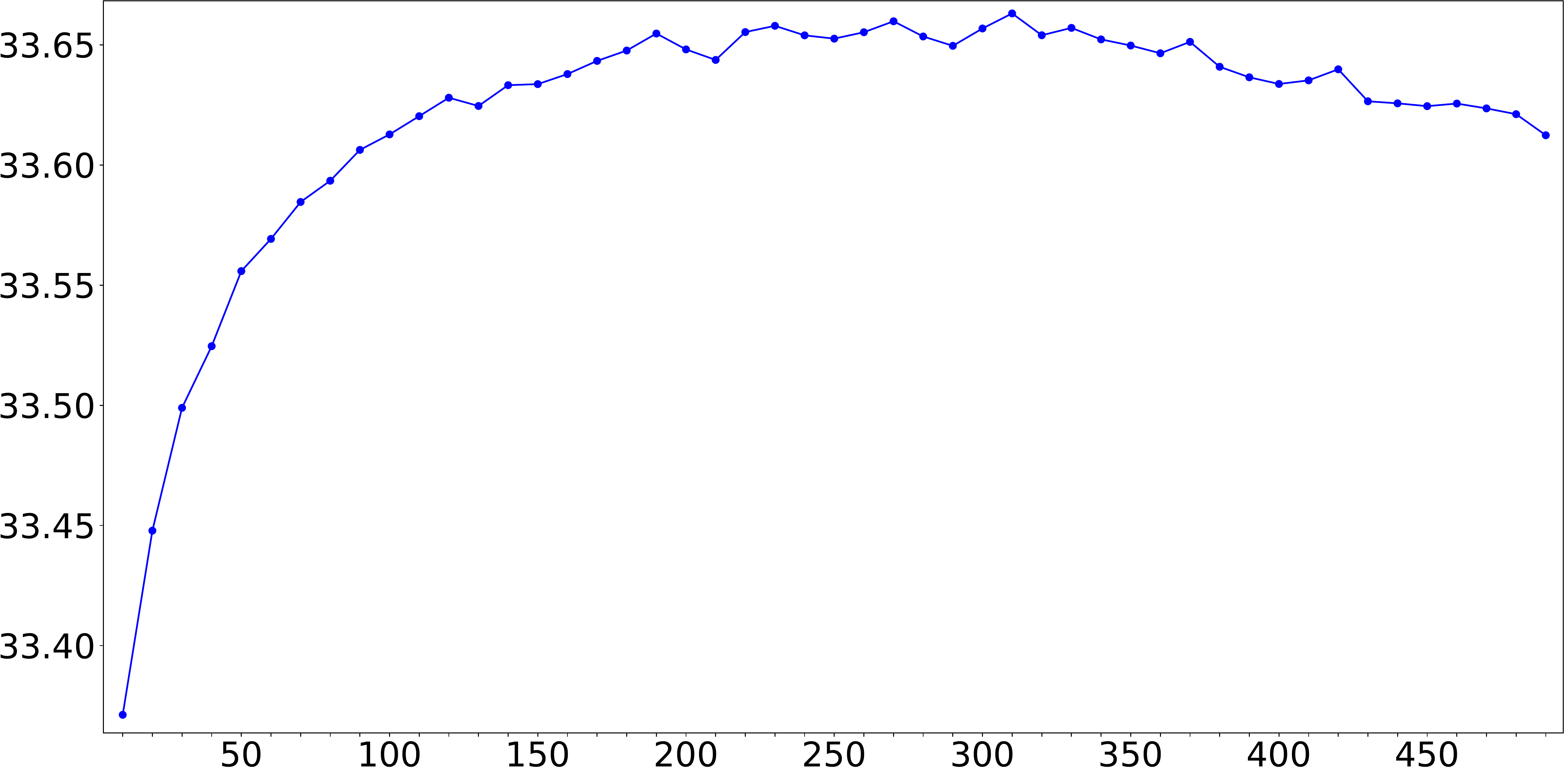}}
        \subfloat[Gaussian $\sigma = 40$]{    \includegraphics[width=0.49\textwidth]{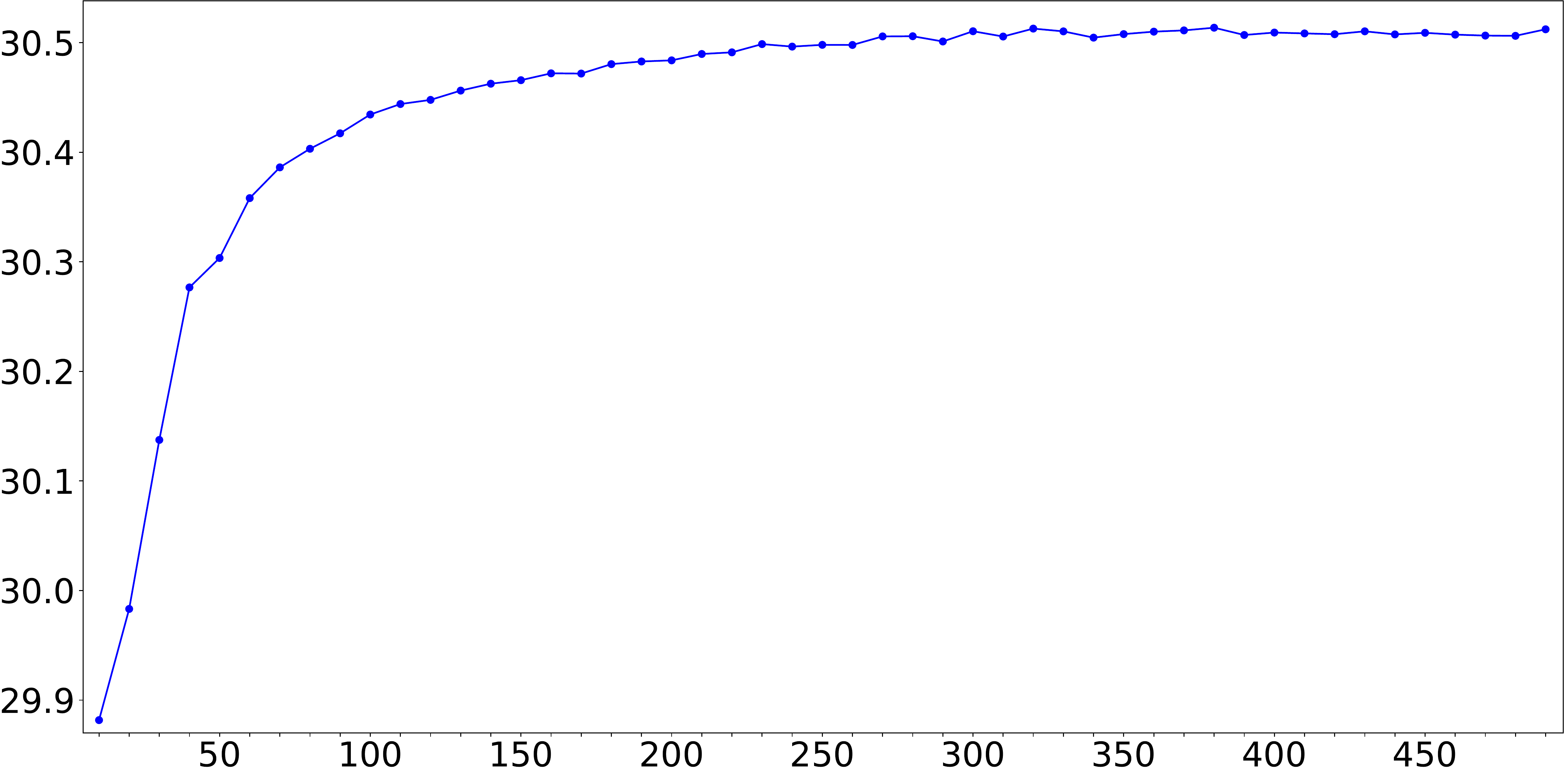}}
        
        \subfloat[Poisson $p = 1$]{    \includegraphics[width=0.49\textwidth]{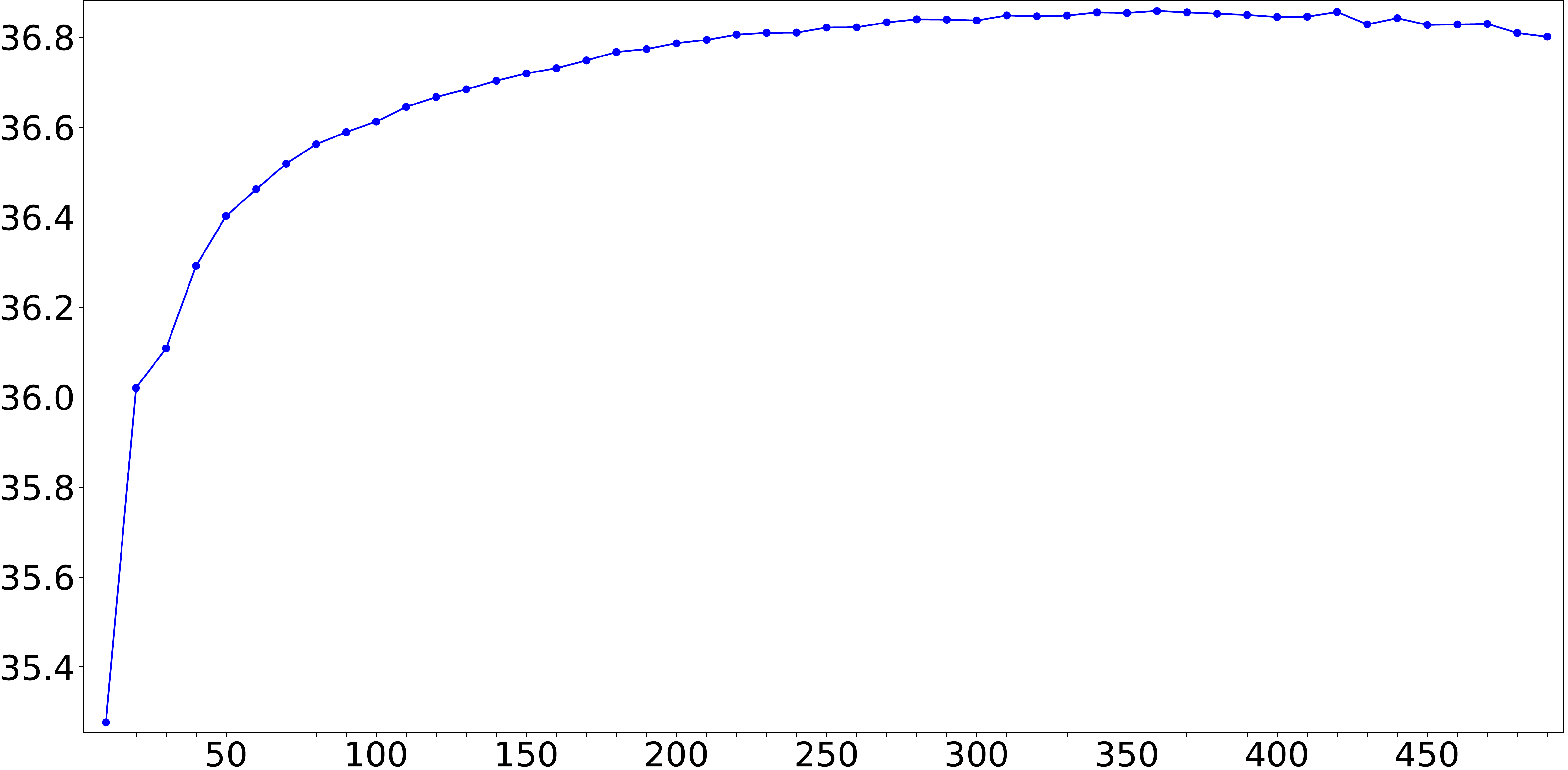}}
        \subfloat[Poisson $p = 8$]{    \includegraphics[width=0.49\textwidth]{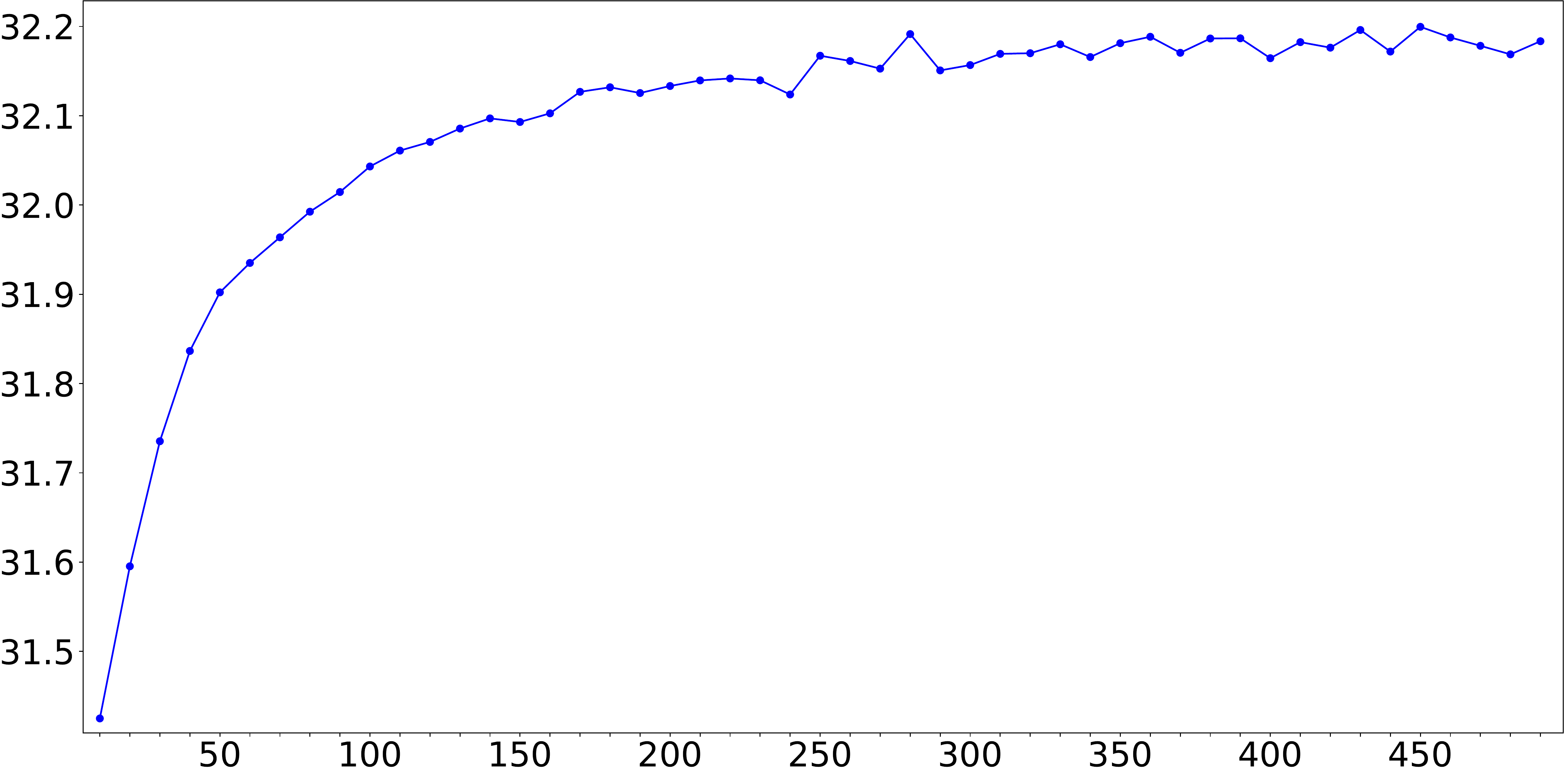}}
        
    \subfloat[Box noise $3 \times 3$]{    \includegraphics[width=0.49\textwidth]{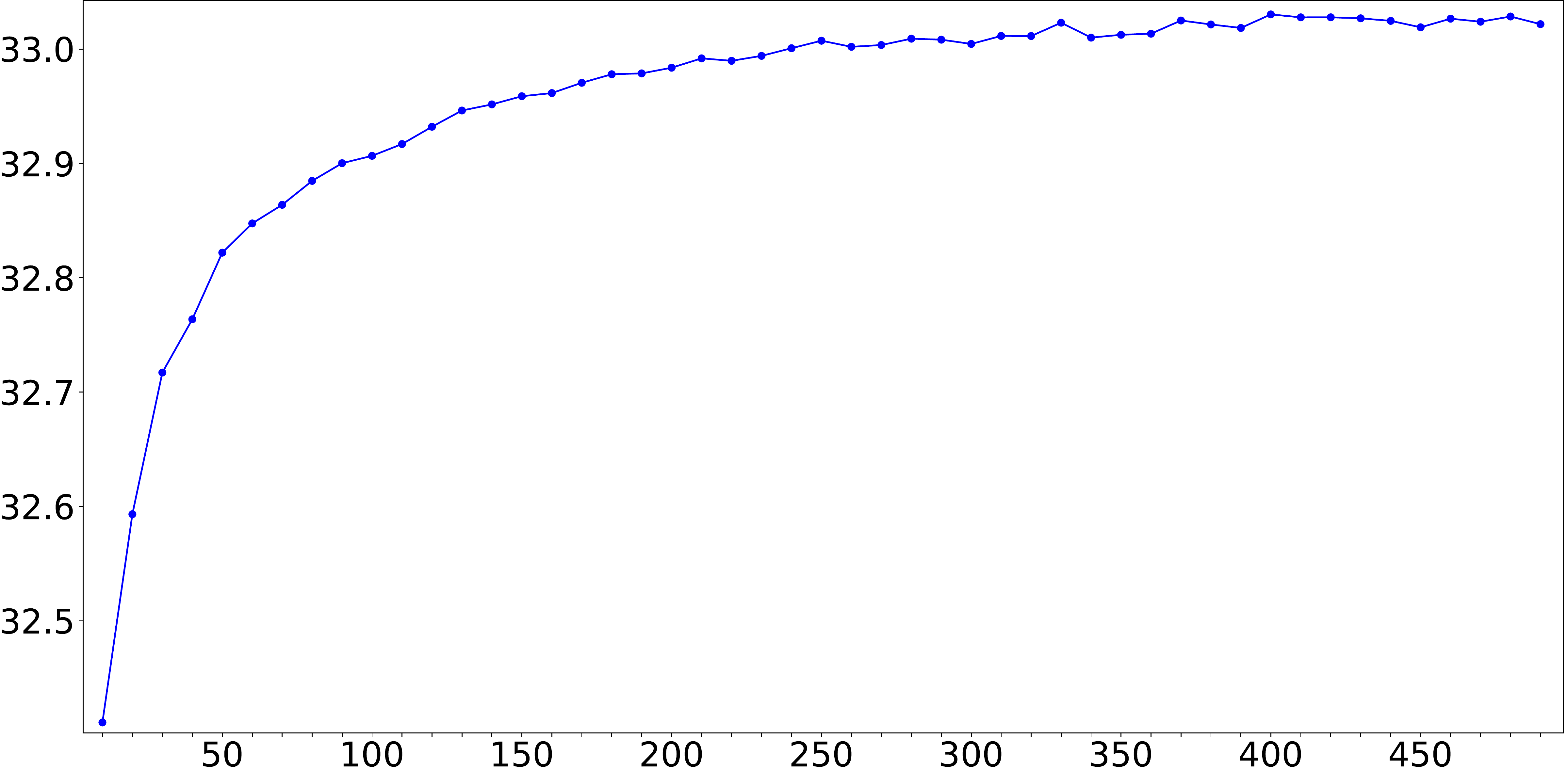}}
        \subfloat[Box noise $5 \times 5$]{    \includegraphics[width=0.49\textwidth]{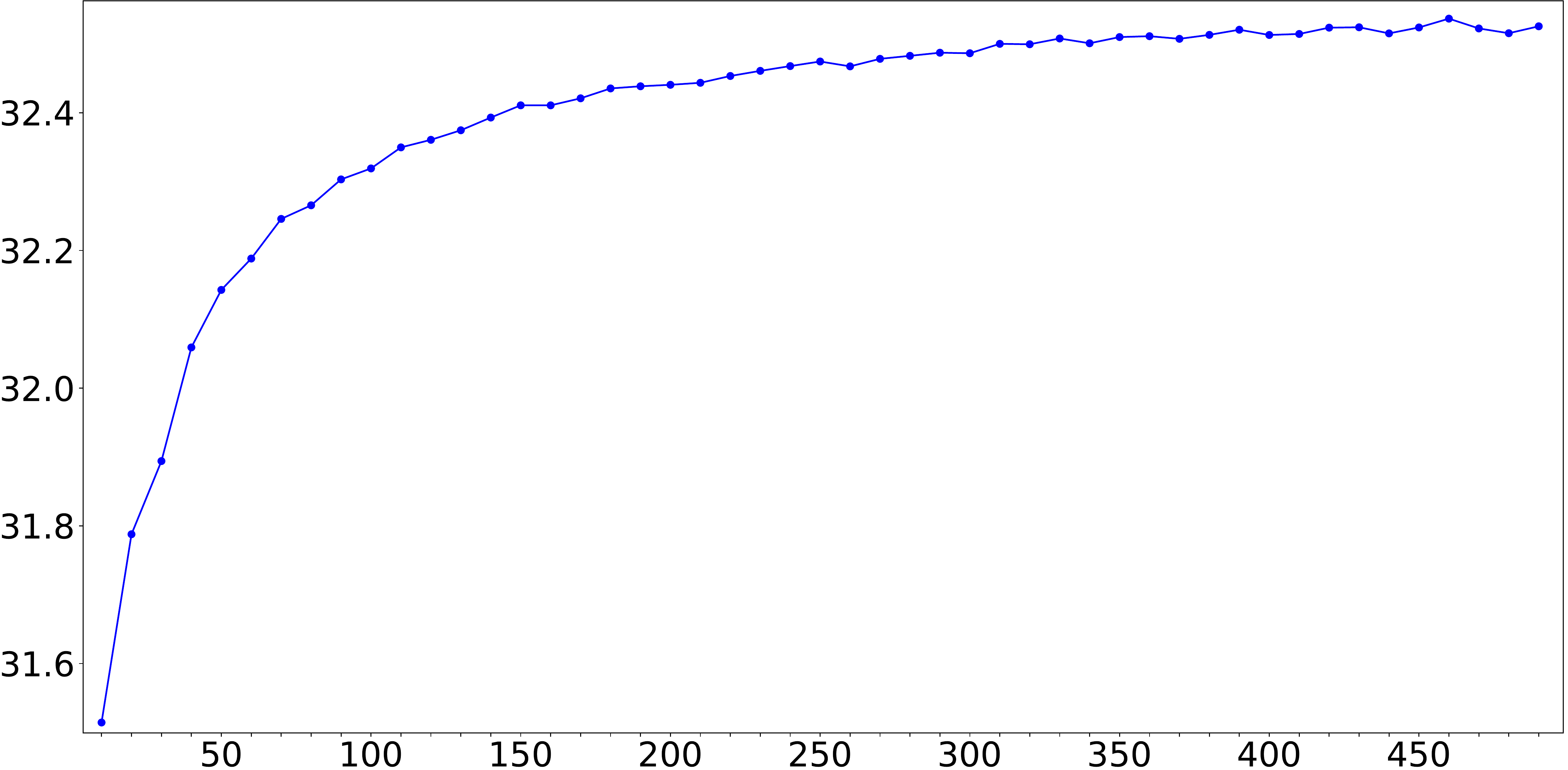}}
        
    	\caption{Justification of the number of iterations in the offline framework: average PSNR on the whole sequence as a function of the number of Adam updates done during the fine-tuning.}
	\label{fig:plot_psnr_average_offline}
\end{figure*}

\section{Computation of the mask}

The proposed MF2F loss penalizes the difference between the network output at $t$ with the previous noisy frame $f_{t-1}$. The network output is aligned to frame $t-1$ by the warping operator $W_{t,t-1}$ which results from the optical flow computed between $t-1$ and $t$. Alignment errors have a negative impact in the training and are removed with a mask $\kappa_t$. This mask is the product (or logical AND) of two binary masks: $\kappa_t(x) = \kappa_t^{\text{OCC}}(x)\kappa_t^{\text{W}}(x)$. The first factor estimates occlusions by looking at collisions in the optical flow, similar in spirit to \cite{ehret2019model}. The second factor is explained below.

The mask $\kappa_t^{\text{W}}$ is zero in areas where the warping residue is larger than a threshold and one elsewhere. For a pixel $x$ at time $t$, the warping residue is computed as
\begin{equation}
    r_{t,t-1}(x) = g_1(x) \ast \vert g_2(x) \ast f_{t-1}(x) - g_2(x)\ast W_{t,t-1}f_t(x) \vert_1,
\end{equation}
where $g_1$ and $g_2$ are two Gaussian convolution kernels and the 1-norm $|\cdot|_1$ means the sum of the errors (in absolute value) for each channel at pixel $x$.
We smooth the images with the kernel $g_2$ to remove some of the noise. Then, this pixelwise norm is smoothed again by the kernel $g_1$. In practice, we used a Gaussian kernel with $\sigma = 2$ for both $g_1$ and $g_2$. 
In order to further reduce noise, we downsample $f_{t-1}$ and $f_t$ by a factor 2. The final warping residue $r_{t,t-1}$ is then upsampled to the original resolution. 

The distribution of the warping residuals can be considered as a mixture of two components. One due to the residuals caused by the noise, and the other due to registration errors. We compute a threshold such that values above that threshold are likely to be registration errors and not just differences caused by the noise. 
Computing such a threshold is difficult without making any assumption on the noise distribution.
We will assume that the distribution of residual caused by the noise is unimodal. We compute the threshold automatically for each frame as
\[\tau_t = m_t + s_t f,\]
where $m_t$ is the mode of the histogram of residuals $r_{t,t-1}$ and $s_t = m_t - p_t$, the difference between the mode and the $10\%$ percentile $p_t$ (thus we are also assuming that the mode is larger than the $10\%$). The mode of the histogram serves as a robust estimation of the position of the distribution, whereas $s$ is a measure of the spread of the distribution. We use the distance between the mode and a low percentile, since we expect warping errors to affect the tail of the distribution (values larger than the mode). The histogram is smoothed by a Gaussian kernel.

\begin{figure*}
    \centering
        \subfloat[Frame $t-1$]{\includegraphics[ width=0.48\textwidth]{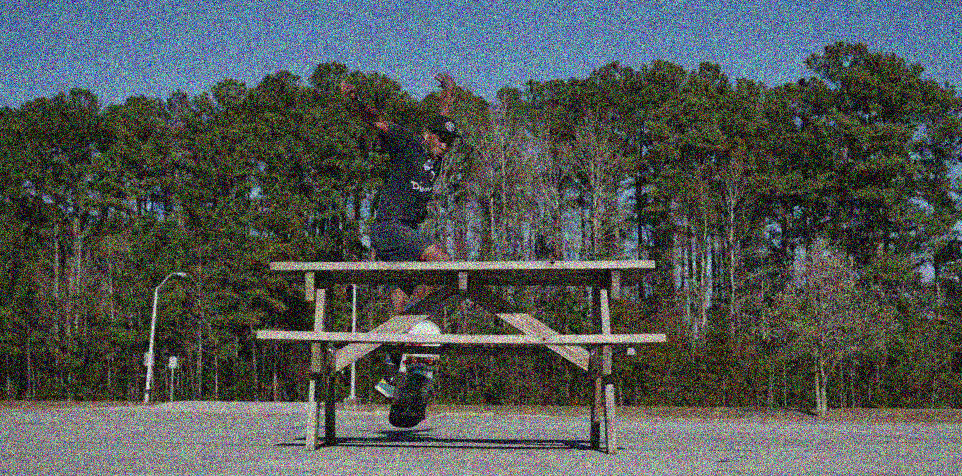}}
        \subfloat[Frame $t$]{\includegraphics[ width=0.48\textwidth]{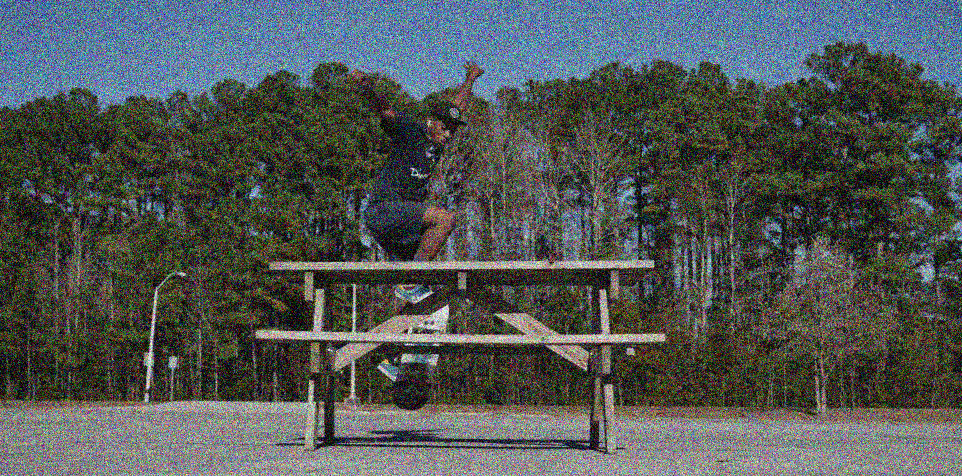}}\\
        \subfloat[Mask $\kappa_t$]{\includegraphics[ width=0.48\textwidth]{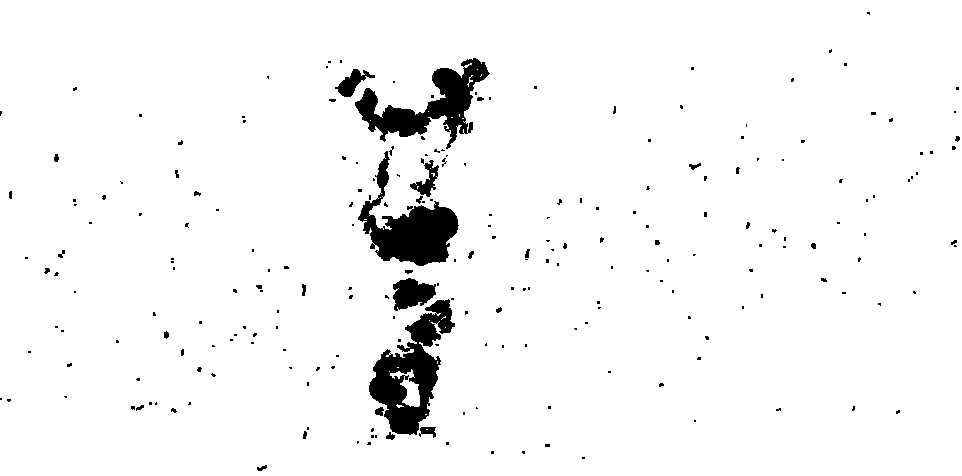}}
        \subfloat[Flow $v_{t-1, t}$]{\includegraphics[clip,trim=0px 0px 0px 0px,width=0.48\textwidth]{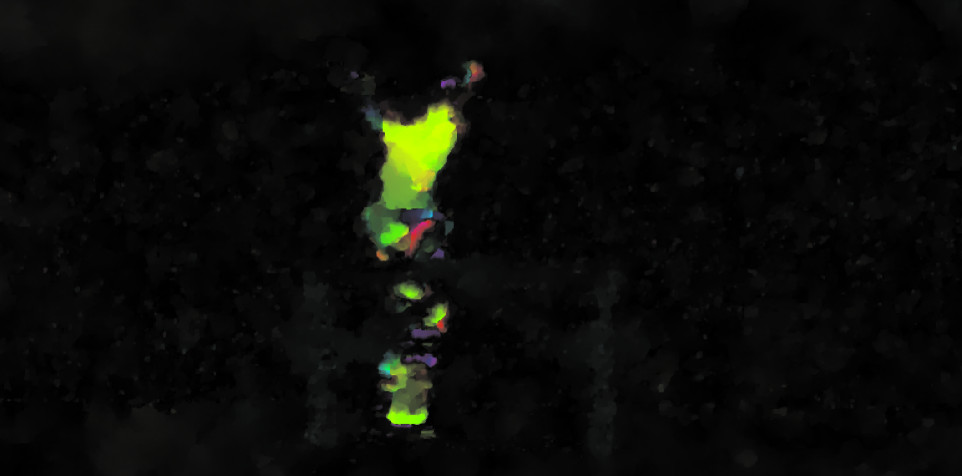}}\\
        \subfloat[Masked target frame $\kappa_t \circ f_{t-1}$ \label{fig:masked_target}]{\includegraphics[ width=0.48\textwidth]{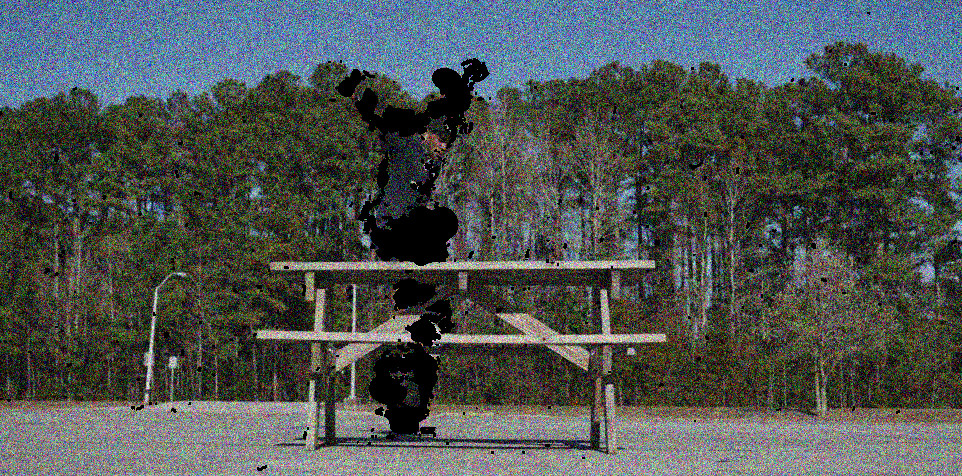}}
        \subfloat[Masked warped frame $\kappa_t \circ W_{t,t-1} f_t$ \label{fig:masked_warped}]{\includegraphics[width=0.48\textwidth]{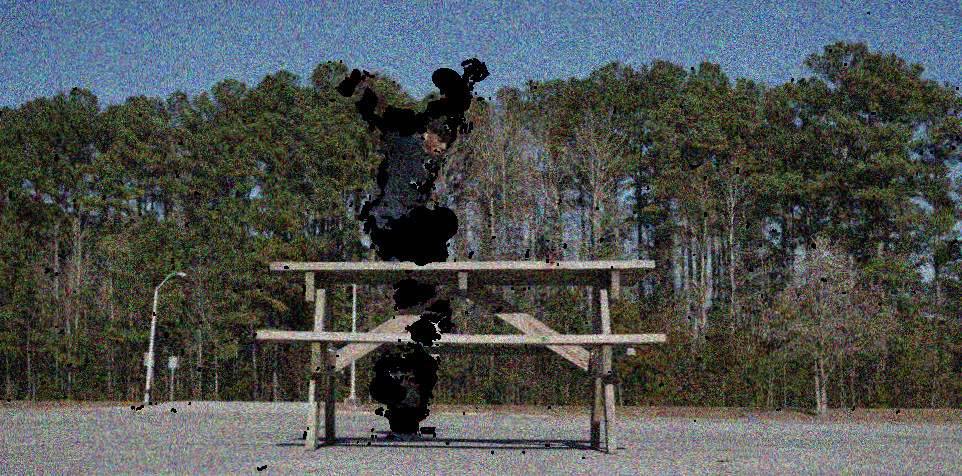}}
    	\caption{Example of mask, flow and warping in case of a very fast motion. The notations are those of section \ref{sec:pseudocodes}.}
	\label{fig:mask,warping}
\end{figure*}

In Figure \ref{fig:mask,warping} we show an example of warping mask computed with this strategy. In this example, the motion is very fast between two consecutive frames $f_{t-1}$ and $f_t$. The arms, the knee of the skater and the skate itself moves quickly. This fast motion is not tracked well by the optical flow and leads to inconsistent warping for those regions. The mask $\kappa_t$ removes these pixels from the loss. Figures~\ref{fig:masked_target} and \ref{fig:masked_warped} illustrate mask overlaid on the target frame  ($f_{t-1}$) and the warped central frame of the stack ($f_t$) .

\section{Experiment on the stack and target position configurations}

\begin{table*}[t]
    \centering
    \begin{tabular}{l| c | c c | c c | c c}
		 & & \multicolumn{2}{c|}{Box $3\times 3$,  40}  & \multicolumn{2}{c|}{Gaussian 20}  & \multicolumn{2}{c}{Poisson 8} \\
		 Training stack $\mathcal S'_t$   & ref. & \multicolumn{2}{c|}{Inference stack} & \multicolumn{2}{c|}{Inference stack} & \multicolumn{2}{c}{Inference stack} \\
		  &  & $\mathcal S_t$ & $\mathcal S'_t$  & $\mathcal S_t$ & $\mathcal S'_t$  & $\mathcal S_t$ & $\mathcal S'_t$   \\\hline
         $f_{t-2}, f_{t-1}, f_{t}, f_{t+1}, f_{t+2}$ & $f_{t-3}$ &   \multicolumn{2}{c|}{28.93} & \multicolumn{2}{c|}{28.78} & \multicolumn{2}{c}{28.13} \\
         $f_{t-3}, f_{t-1}, f_{t}, f_{t+1}, f_{t+2}$ & $f_{t-2}$ &    32.02  & 31.90 & 32.25 & 32.12 & 31.18 & 31.03\\
         $f_{t-3}, f_{t-2}, f_{t}, f_{t+1}, f_{t+2}$ & $f_{t-1}$ & \textbf{36.23}  & 35.98 & 37.25 & 37.10 & \textbf{35.20} & 35.02\\
         $f_{t-4}, f_{t-2}, f_{t}, f_{t+2}, f_{t+4}$ & $f_{t-1}$ & \textbf{36.22} & 35.78 & \textbf{37.32} & 36.94 & \textbf{35.21} & 34.86\\
         \hline
       FastDVDnet superv. & n/a & 36.58  & n/a & 37.29 & n/a & 35.82 & n/a\\
    \end{tabular}
    \caption{PSNR results for different reference frames and training stacks $\mathcal S'_t$. $\mathcal S_t$ denotes the natural input stack.
    This test was carried out on the datasets Derf~\cite{montgomeryxiph} and Vid30C-10~\cite{kim2019vid3oc} using the online MF2F fine-tuning. The reported PSNRs are the average on all the sequences, but excluding the first 10 frames (to avoid perturbations due to the adaptation time).
    }
    \label{tab:stacks}
\end{table*}

The position of the target frame and the choice of the stack have already been discussed in the main paper. The table \ref{tab:stacks} extends the table 1 from the main paper by showing in addition the gain obtained by switching to the natural stack at inference compared with keeping the training one. Note that for the first row, the training stack $\mathcal S'_t = [f_{t-2}, f_{t-1}, f_{t}, f_{t+1}, f_{t+2}]$ is precisely the natural stack ($\mathcal S_t$). Thus both results for $\mathcal S_t$ and $\mathcal S'_t$ are equal. In the same way, for comparison, a row with the noise-specific supervised FastDVDnet was added. FastDVDnet was trained on the natural stack, The evalutation of FastDVDnet on our training stack does not make sense as it absurdly degrades its performance. We omitted them in the table.

\section{Impact of pre-trained network}

The proposed fine-tuning scheme can be applied to any denoising network and any pre-trained weights can be used as a starting point. 
Here we evaluate the impact of the choice of the pre-trained weights.
This issue is related to transfer learning and domain adaptation, where a pre-trained network is re-targeted for a different task or input data distribution. 
In \cite{zamir2018taskonomy} it is shown that effectiveness of the transference depends on the similarity between the source and target tasks.

In the same spirit as \cite{zamir2018taskonomy}, 
we tested our fine-tuning starting from weights pre-trained for four types of noise: AWGN with $\sigma = 15, 25, 35$ and box noise with with kernel size $3 \times 3$  and $\sigma=40$.
For the AWGN noise we used the pre-trained network provided by~\cite{tassano2019fastdvdnet}.
We fine-tuned those pre-trained networks for three different target noises: AWGN with small $\sigma = 10$, a stronger AWGN with $\sigma = 40$ and box noise with kernel size $5 \times 5$ and $\sigma=65$. 

\begin{figure*}
    \centering
	 \begin{subfigure}{.32\linewidth}
			\includegraphics[width=\linewidth]{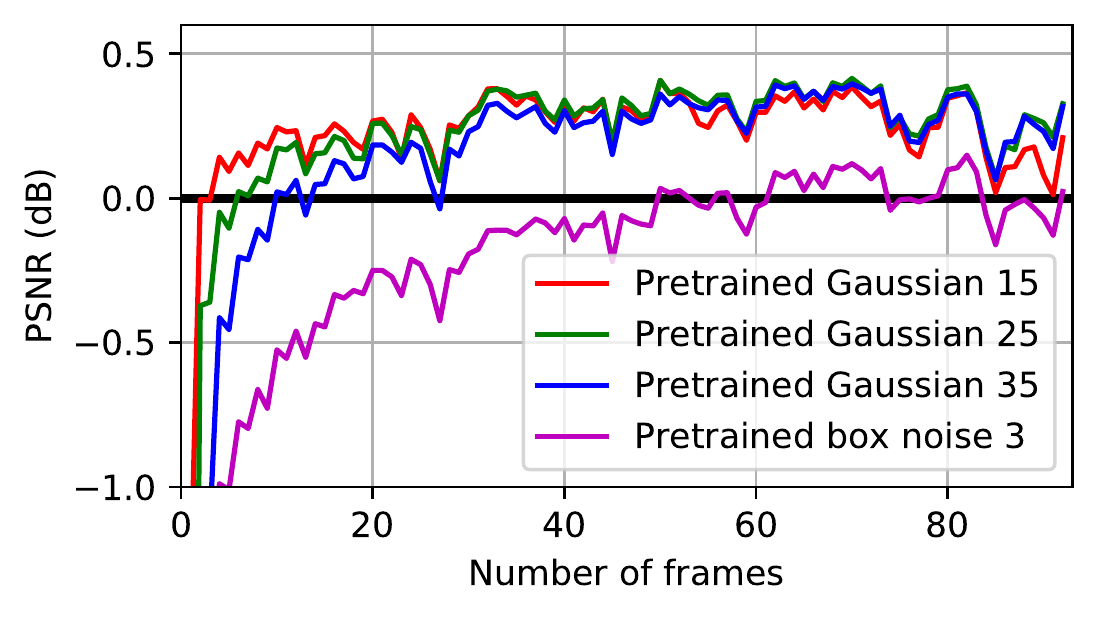}
			\caption{Target noise AWGN with $\sigma=10$}
	 	\end{subfigure}
	 	\begin{subfigure}{.32\linewidth}
			\includegraphics[width=\linewidth]{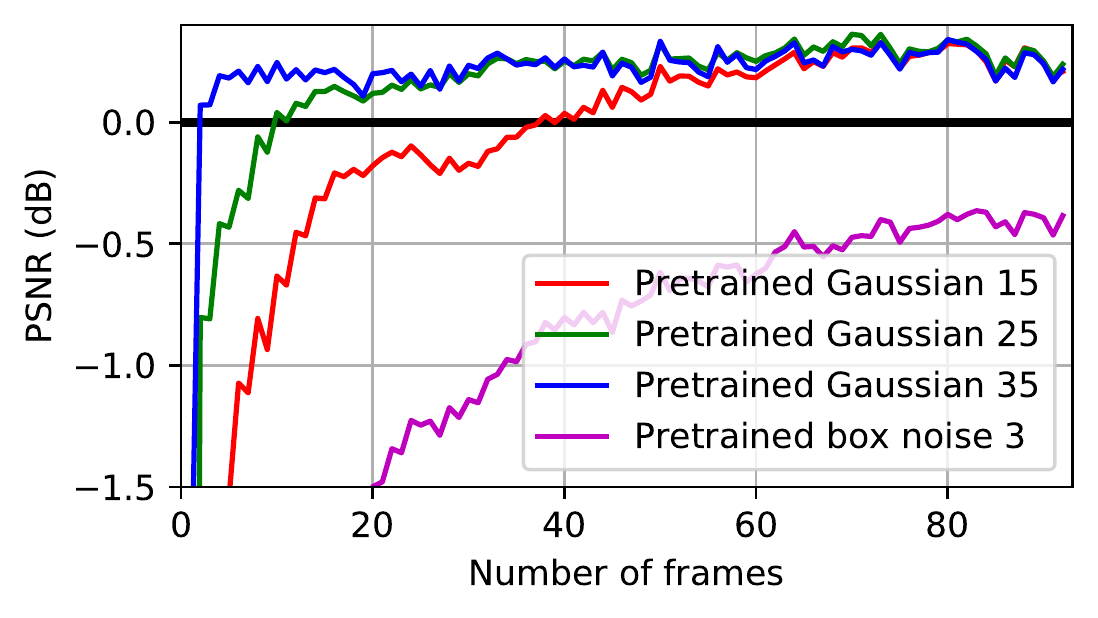}
			\caption{Target noise AWGN with $\sigma=40$}
	 	\end{subfigure}
	 	\begin{subfigure}{.32\linewidth}
         \includegraphics[width=\linewidth]{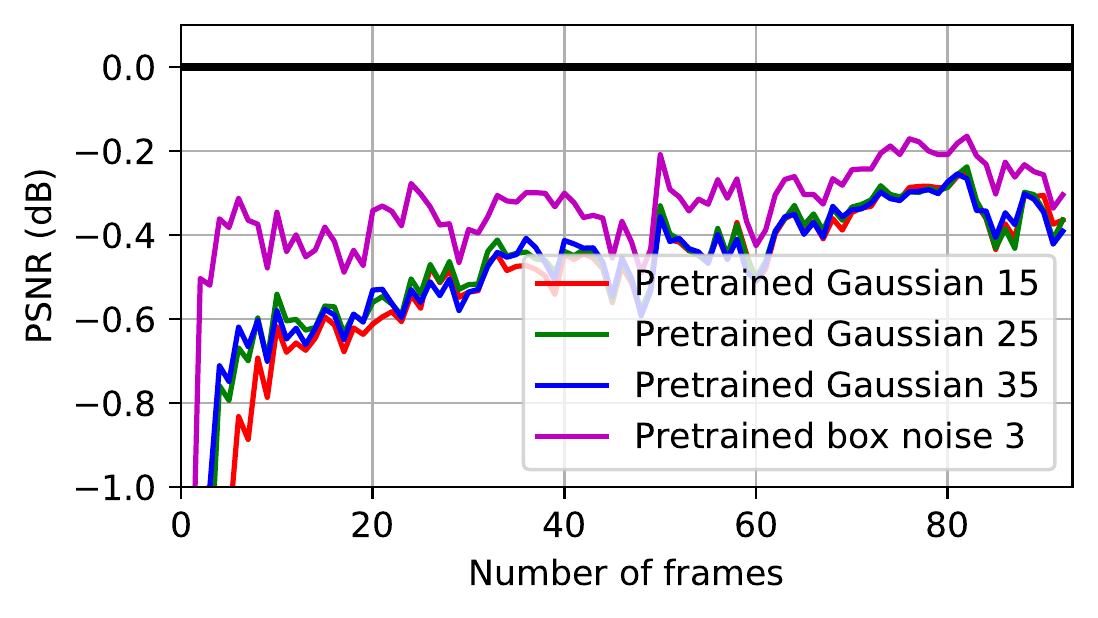}
			\caption{Target noise $5\times 5$ box noise with $\sigma=65$}
	 	\end{subfigure}
        \vspace{-.5em}
    	\caption{Online MF2F fine-tuning starting from different pre-trained weights for different target noise types. For each frame, we plot the difference in PSNR with respect to the result of a noise-specific network trained with supervision. The per-frame PSNRs are averaged over the seven videos of the Derf dataset.}
	\label{fig:exp_choice_pretrained}
\end{figure*}

Fig.~\ref{fig:exp_choice_pretrained} shows three plots, one per target noise. 
We consider the online version of our fine-tuning to evaluate the convergence speed. The fine-tunings were performed independently on  sequences of 100 frames. For each frame, we average the PSNR obtained at that frame for the seven sequences of the Derf dataset. We plot the evolution of the difference between the average per-frame PSNR for our fine-tuned network and a network which was trained with supervision specifically for each target noise type.

As expected, the similarity between the source and target noise distributions impacts the convergence speed of the online fine-tuning.  
For both AWGN noise targets the weights pre-trained for AWGN with the closest $\sigma$ show the fastest convergence. Similarly, the weights pre-trained for box noise work better when the target noise is also box noise. In all cases the fine-tuned network achieves a performance comparable to the supervised network (within a 0.4dB range), and even surpasses it in the case of AWGN. It seems to be easier for the weights pre-trained for AWGN to adapt to the box noise than the other way around. For this reason, all our experiments were done starting our fine-tunings from the weights pre-trained for AWGN with $\sigma = 25$.

\section{Fine-tuning only the variance map}

Figure~\ref{fig:variance_map} shows results obtained for Poisson noise by fine-tuning the variance map. We compare the results obtained for the constant variance map and the per-level variance map (with $K = 8$ levels). Figure~\ref{fig:noisemap_8sigmas} shows an example of the per-level variance map. We recall this variance map is built by first segmenting the image in $K$ regions based on the pixel intensity of the noisy image. To each region we assign a variance $\sigma_i
^2$, and the fine-tuning is applied to these $K = 8$ variances.

The results with constant variance map clearly contain remaining noise and also over-smoothed areas, whereas the results with our variance map are uniformly denoised.

In Fig. \ref{fig:comparison_spatial_variance_map}, we compare some results of the MF2F fine-tuning using the per-level noise map and the spatially variant noise map on Poisson noise. The spatially variant noise map is particularly suitable for Heteroscedastic AWGN. Although in principle, the same strategy as for the space varying noise case could be also used to estimate a time-varying variance map $\Sigma_t(t)$, the results with the per-level noise map are more accurate. This is because the spatially variant noise map requires more iterations per frame (due to the slower convergence) to adapt to a temporally varying noise pattern (leading to a higher computational cost per-frame).

\begin{figure*}
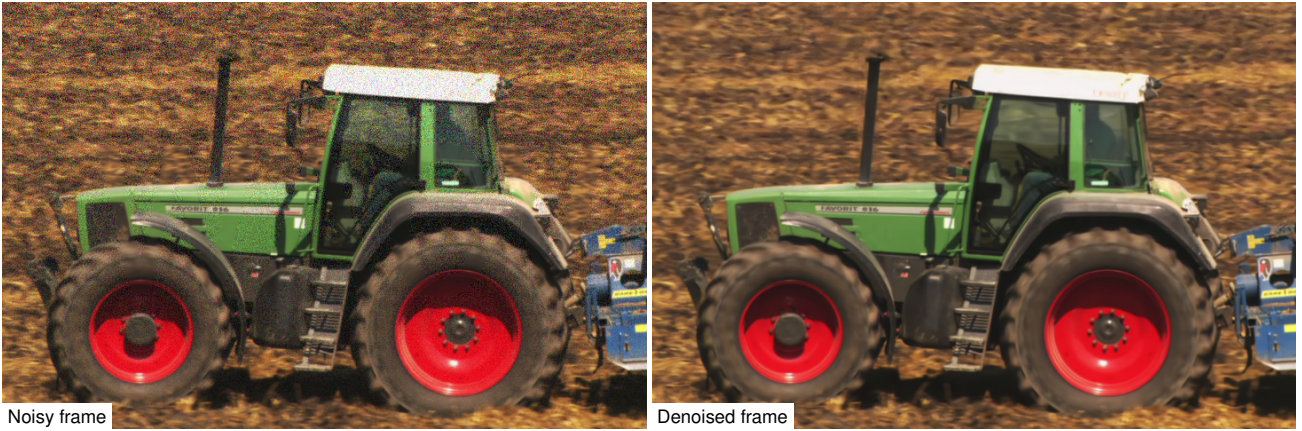

	\begin{center}
		\def\imagesize{0.49\textwidth}
		\overimg[width=\imagesize]{fig/spatial_noise_map/noisy096.png}{Noisy frame} 
		\overimg[width=\imagesize]{fig/spatial_noise_map/096.png}{Denoised frame}
	\caption{A noisy frame with spatial noise map from figure 6 in the main paper and the corresponding denoised by the self-supervised online MF2F, when fine-tuning the noise map input and keeping the network weights fixed.}
	\label{fig:spatial-noise}
	\end{center}
\end{figure*}

In Fig \ref{fig:spatial-noise}, we show an example frame of the noisy video, contaminated with the spatial Gaussian noise from the spatial noise map in figure 6 of the main paper. In the same figure, we show the corresponding denoised frame.

\begin{figure*}
    \centering
    
        \subfloat[Noisy]{    \includegraphics[width=0.32\textwidth]{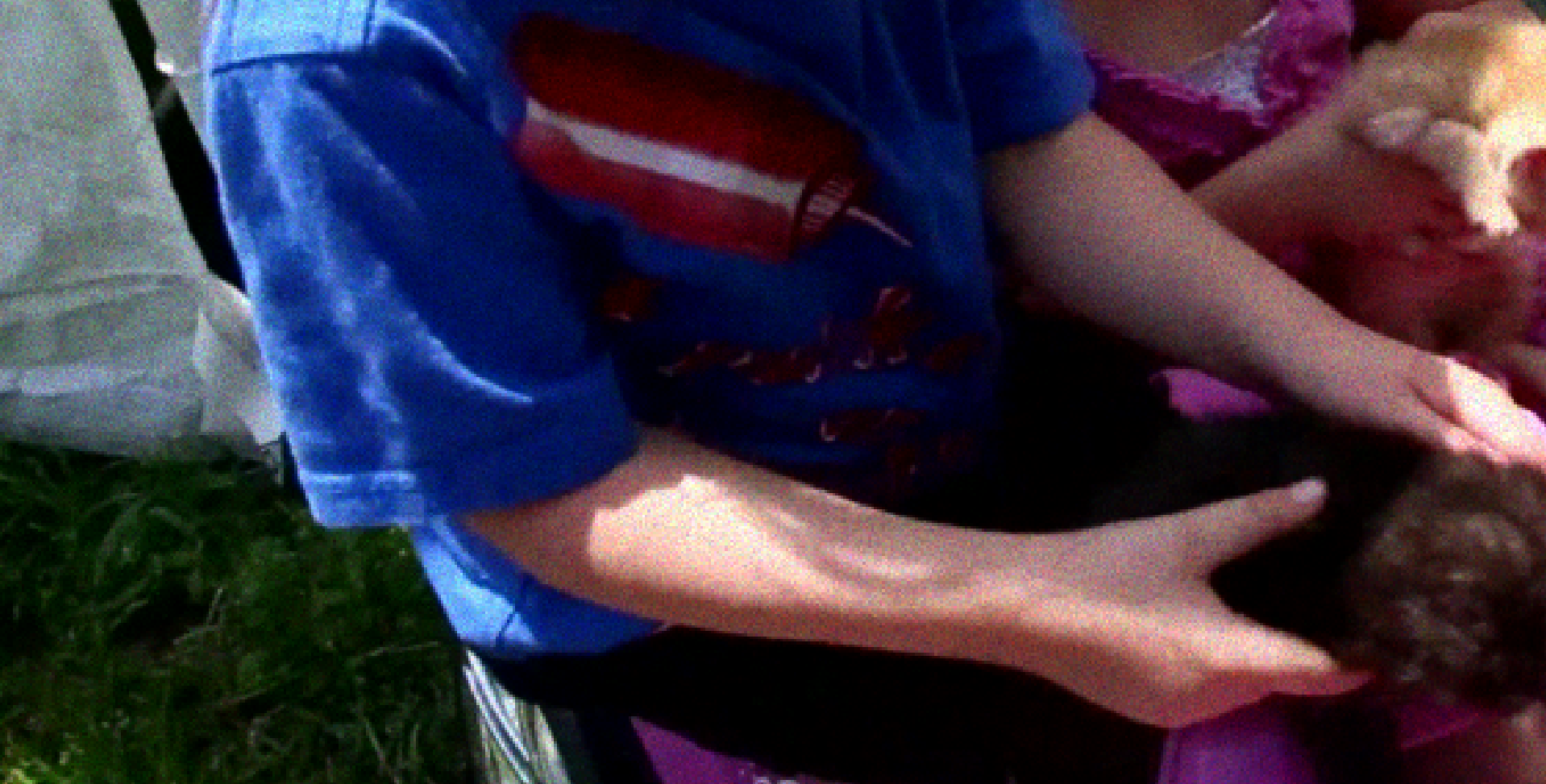}}
        \subfloat[FastDVDnet scalar variance map]{   \includegraphics[width=0.32\textwidth]{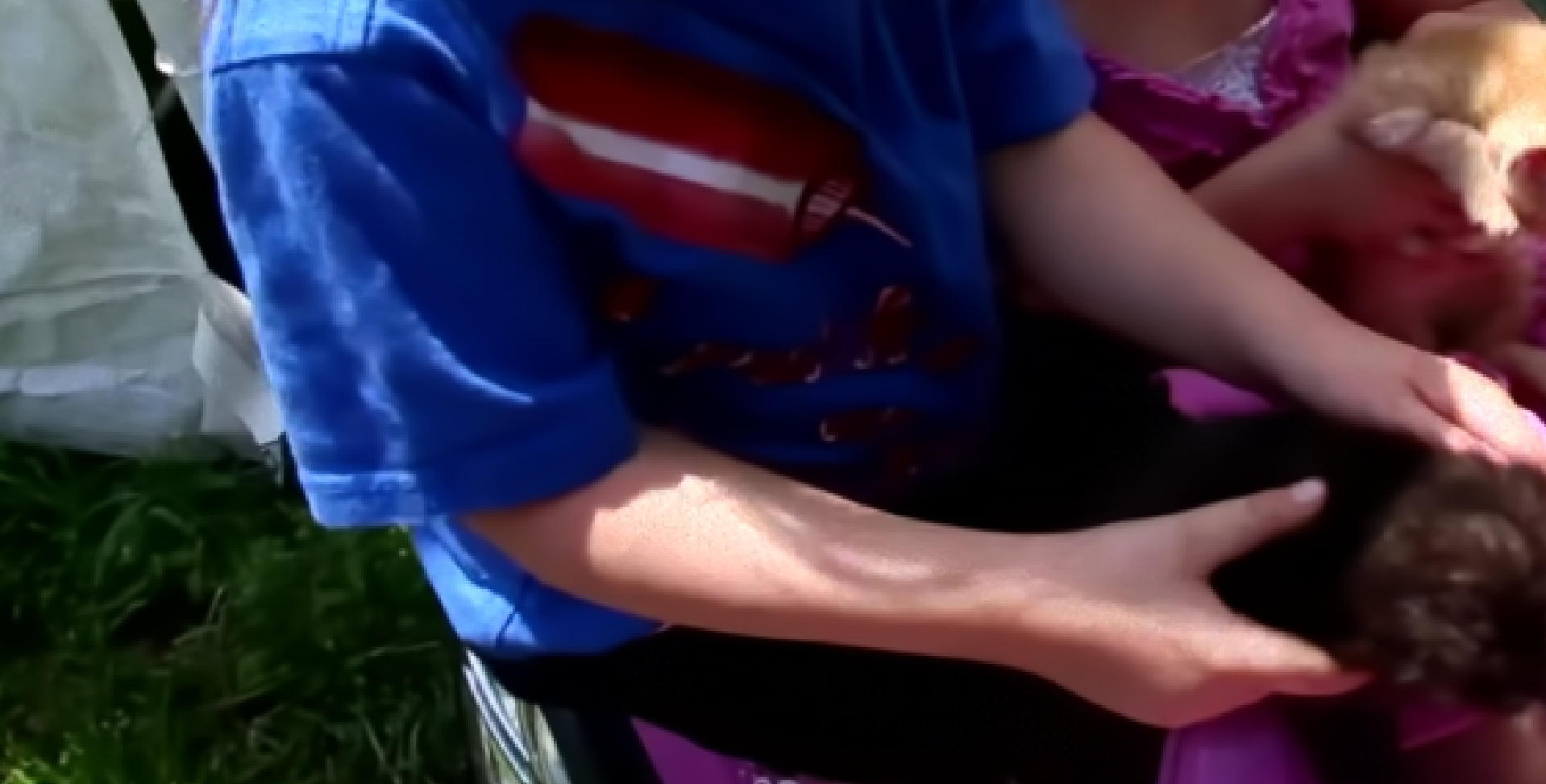}}
        \subfloat[FastDVDnet per-level variance map]{    \includegraphics[width=0.32\textwidth]{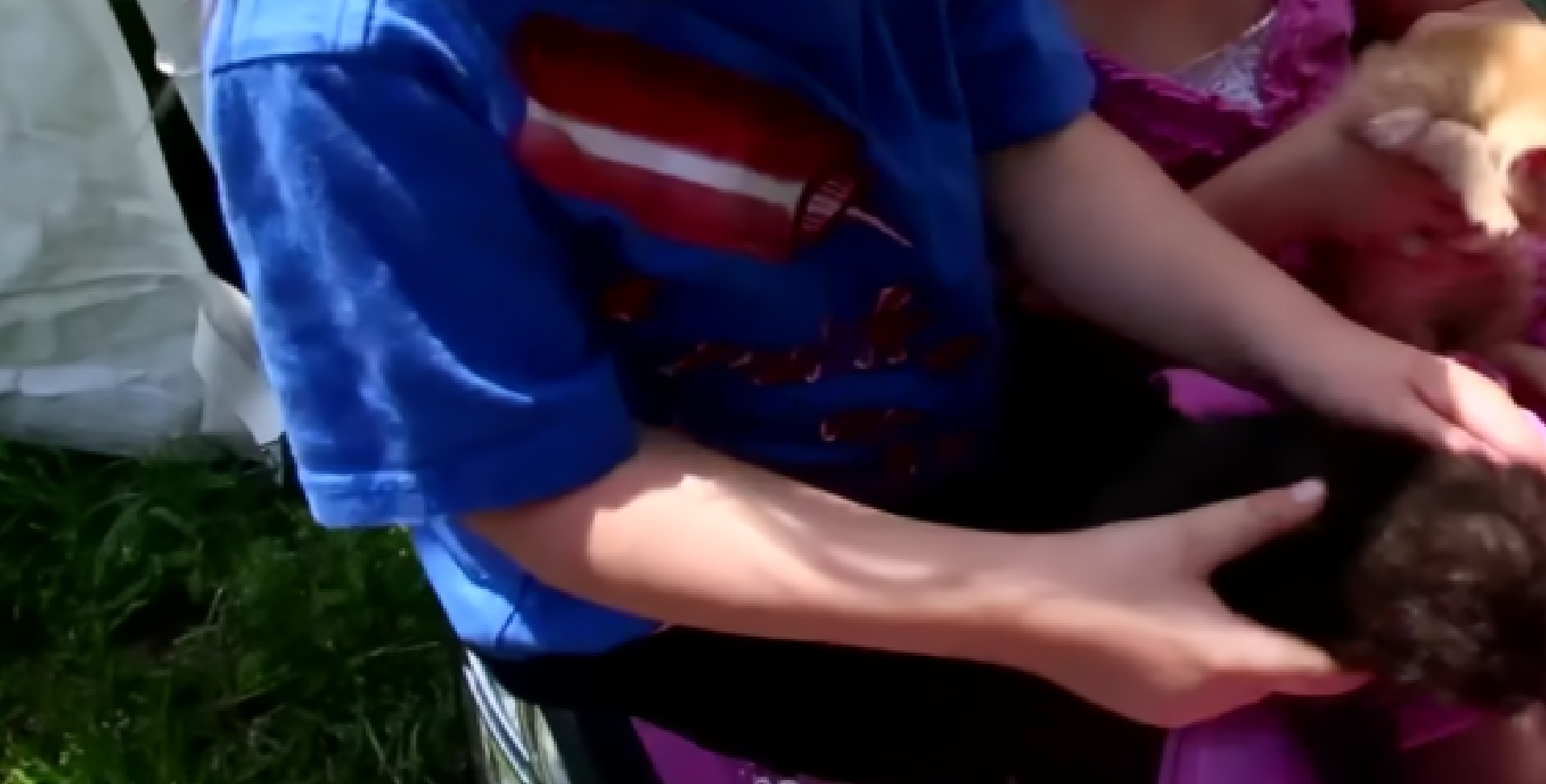}} 
        
        \setcounter{subfigure}{0}
        \subfloat[Noisy]{    \includegraphics[width=0.32\textwidth]{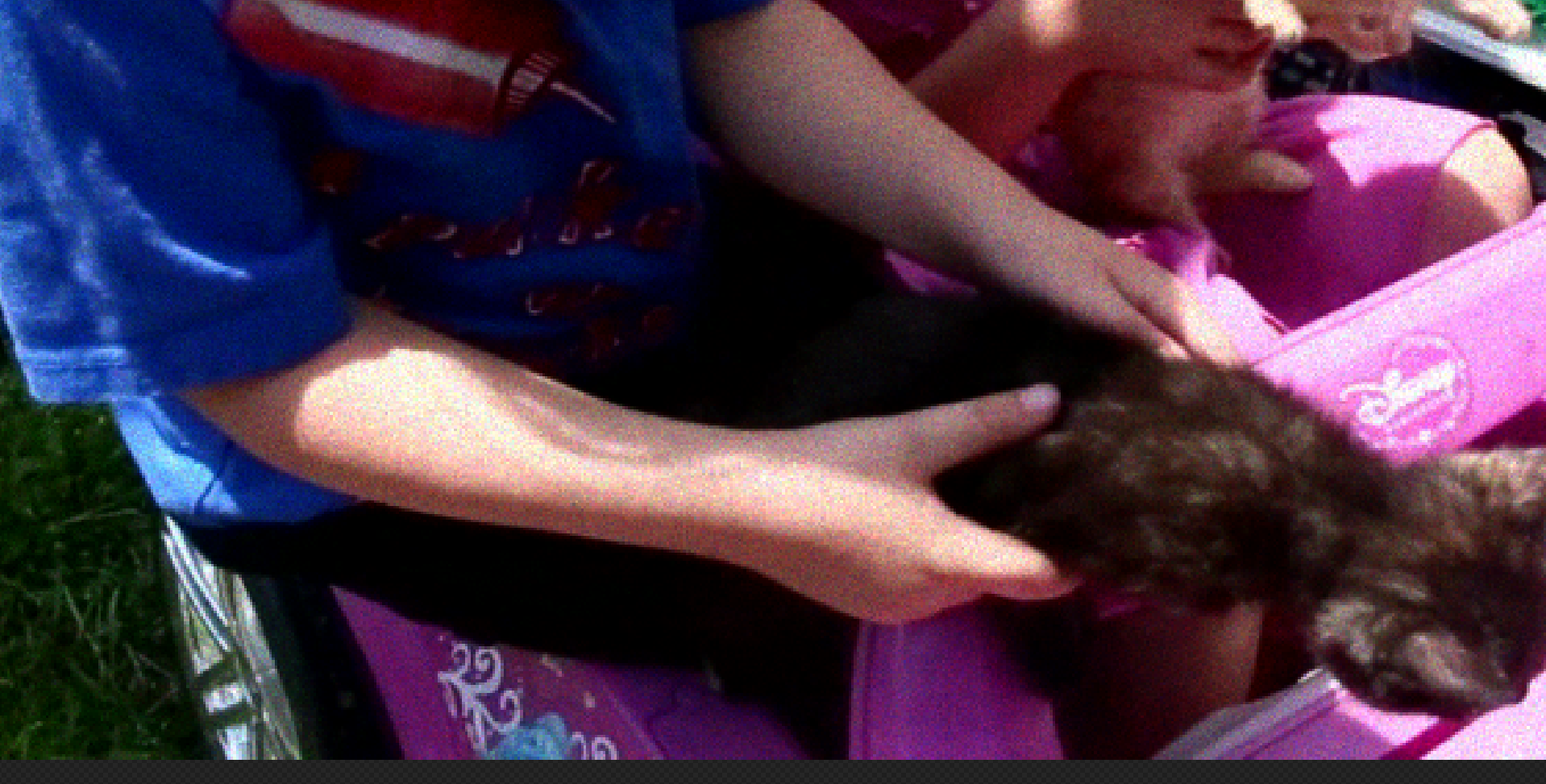}}
        \subfloat[FastDVDnet scalar variance map]{   \includegraphics[width=0.32\textwidth]{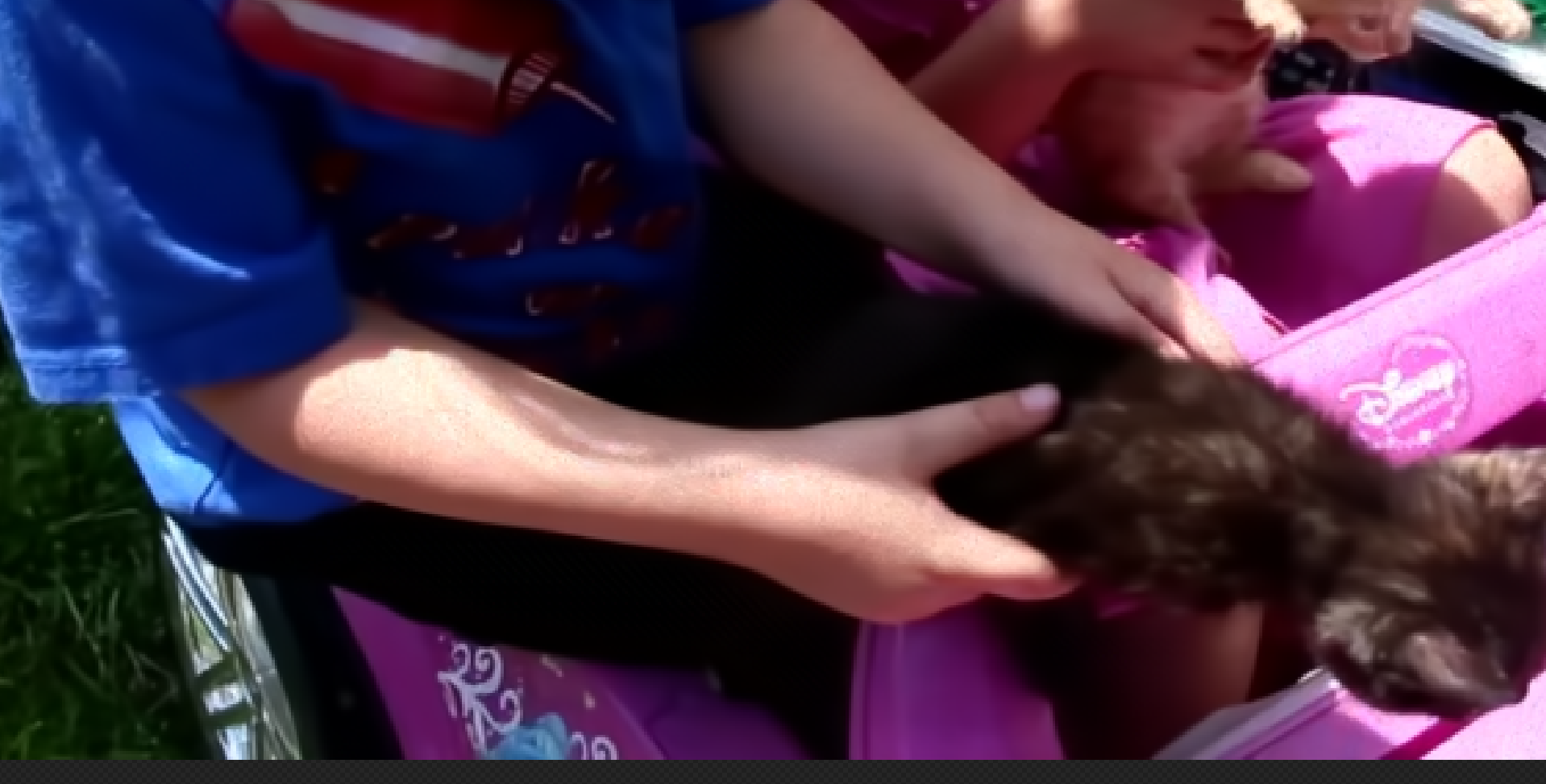}}
        \subfloat[FastDVDnet per-level variance map]{    \includegraphics[width=0.32\textwidth]{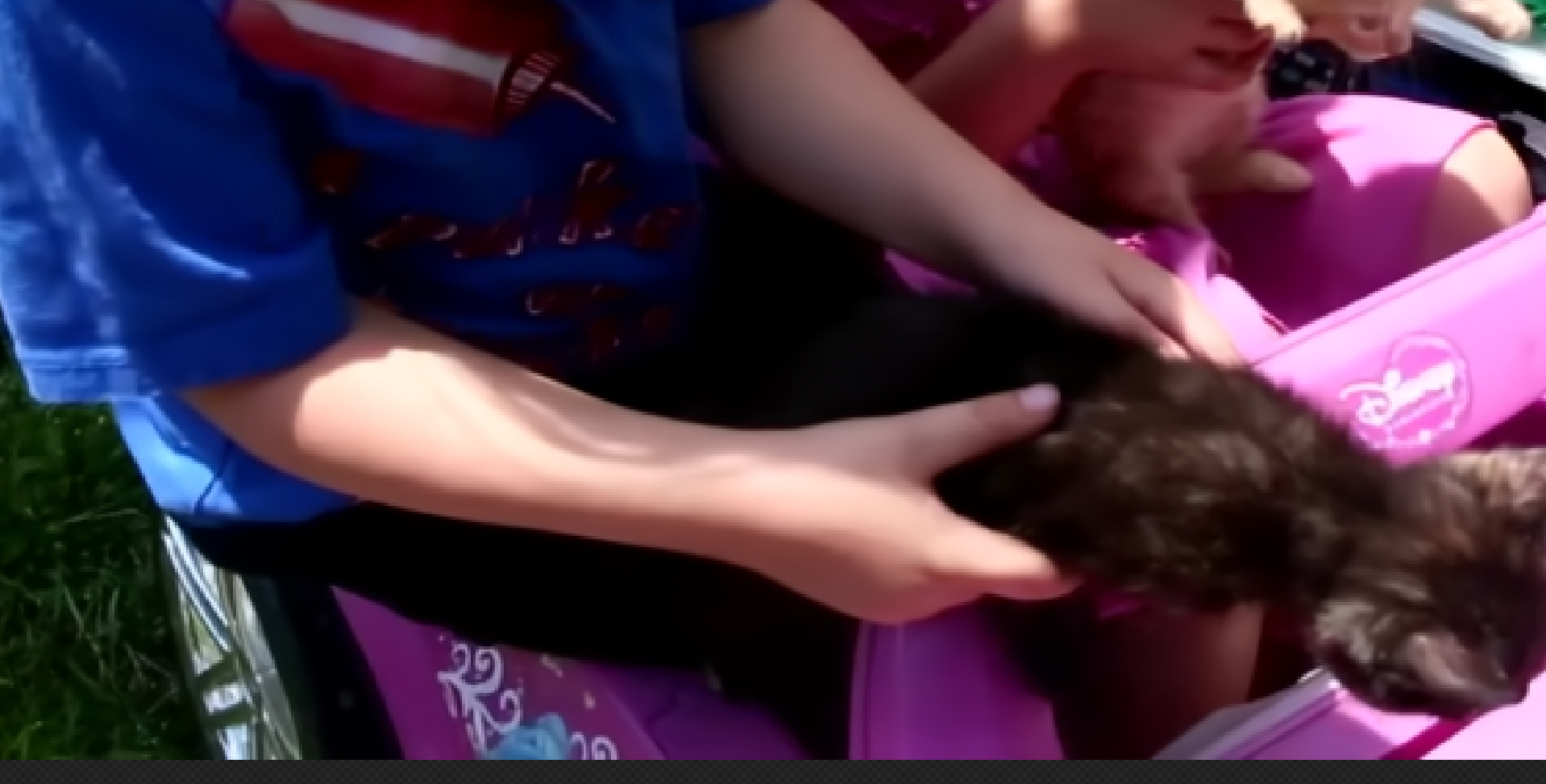}} 
        
        \setcounter{subfigure}{0}
        \subfloat[Noisy]{    \includegraphics[width=0.32\textwidth]{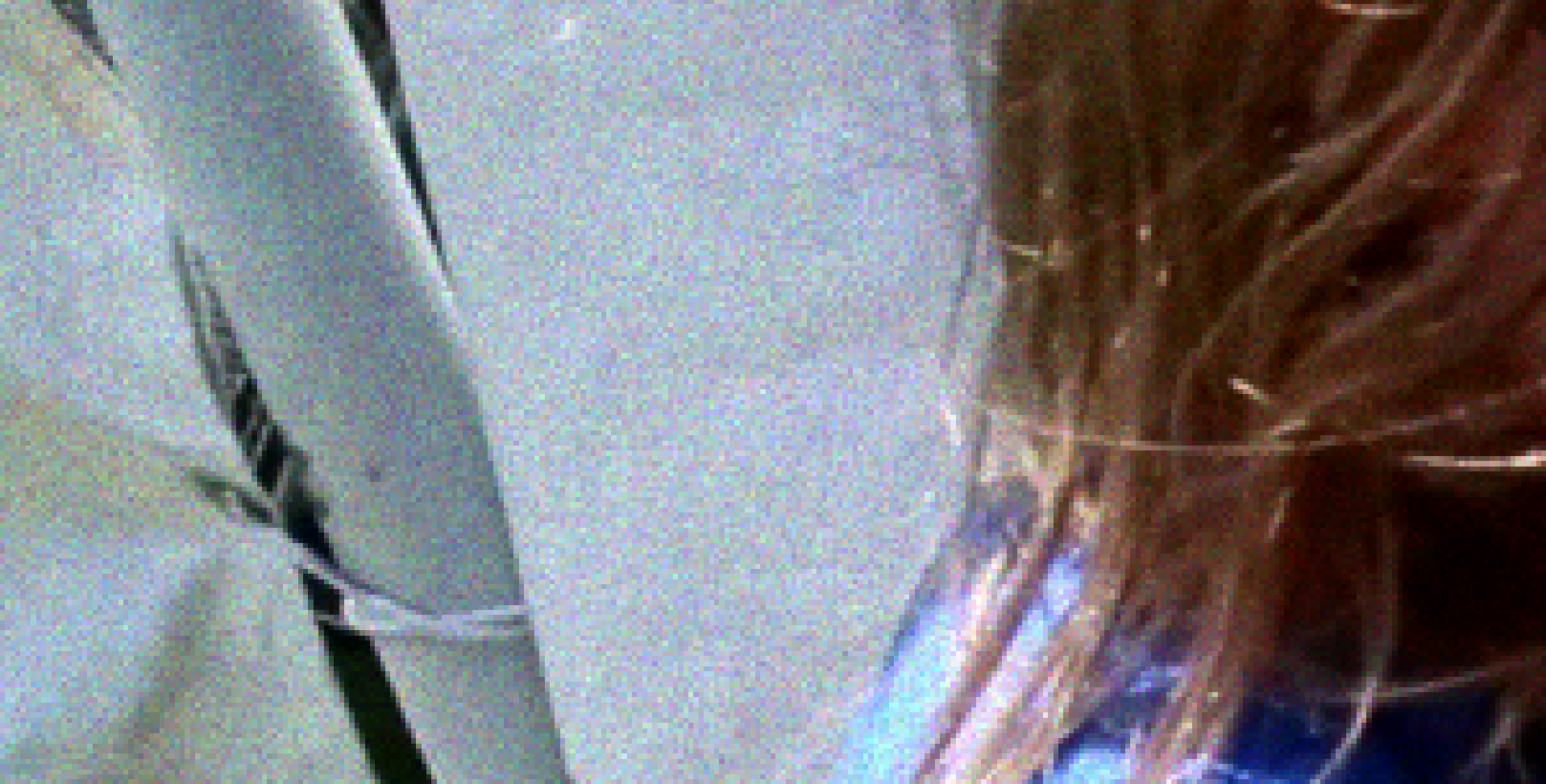}}
        \subfloat[FastDVDnet scalar variance map]{   \includegraphics[width=0.32\textwidth]{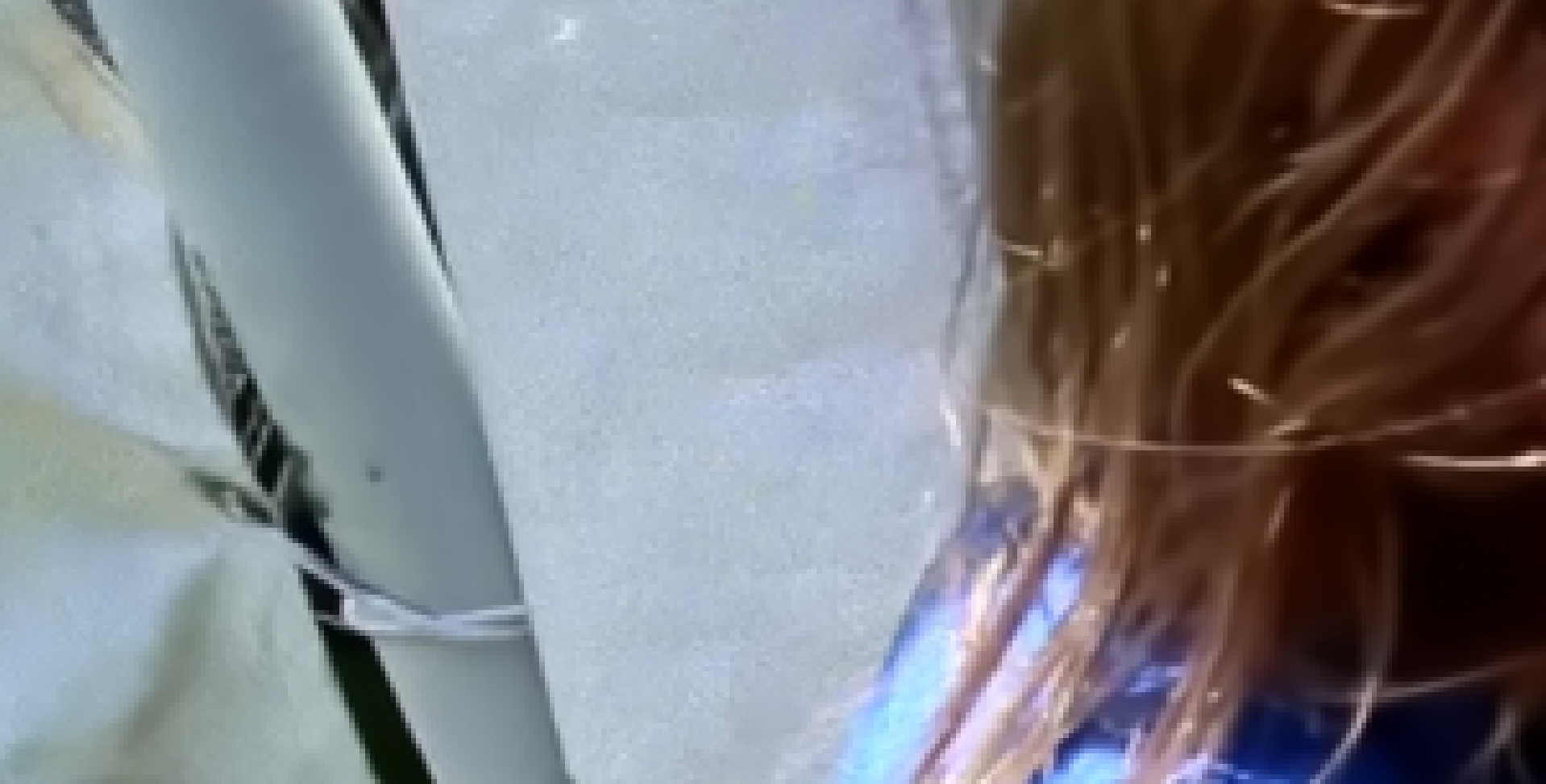}}
        \subfloat[FastDVDnet 8 sigmas variance map]{    \includegraphics[width=0.32\textwidth]{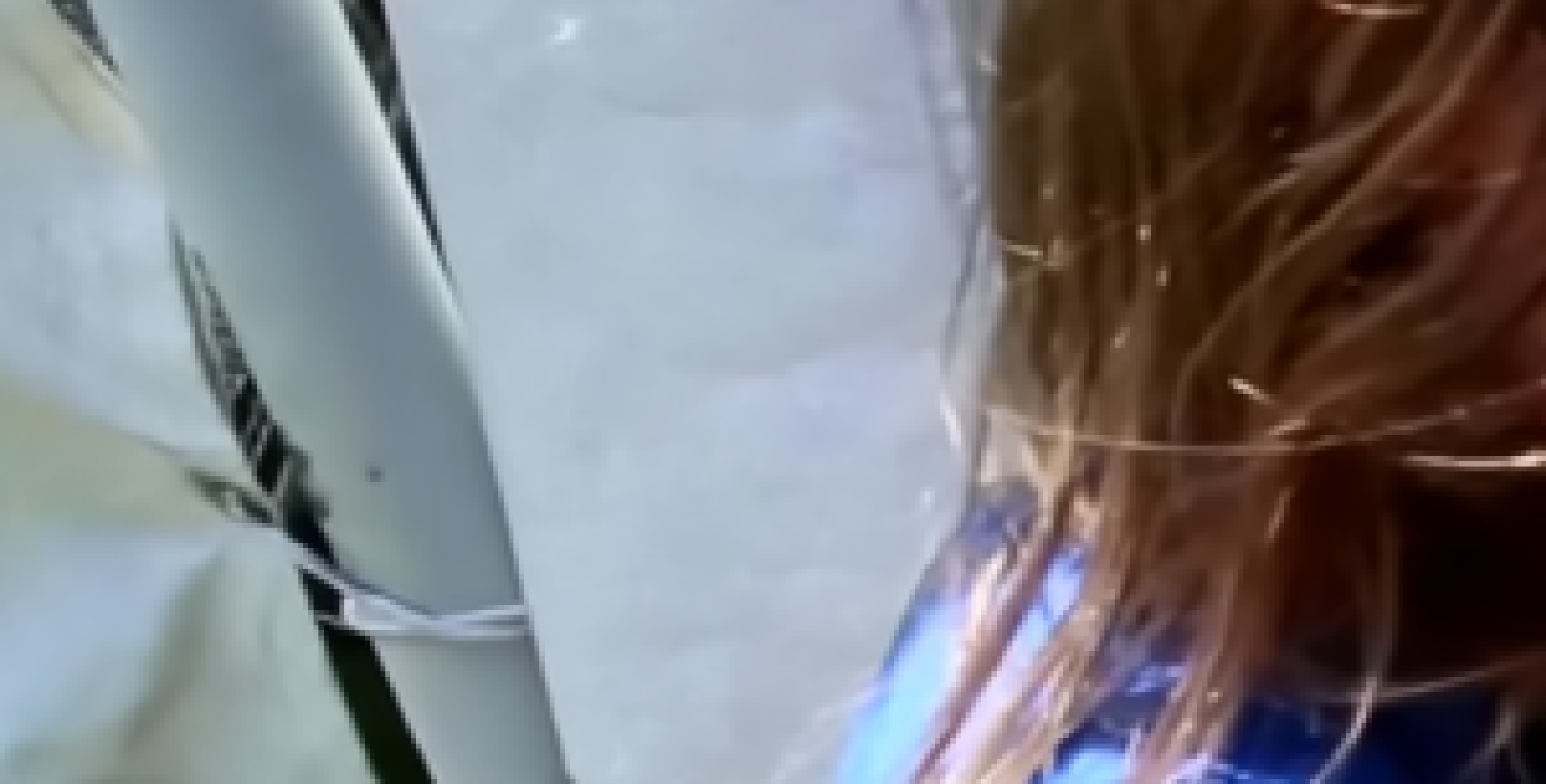}} 
        
        \setcounter{subfigure}{0}
        \subfloat[Noisy]{    \includegraphics[width=0.32\textwidth]{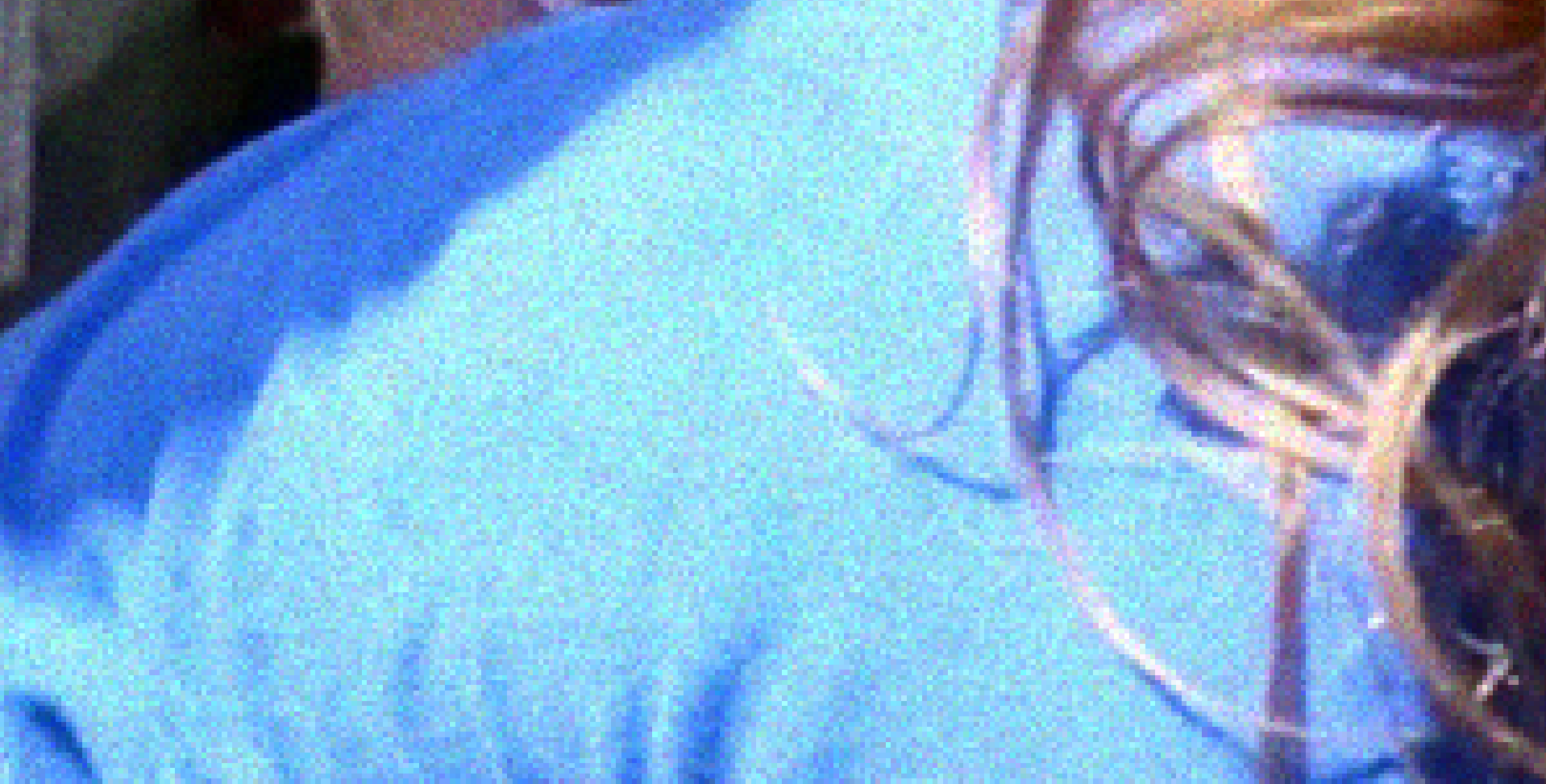}}
        \subfloat[FastDVDnet scalar variance map]{   \includegraphics[width=0.32\textwidth]{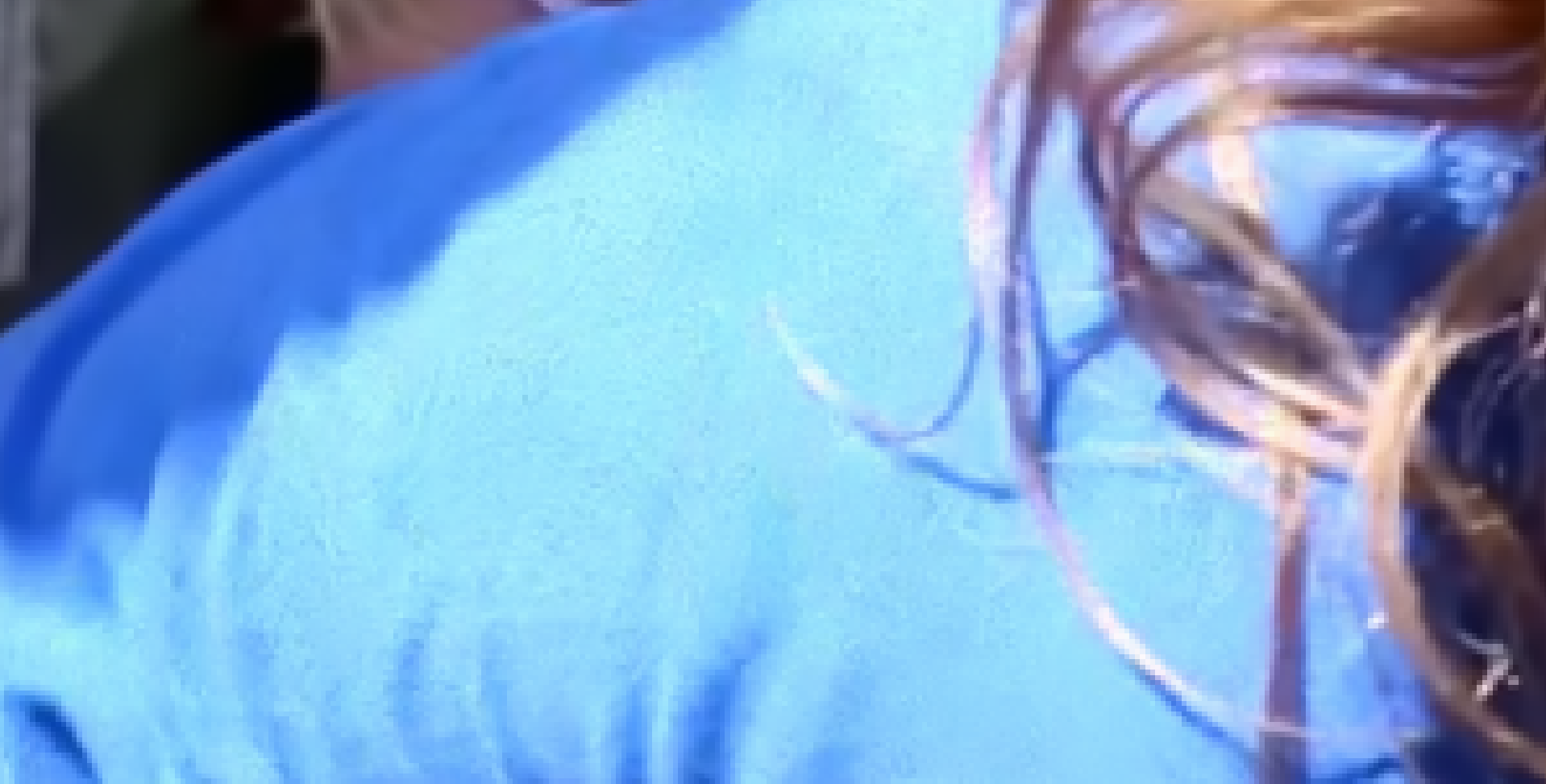}}
        \subfloat[FastDVDnet per-level variance map]{    \includegraphics[width=0.32\textwidth]{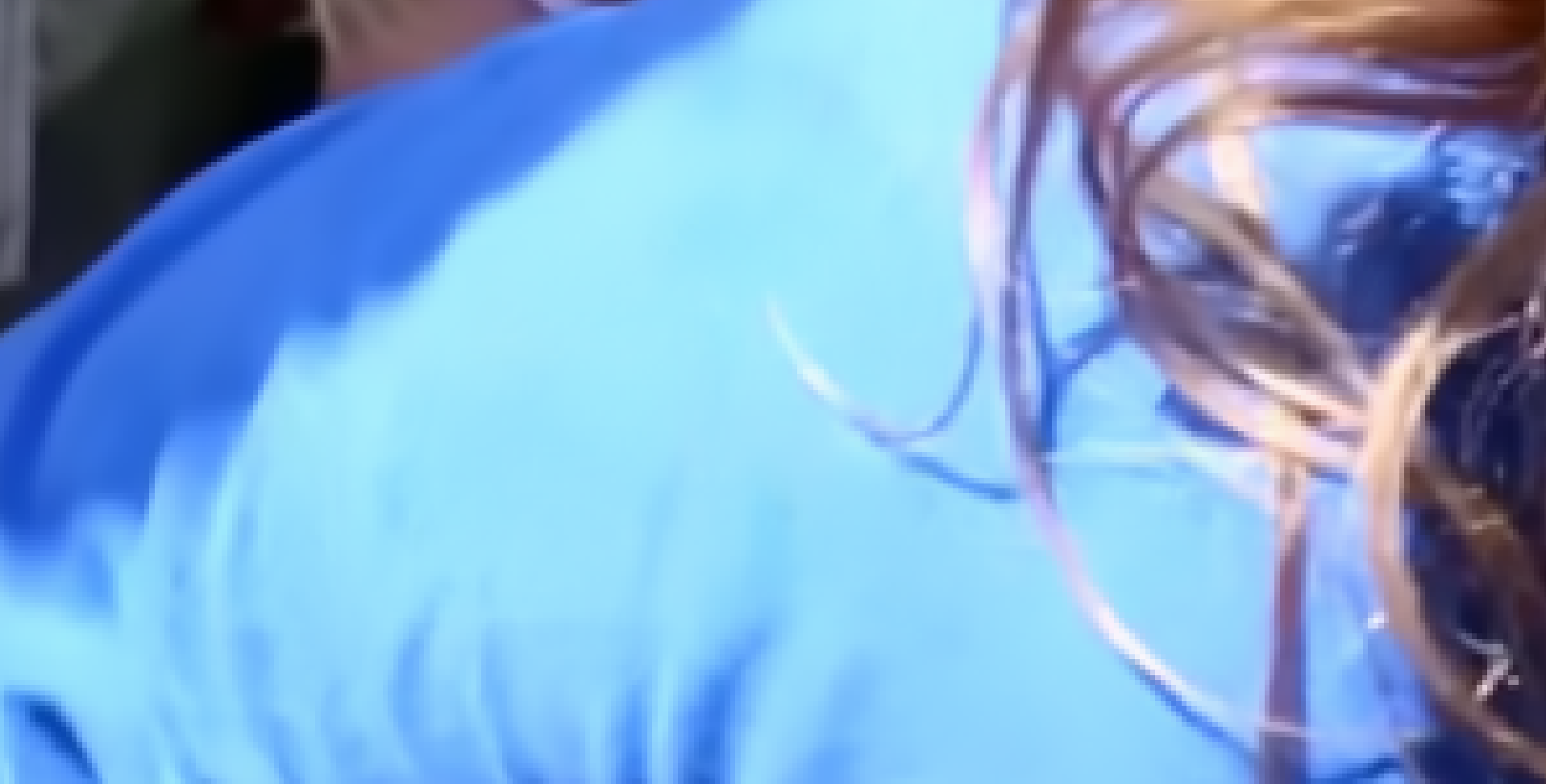}}

	\caption{Comparison between a constant variance map and per-level variance map for an image with Poisson noise of $p=1$. Results with the constant variance map still contain remaining noise for bright areas. Contrast has been linearly scaled for visualization. Notice that no color variance was applied (it is not within the scope of this work to reproduce a complete image pipeline)}
	\label{fig:variance_map}
\end{figure*}

\begin{figure*}
    \centering
        \includegraphics[clip,trim=0cm 0cm 0cm 0cm, width=0.7\linewidth]{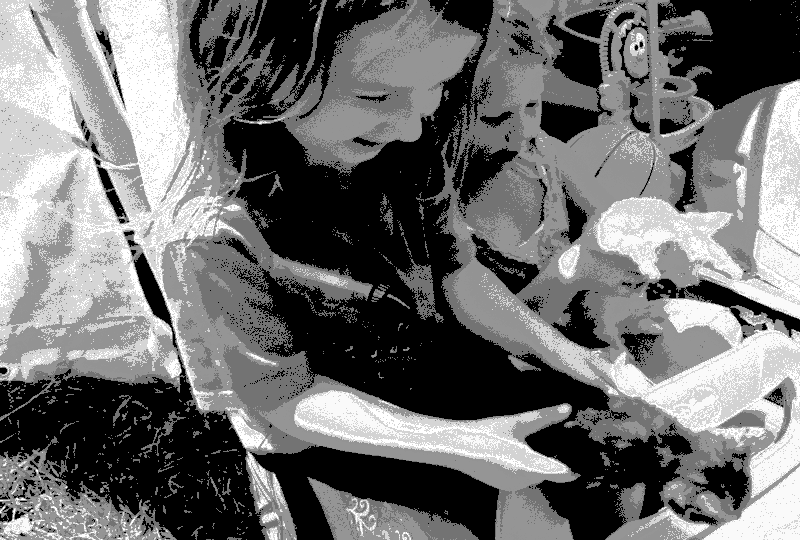}
	\caption{Obtained variance map for Poisson noise $p=1$ (we display the square root of the variance map, \textit{i.e.} the standard deviation). The $\sigma$ found by fine-tuning the constant variance map was 10.41.}
	\label{fig:noisemap_8sigmas}
\end{figure*}

\section{Fine-tuning half of the weights}
The FastDVDnet architecture~\cite{tassano2019fastdvdnet} consists of  two cascaded blocs of U-net. In the previous section, we showed that fine-tuning the parameters of the noise map  while leaving the weights  fixed can achieve good results for certain  types of noise. While, in the main article, we have seen that fine-tuning all the weights of the network permits to adapt to a wider range of  noise types. In this section we investigate if we can update a smaller part of the network in order to attain the same adaptation capacity. 

We will fine-tune half of the network weights. For that we consider four ways of splitting the weights. 
We can fine-tune the weights corresponding to the first Unet (denoted \textit{first bloc}), or the second one  (\textit{second bloc}), while leaving the other fixed. 
But also we can fine-tune the weights of the encoder parts of both Unets (denoted \textit{``encoder"}) or the decoder parts (\textit{``decoder''}).

The average PSNR obtained with this  fine-tuning experiments are reported in Table \ref{tab:half_weights_FT}. The averages are computed over seven video sequences of the Derf dataset and ten video sequences of the Vid3oC-10 dataset.
Surprisingly, one of the configuration for half fine-tuning competes with the full training. Indeed training only the encoder parts of both Unet consistently attains the performance obtained by fine-tuning the \textit{full weights}. This is true for all the tested noises. 
This seems to indicate that most of the "noise-specific" work is being done in the encoders. 

We can also observe that the other half-fine-tuning configurations reach a good performance and sometimes overtaking  the noise-specific FastDVDnet trained with supervision. 

Furthermore, in case of fine-tuning the end of the network (decoder of both Unet or encoder \& decoder of the last Unet), fine-tuning half of the network does not require to back-propagate through the whole network . Thus, this allows to reduce the computational memory needed (however, the table \ref{tab:half_weights_FT} shows it slightly affects the performance compared with a full-weights training).

\begin{table*}
\centering
\begin{tabular}{|c|c|c|c||c|c||c|} 
\hline
\multicolumn{2}{|c|}{Dataset \& noise} & Encoder & Decoder & Bloc1 & Bloc2 & Full weights\\
\hline
\multirow{7}{*}{\rotatebox{90}{Derf}} &
Gaussian 20 & \textbf{\color{gray}37.28} & \textbf{\color{gray}37.28} & 37.03
 & 37.21 & \textbf{37.42}\\
&Gaussian 40 & \textbf{\color{gray}34.19} & 33.89 & 33.94 & 33.27 & \textbf{34.24}\\
&Poisson 1 & \textbf{\color{gray}40.32} & 40.07 & 39.92 & 38.55 & \textbf{40.39} \\
&Poisson 8 & \textbf{35.56}  & \textbf{\color{gray}35.49} & 35.34 &  \textbf{\color{gray}35.48} & \textbf{35.57} \\
&Box 40 3 &  \textbf{35.47}  &\textbf{\color{gray}35.39} & 35.18 & 35.12 & \textbf{35.50} \\
&Box 65 5 & \textbf{\color{gray}33.85} & 33.64 & 33.43 & 32.96 & \textbf{34.29}\\
& Demosaicked 4 & \textbf{\color{gray}34.70} & 34.60 & 34.46 & 34.49 & \textbf{34.75} \\
\hline
\multirow{7}{*}{\rotatebox{90}{Vid3oC-10}} &
Gaussian 20 & \textbf{37.35} & \textbf{37.33} & \textbf{\color{gray}37.26} & 37.09  & \textbf{37.32}\\
&Gaussian 40 & \textbf{34.19} & 33.90 & \textbf{\color{gray}34.10} & 33.12 & \textbf{34.17}\\
&Poisson 1 & \textbf{40.05} & 39.70 & \textbf{\color{gray}39.79} & 38.00 & \textbf{40.01} \\
&Poisson 8 & \textbf{35.00} & 34.76 & \textbf{\color{gray}34.90} & 34.44 & \textbf{34.99} \\
&Box 40 3 & \textbf{36.65} & \textbf{\color{gray}36.56} & 36.47 & 35.72 & \textbf{36.65}\\
&Box 65 5  & \textbf{35.65} & \textbf{\color{gray}35.46} & \textbf{\color{gray}35.42} & 34.32 & \textbf{35.65} \\
&Demosaicked4 & \textbf{33.96} & 33.66 & \textbf{\color{gray}33.83} & 33.26 & \textbf{33.95}\\
\hline
\end{tabular}
\caption{Comparison of average PSNR over all the sequences for a given dataset and type of noise when fine-tuning the all weights or only half of them. The best PSNR is in bold. The second blind is in gray.} 
\label{tab:half_weights_FT}
\end{table*}

\section{Additional results}

In this section we present more results obtained with the proposed methods for different noise types. All the fine-tuned networks are obtained from the same  pre-trained FastDVDnet network, trained in a supervised setting for Gaussian noise with noise level $\sigma = 25$. 
This network is fine-tuned blindly with the proposed method and we show that it behaves as the supervised one for many types and levels of noise. 

The MF2F method was tried on two AWGN, two correlated noise that we call ``box noise'' consisting of AWGN and filtered with a box filter. Finally we also tested on two scaled Poisson noise as well as on demosaicking noise (Poisson noise follows by a demosaicking algorithm).

\begin{figure*}
	\begin{center}
		\def\imagesize{0.25\textwidth}
		\overimg[width=\imagesize,trim={0 0 0 0 },clip]{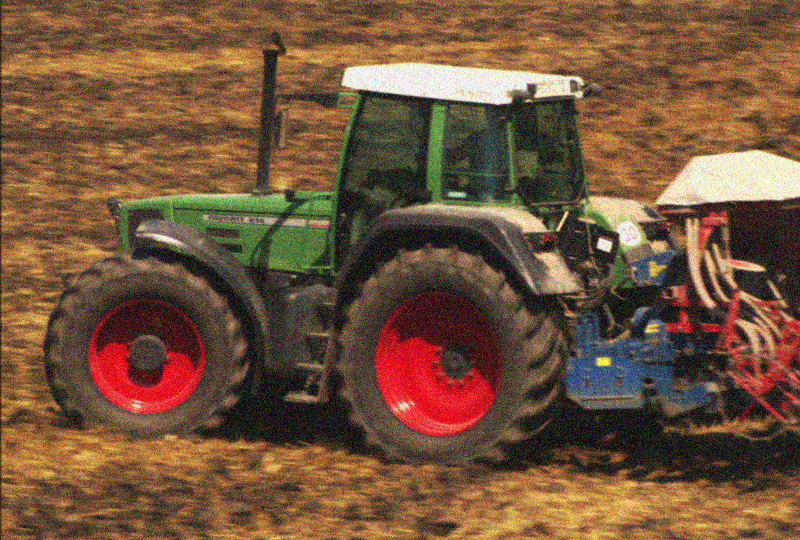}{Gaussian 20}%
		\includegraphics[width=\imagesize,trim={0 0 0 0 },clip]{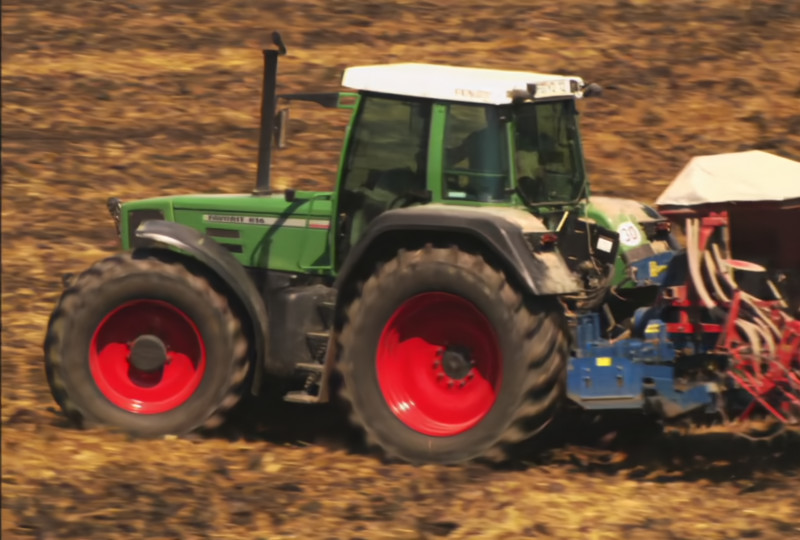}%
		\includegraphics[width=\imagesize,trim={0 0 0 0 },clip]{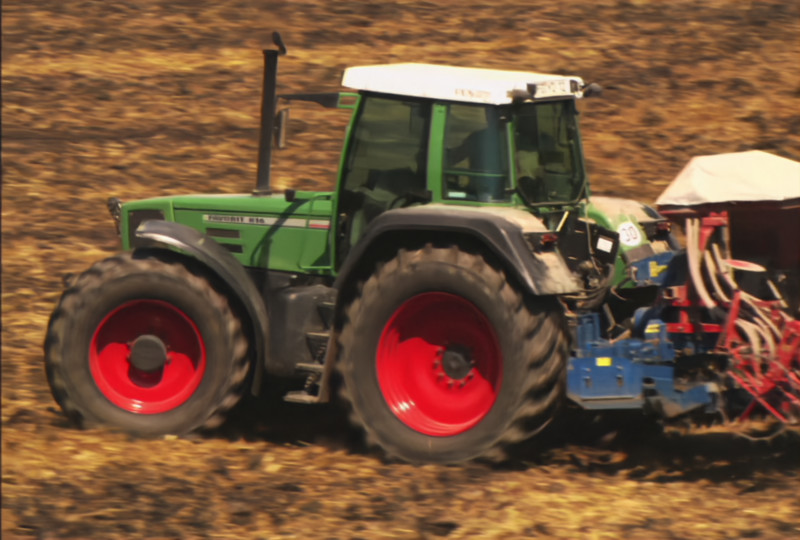}%
		\includegraphics[width=\imagesize,trim={0 0 0 0 },clip]{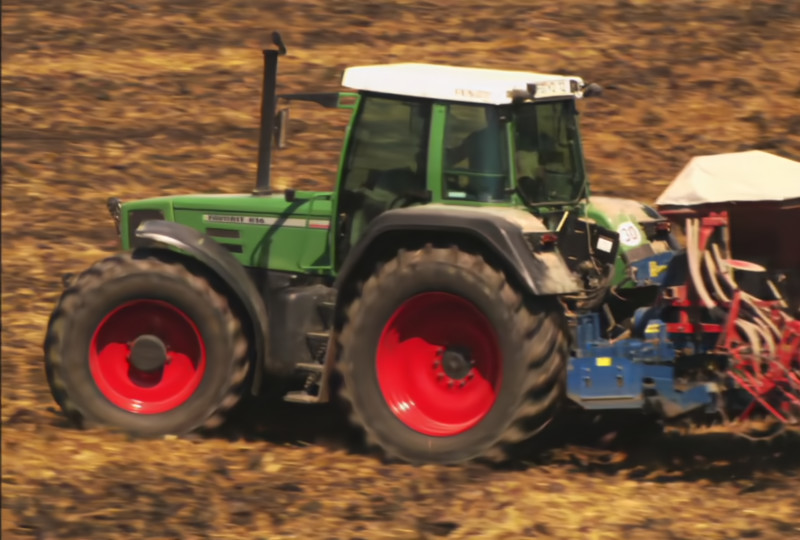}%
				
		\overimg[width=\imagesize,trim={0 0 0 0 },clip]{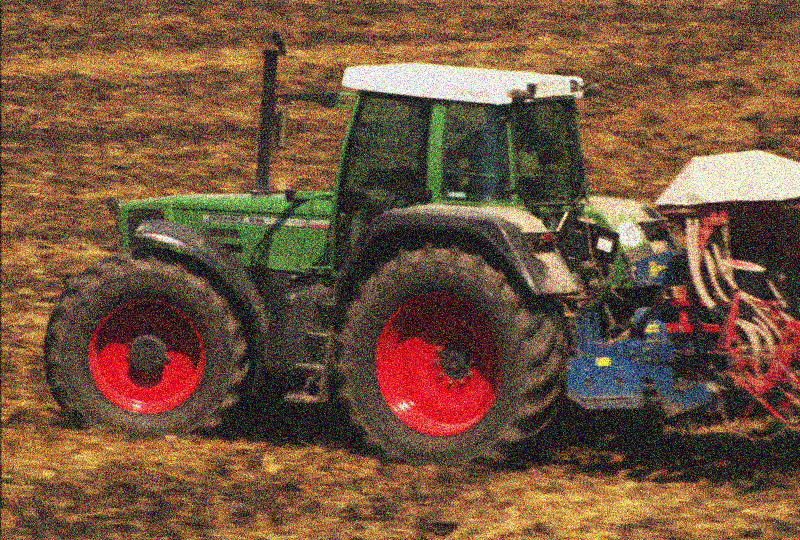}{Gaussian 40}%
		\includegraphics[width=\imagesize,trim={0 0 0 0 },clip]{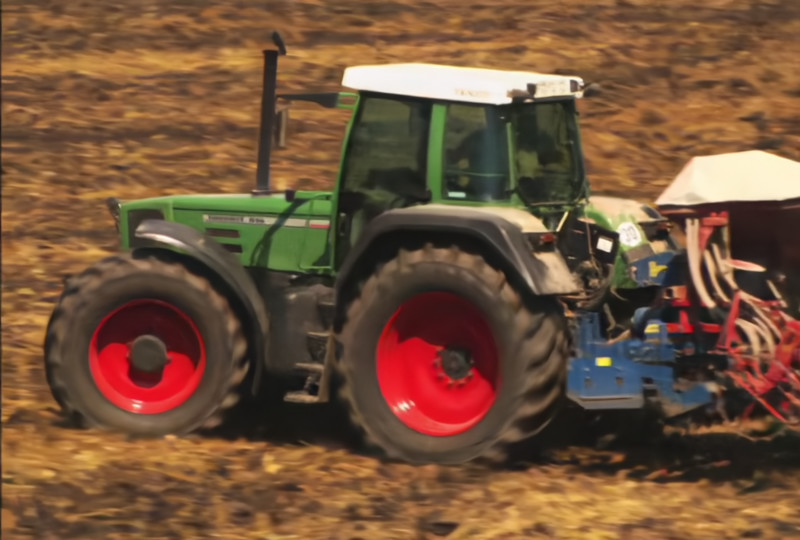}%
		\includegraphics[width=\imagesize,trim={0 0 0 0 },clip]{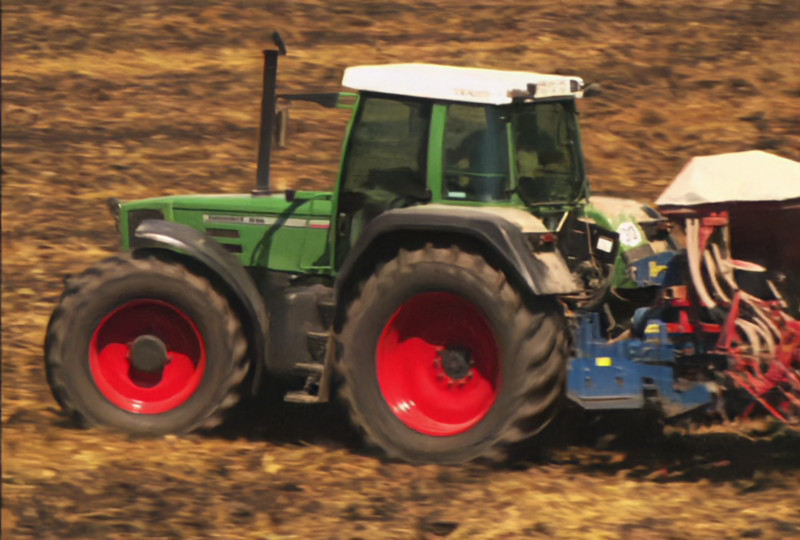}%
		\includegraphics[width=\imagesize,trim={0 0 0 0 },clip]{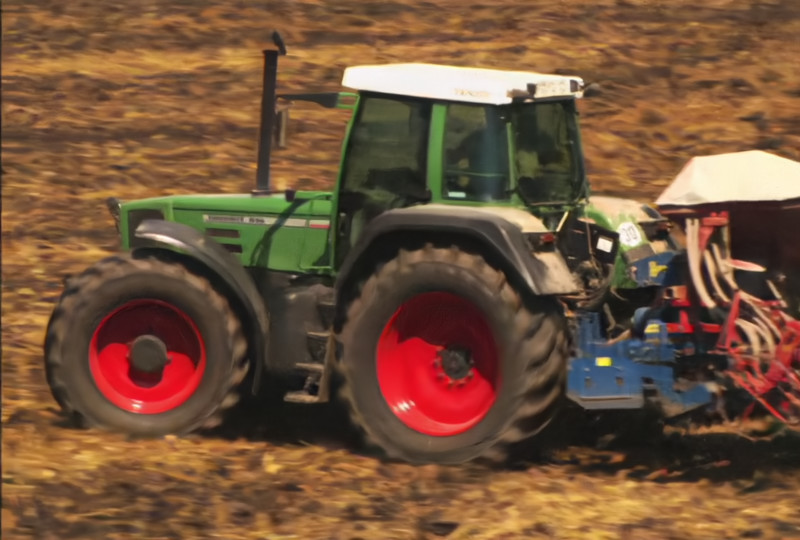}%
				
		\overimg[width=\imagesize,trim={0 0 0 0 },clip]{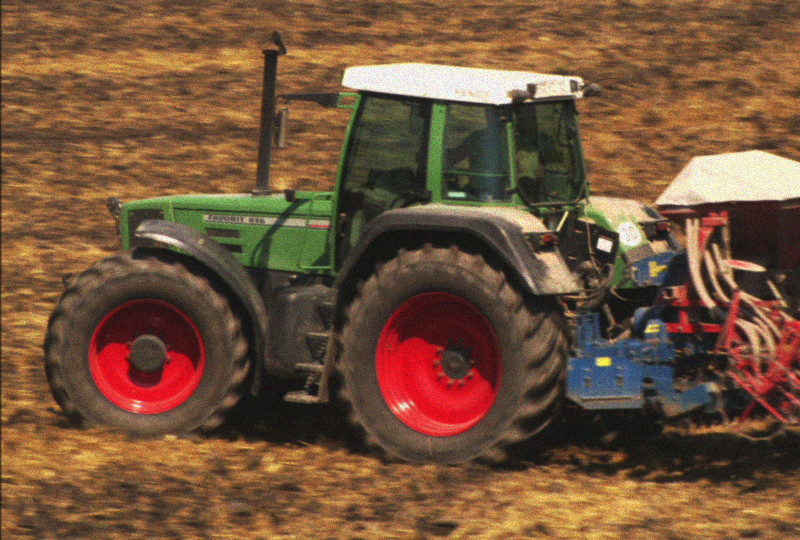}{Poisson 1}%
		\includegraphics[width=\imagesize,trim={0 0 0 0 },clip]{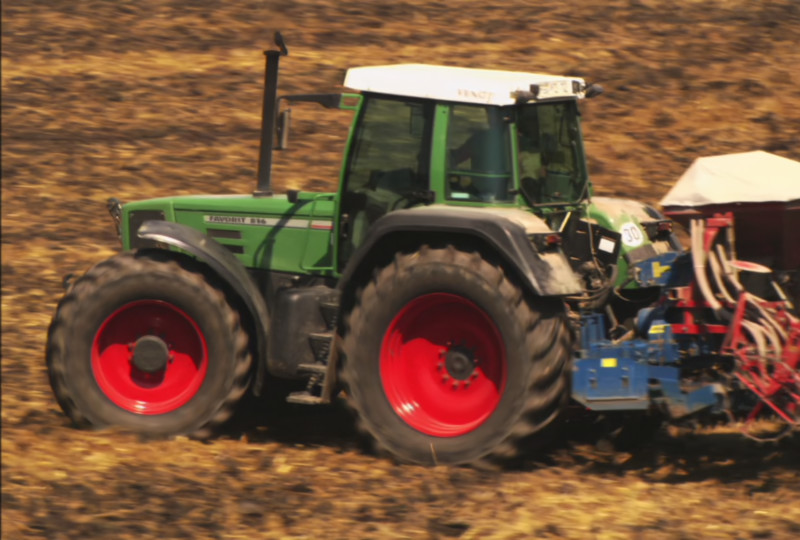}%
		\includegraphics[width=\imagesize,trim={0 0 0 0 },clip]{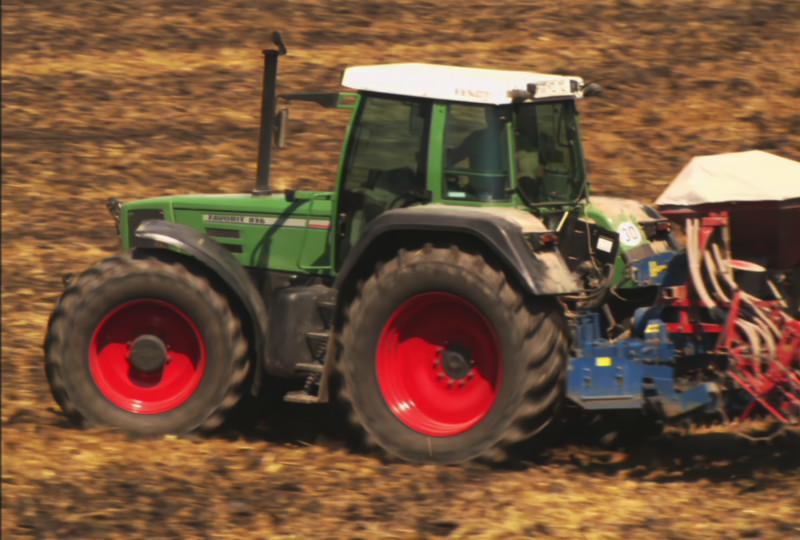}%
		\includegraphics[width=\imagesize,trim={0 0 0 0 },clip]{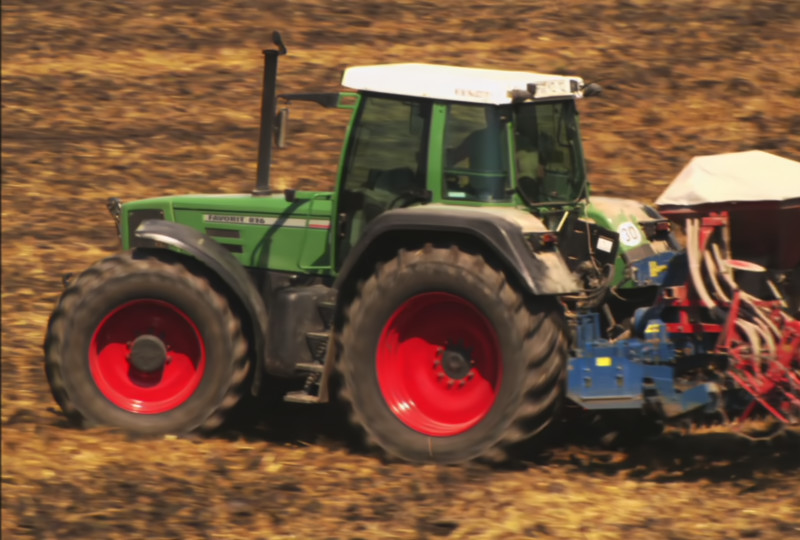}%
					
		\overimg[width=\imagesize,trim={0 0 0 0 },clip]{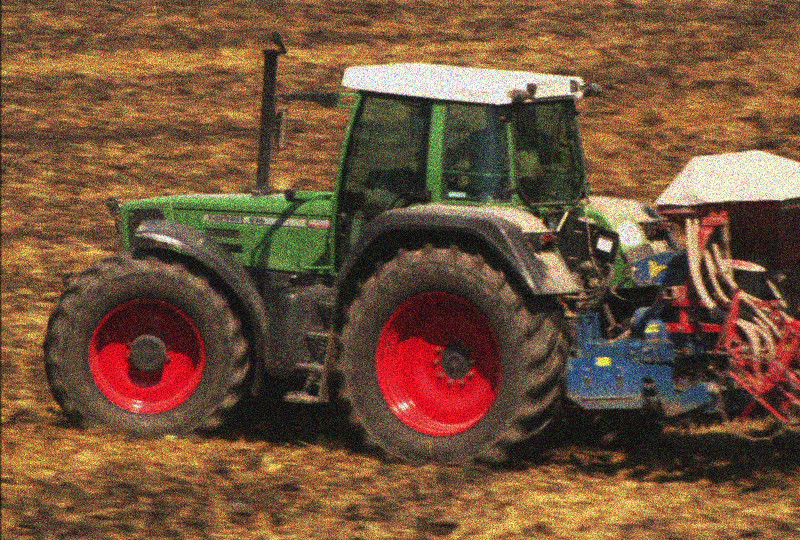}{Poisson 8}%
		\includegraphics[width=\imagesize,trim={0 0 0 0 },clip]{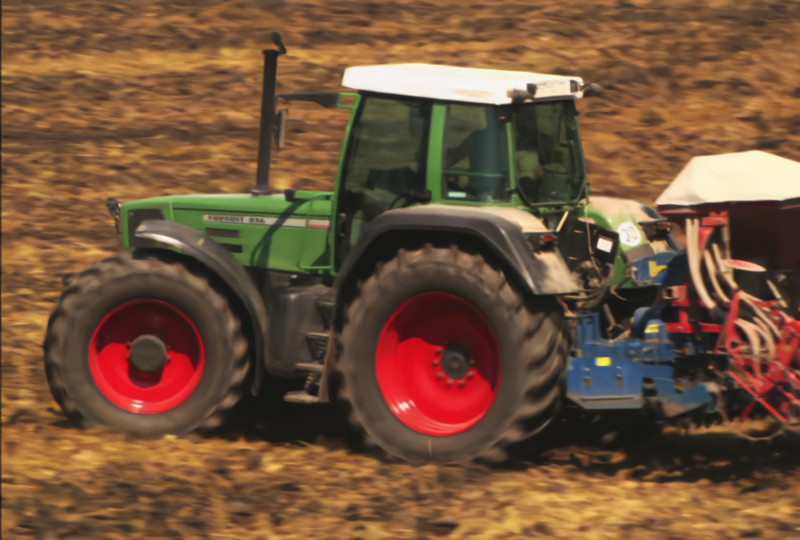}%
		\includegraphics[width=\imagesize,trim={0 0 0 0 },clip]{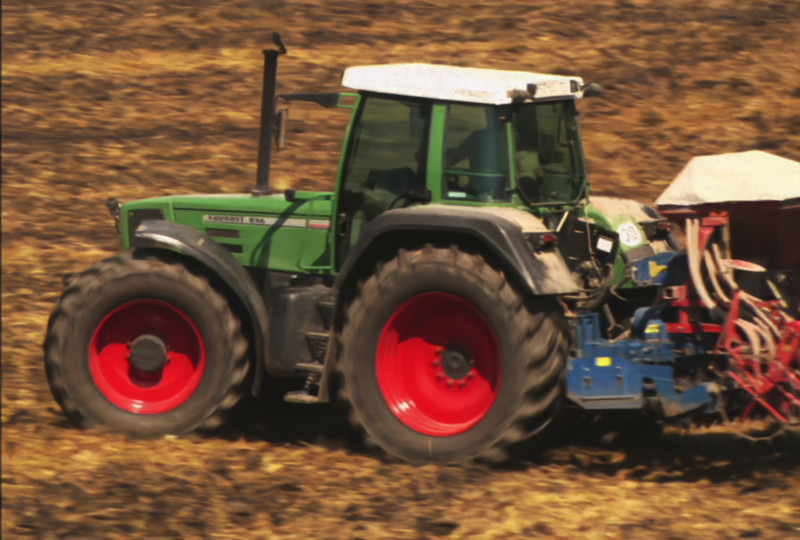}%
		\includegraphics[width=\imagesize,trim={0 0 0 0 },clip]{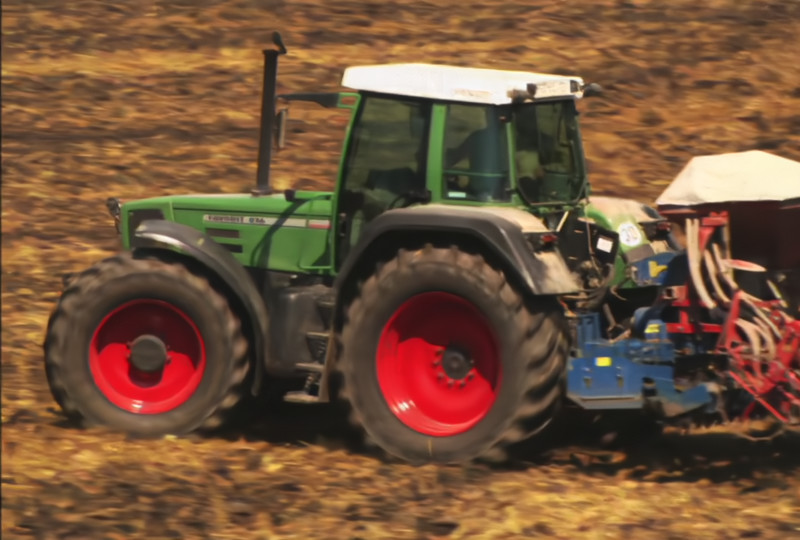}%
				
		\overimg[width=\imagesize,trim={0 0 0 0 },clip]{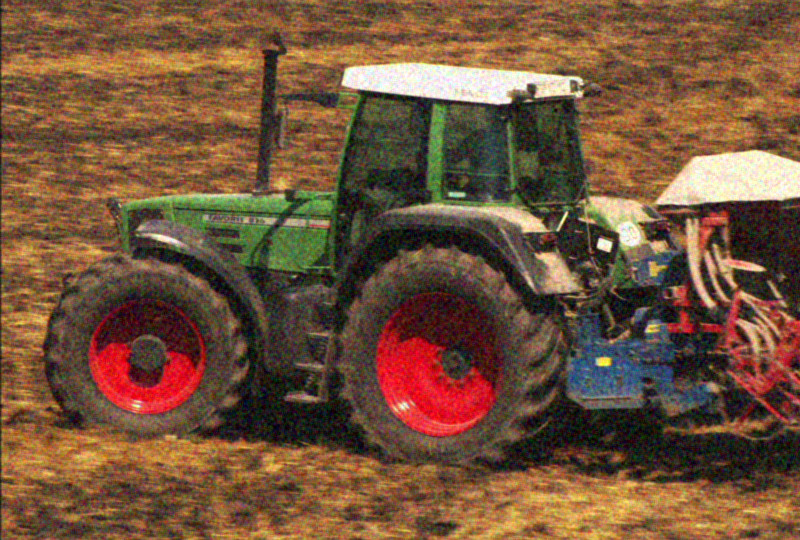}{Box 40 3}%
		\includegraphics[width=\imagesize,trim={0 0 0 0 },clip]{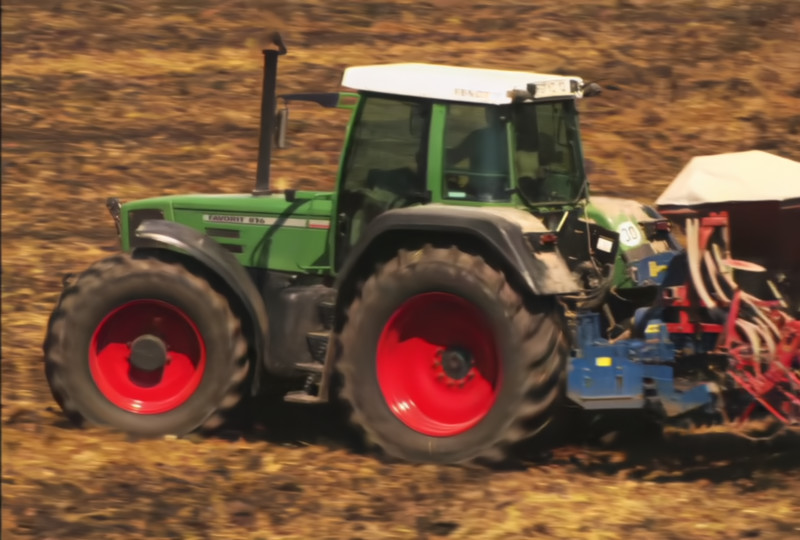}%
		\includegraphics[width=\imagesize,trim={0 0 0 0 },clip]{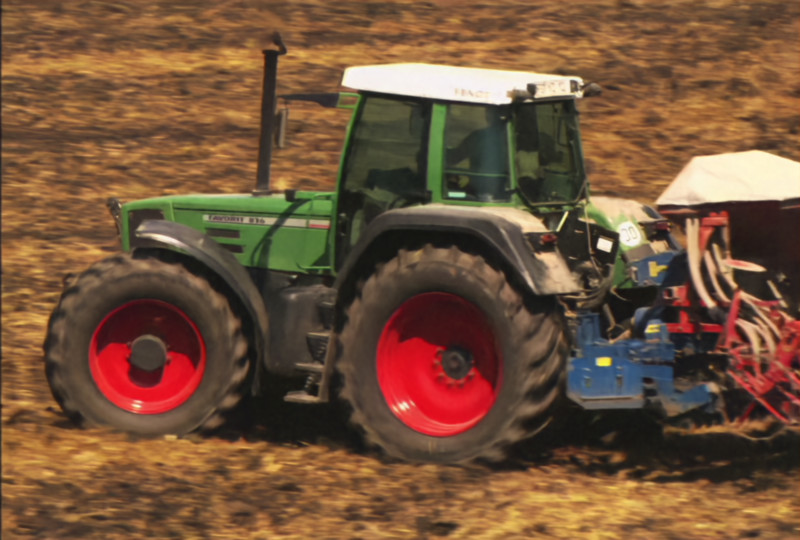}%
		\includegraphics[width=\imagesize,trim={0 0 0 0 },clip]{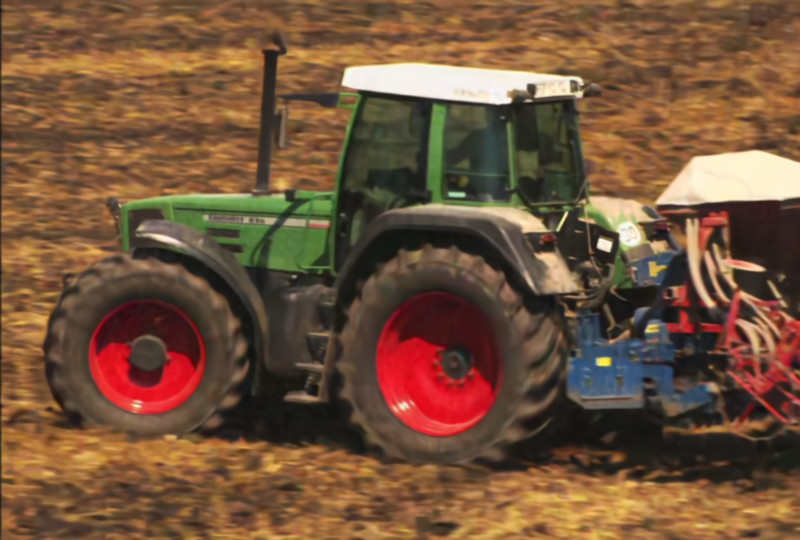}%
					
		\overimg[width=\imagesize,trim={0 0 0 0 },clip]{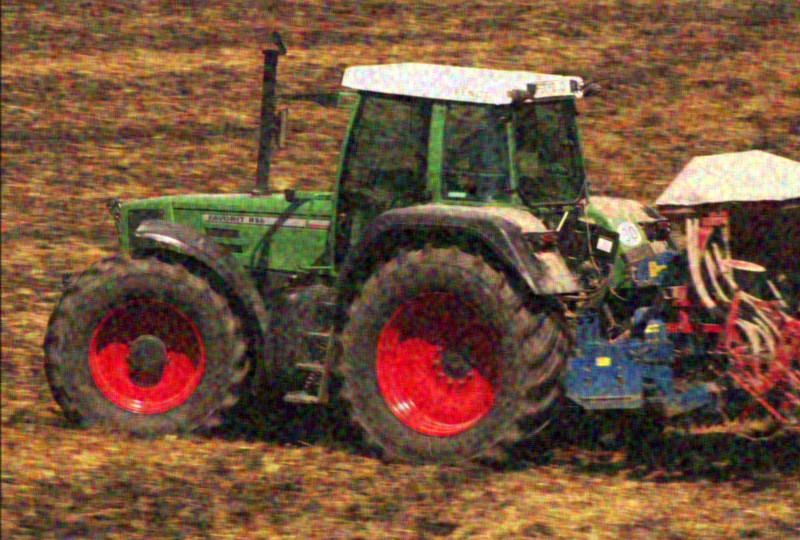}{Box 65 5}%
		\includegraphics[width=\imagesize,trim={0 0 0 0 },clip]{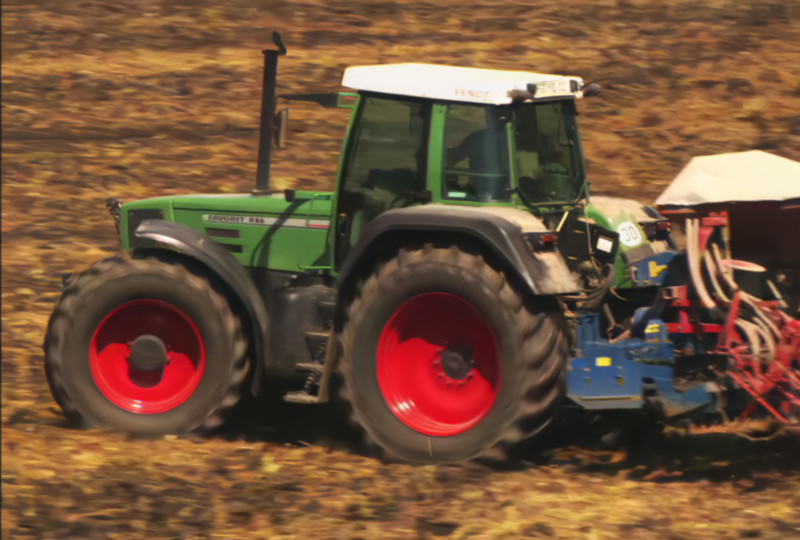}%
		\includegraphics[width=\imagesize,trim={0 0 0 0 },clip]{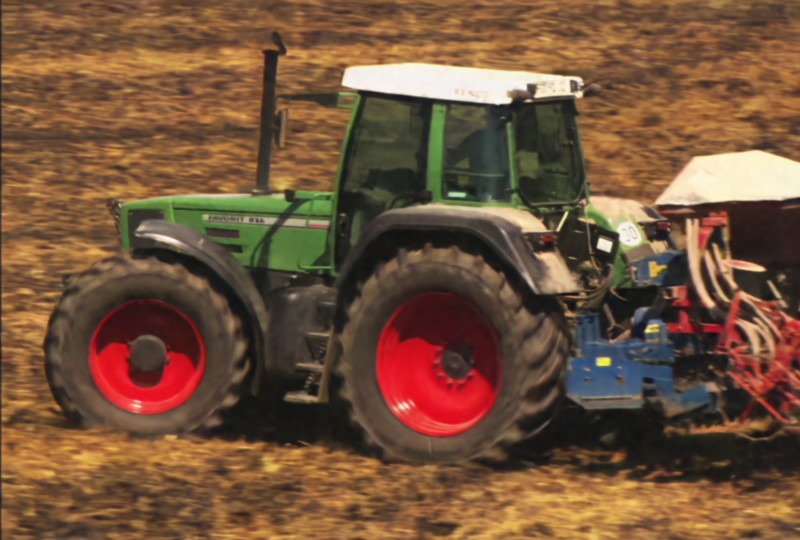}%
		\includegraphics[width=\imagesize,trim={0 0 0 0 },clip]{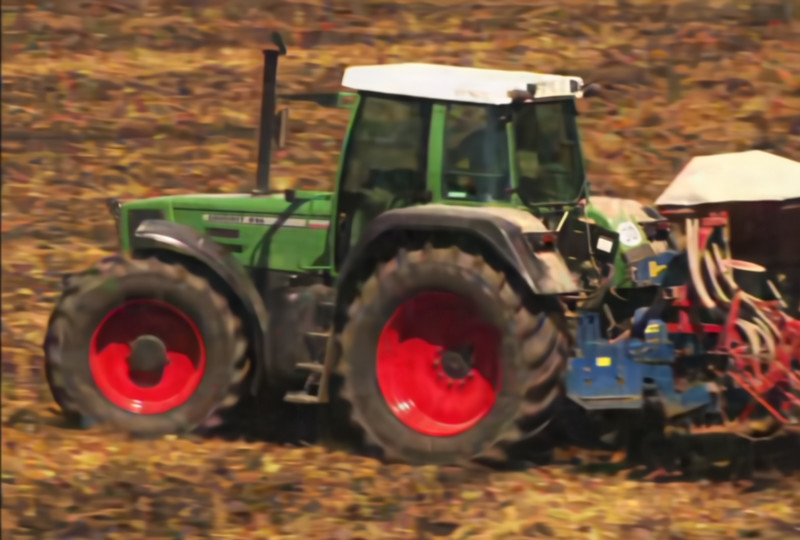}%
		
		\overimg[width=\imagesize,trim={0 0 0 0 },clip]{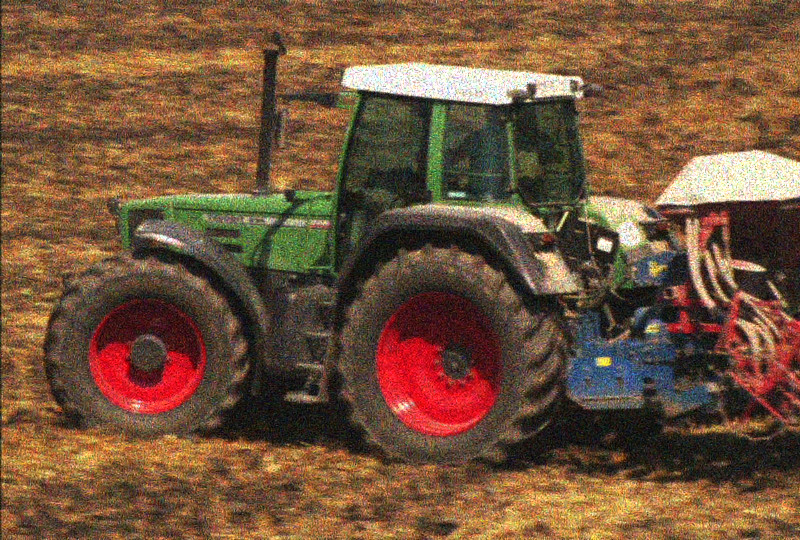}{Demosaicked 4}%
		\includegraphics[width=\imagesize,trim={0 0 0 0 },clip]{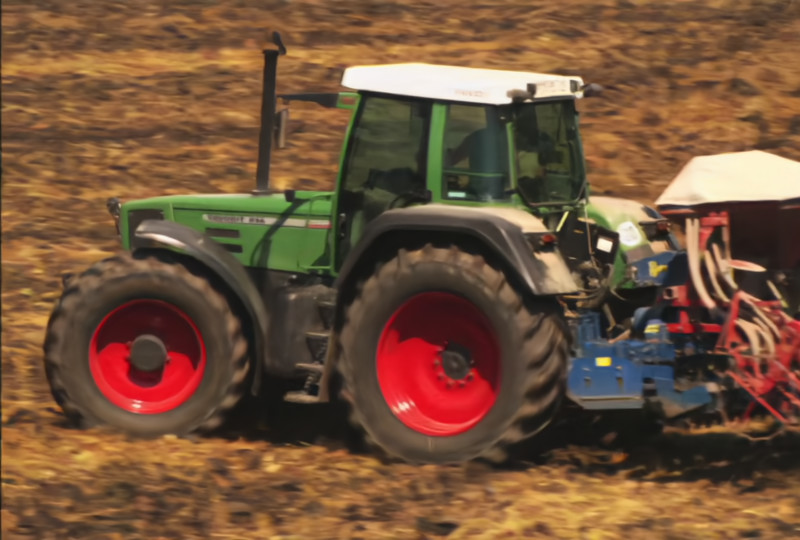}%
		\includegraphics[width=\imagesize,trim={0 0 0 0 },clip]{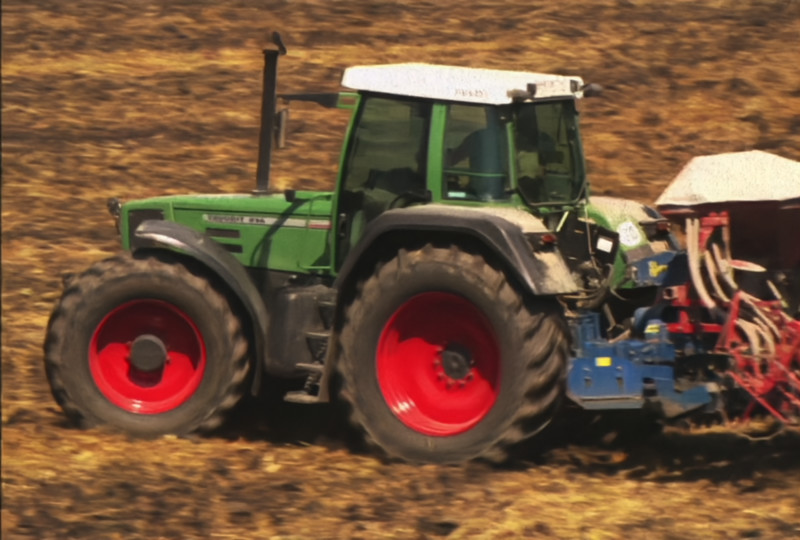}%
		\includegraphics[width=\imagesize,trim={0 0 0 0 },clip]{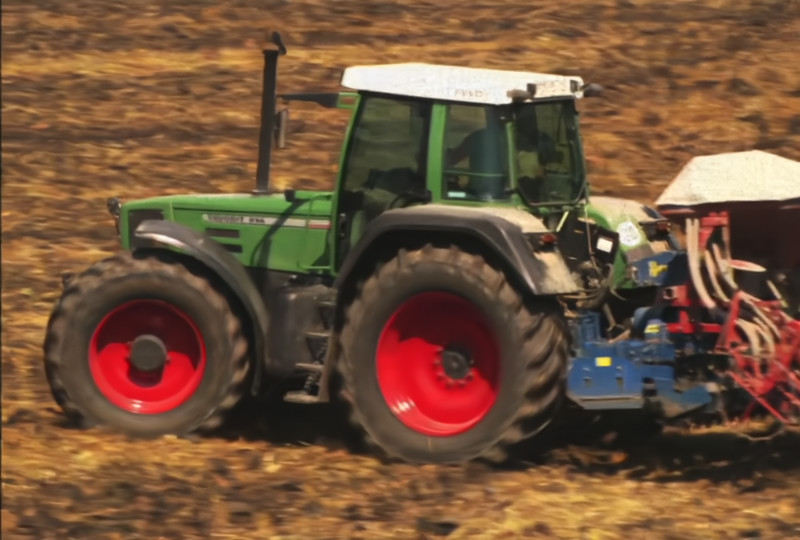}%
				
	\caption{Comparison on all synthetic noise types. From left to right: the noisy image, the result of the noise-specific  FastDVDnet (\textit{supervised}), the result of our offline MF2F fine-tuning (\textit{self-supervised}) and the per-level variance map MF2F (\textit{self-supervised}). From the top to the bottom: AWGN20, AWGN40, Poisson1, Poisson8, box noise $3 \times 3, \sigma=40$, box noise $5\times5, \sigma=65$ and the demosaicking noise.}
	\label{fig:big_fig_all_noises_tractor}
	\end{center}
\end{figure*}

\begin{figure*}
	\begin{center}
		\def\imagesize{0.25\textwidth}
		\overimg[width=\imagesize,trim={0 0 0 0 },clip]{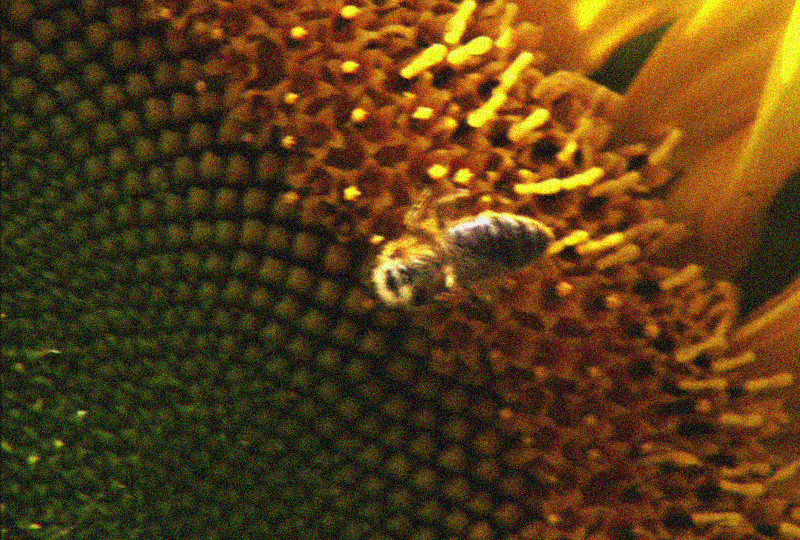}{Gaussian 20}%
		\includegraphics[width=\imagesize,trim={0 0 0 0 },clip]{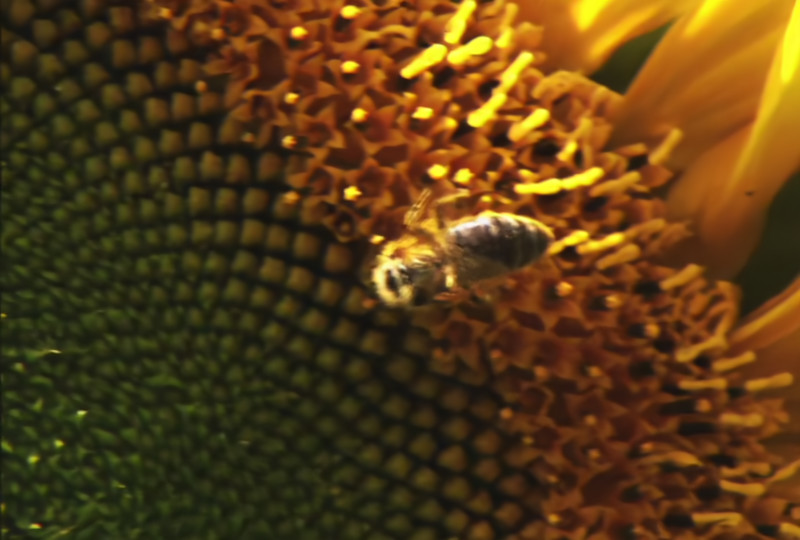}%
		\includegraphics[width=\imagesize,trim={0 0 0 0 },clip]{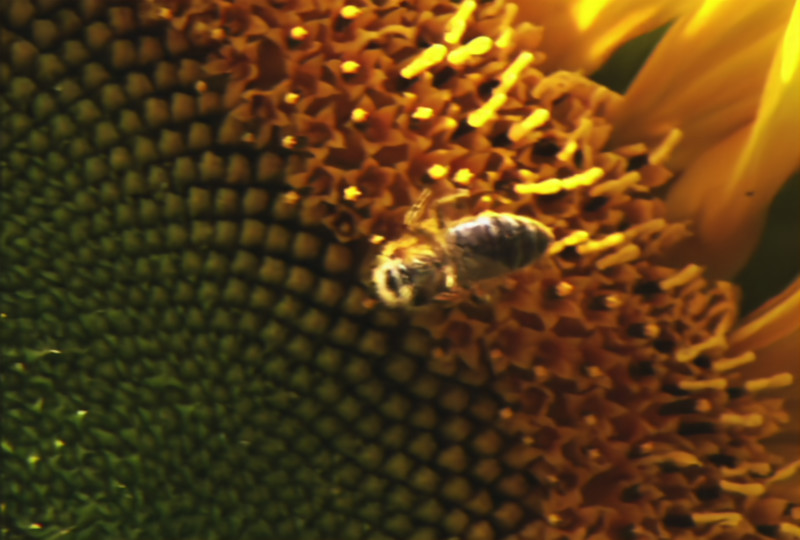}%
		\includegraphics[width=\imagesize,trim={0 0 0 0 },clip]{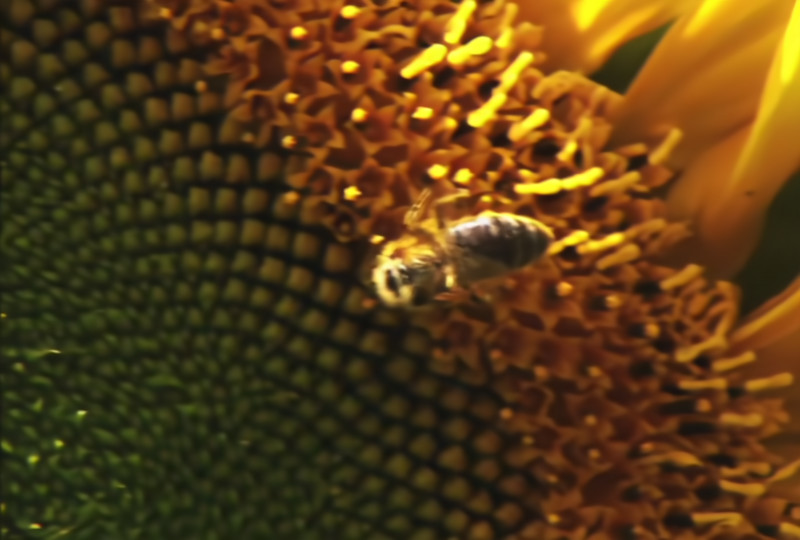}%
				
		\overimg[width=\imagesize,trim={0 0 0 0 },clip]{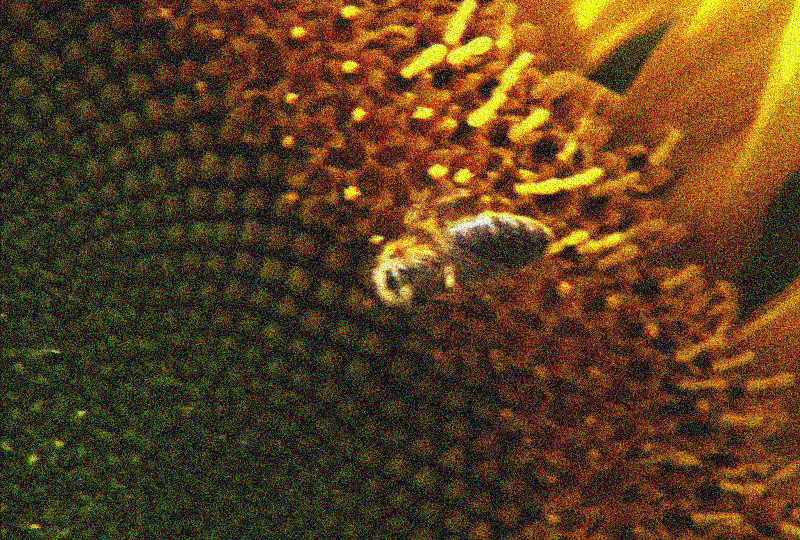}{Gaussian 40}%
		\includegraphics[width=\imagesize,trim={0 0 0 0 },clip]{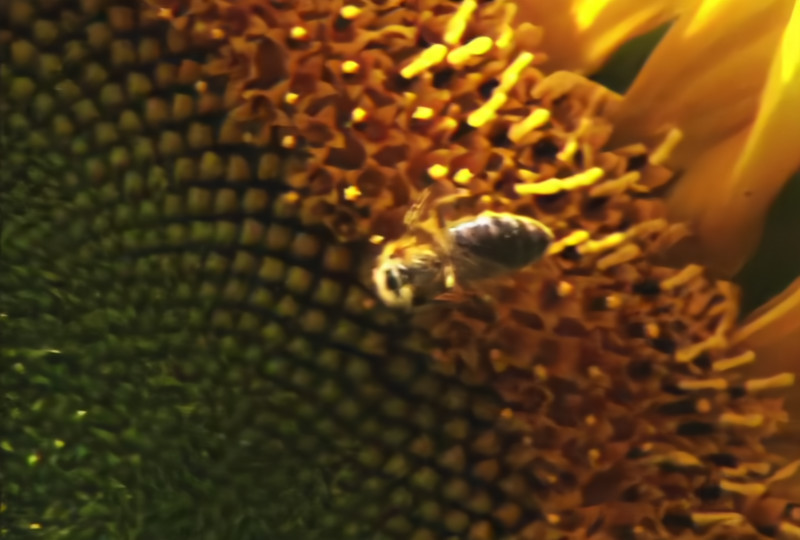}%
		\includegraphics[width=\imagesize,trim={0 0 0 0 },clip]{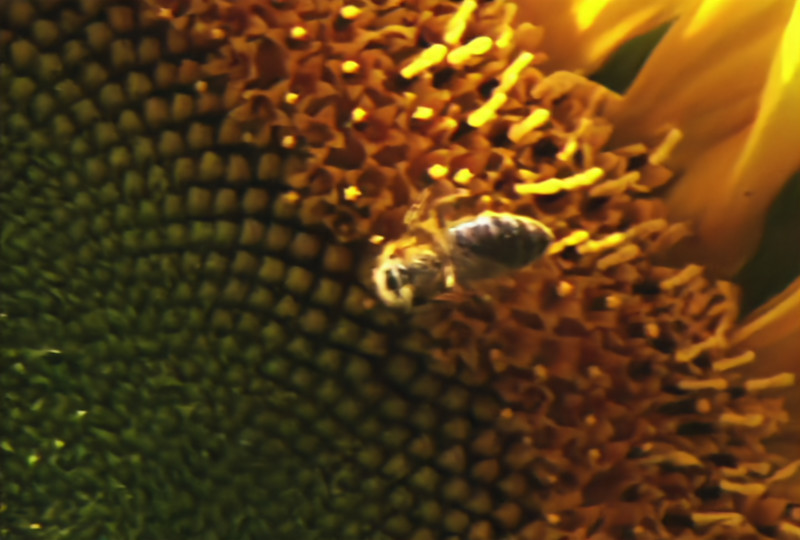}%
		\includegraphics[width=\imagesize,trim={0 0 0 0 },clip]{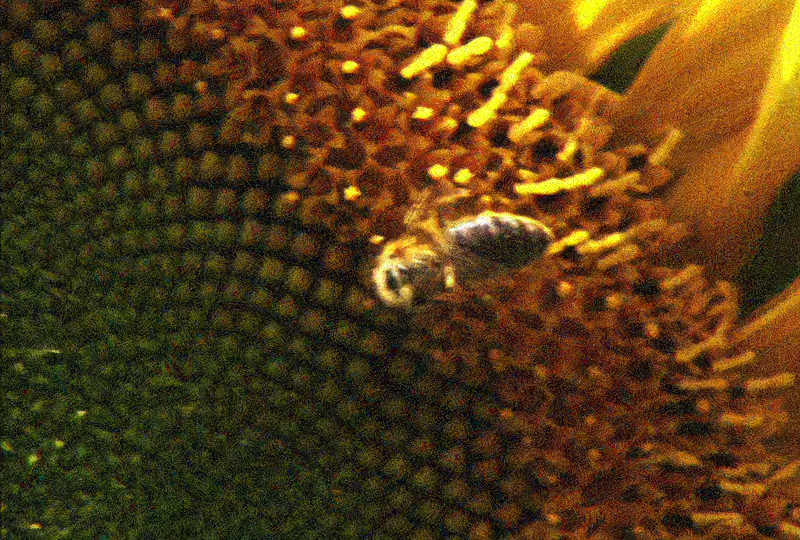}%
				
		\overimg[width=\imagesize,trim={0 0 0 0 },clip]{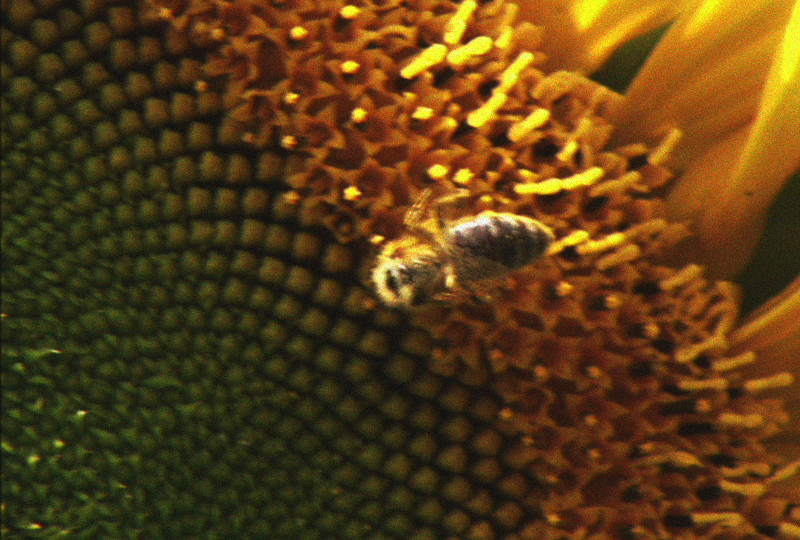}{Poisson 1}%
		\includegraphics[width=\imagesize,trim={0 0 0 0 },clip]{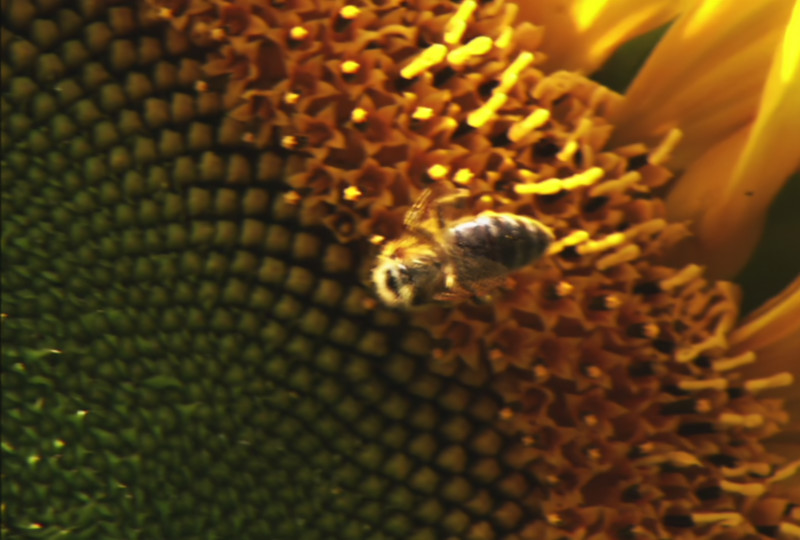}%
		\includegraphics[width=\imagesize,trim={0 0 0 0 },clip]{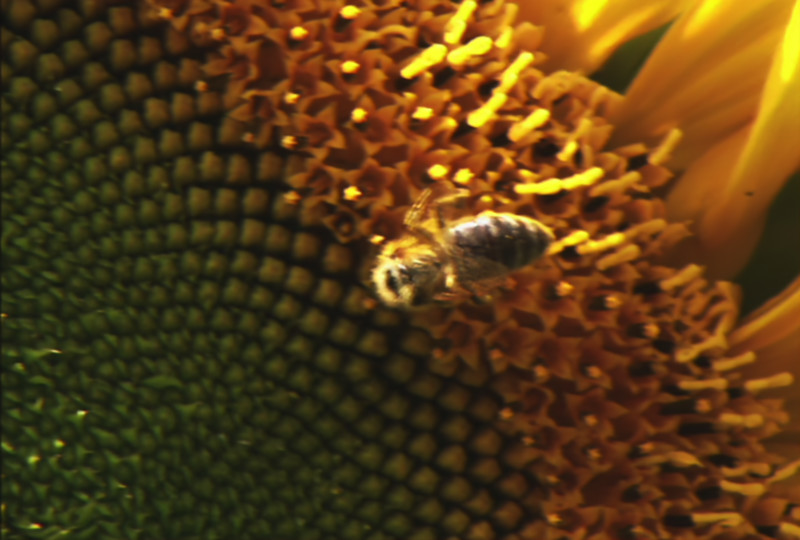}%
		\includegraphics[width=\imagesize,trim={0 0 0 0 },clip]{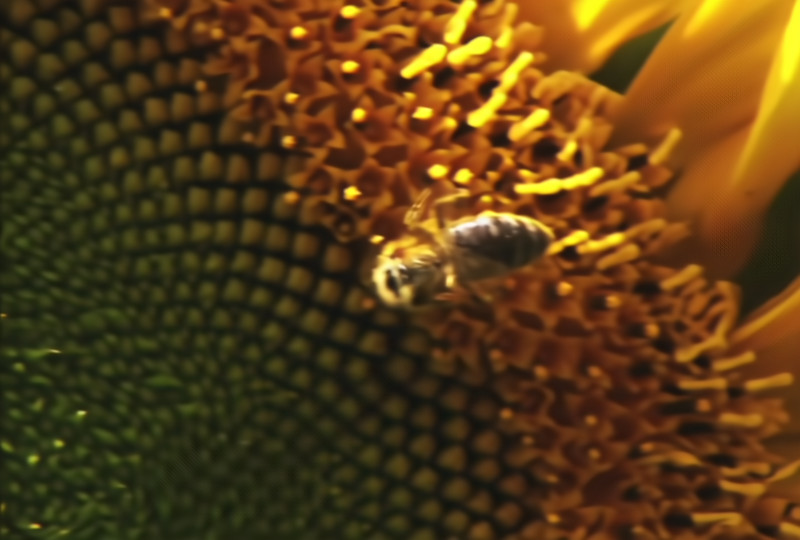}%
					
		\overimg[width=\imagesize,trim={0 0 0 0 },clip]{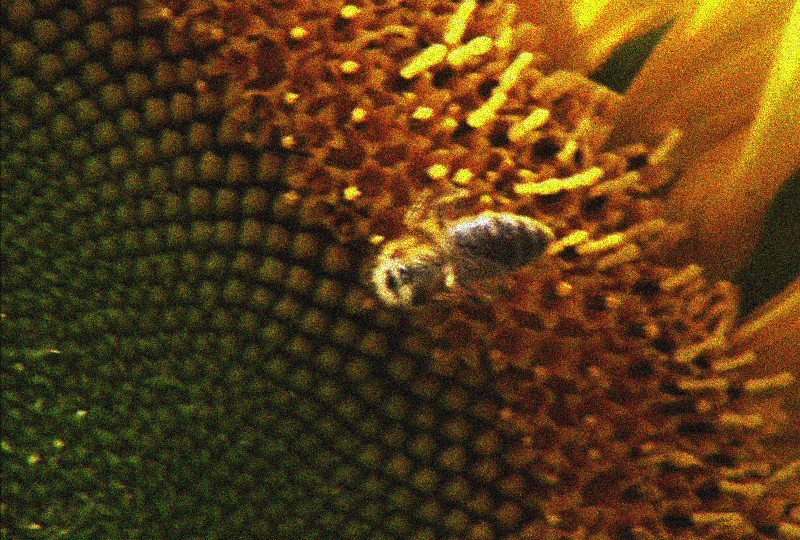}{Poisson 8}%
		\includegraphics[width=\imagesize,trim={0 0 0 0 },clip]{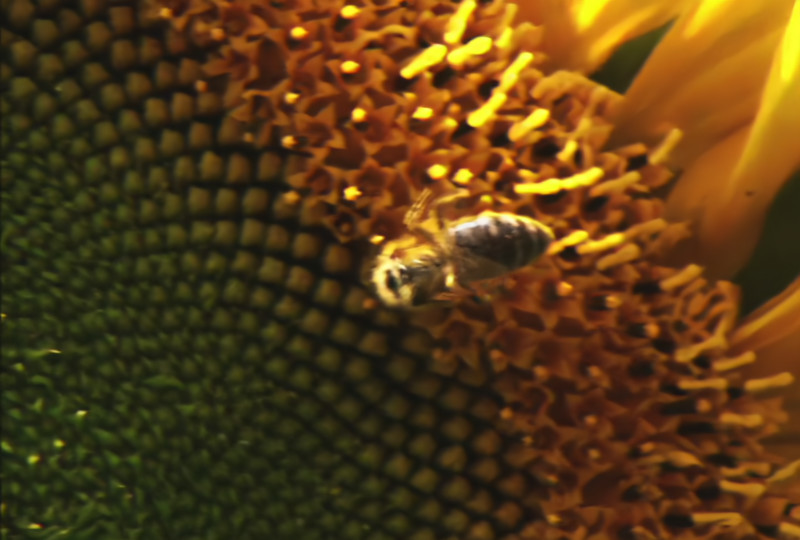}%
		\includegraphics[width=\imagesize,trim={0 0 0 0 },clip]{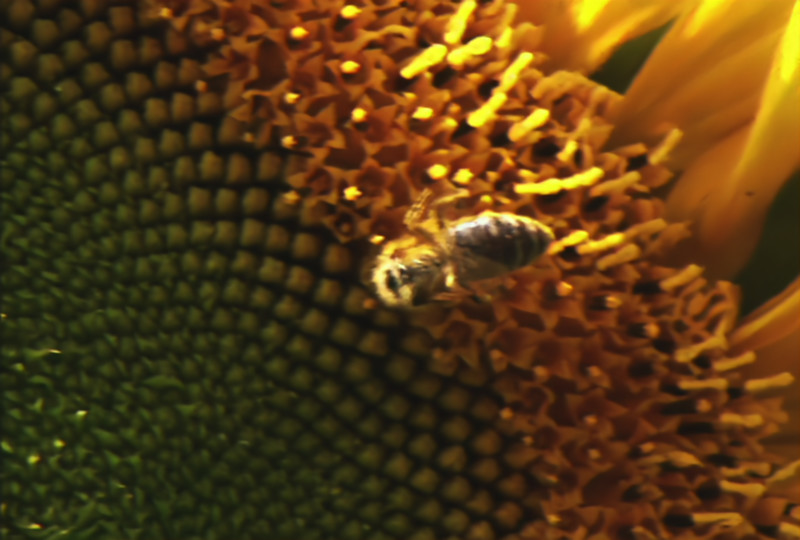}%
		\includegraphics[width=\imagesize,trim={0 0 0 0 },clip]{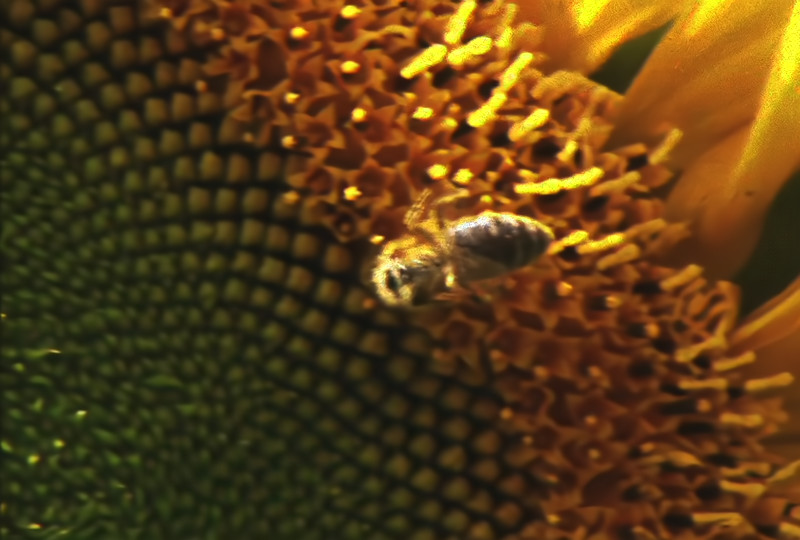}%
				
		\overimg[width=\imagesize,trim={0 0 0 0 },clip]{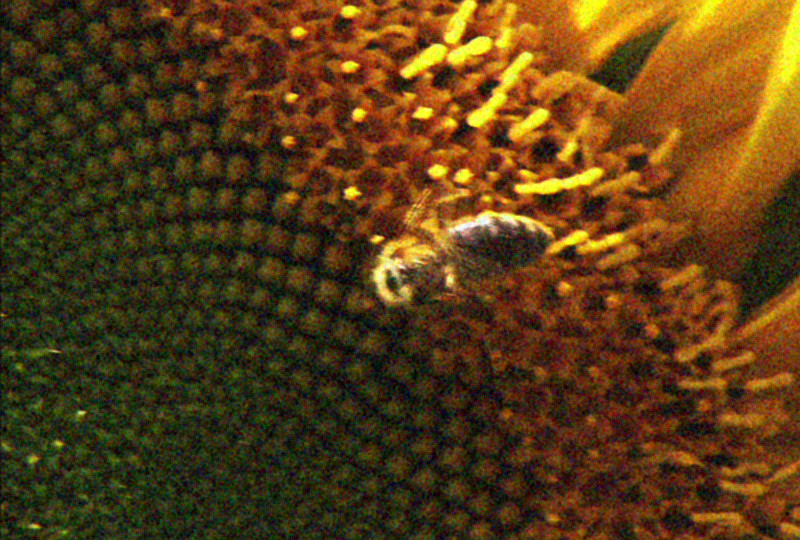}{Box 40 3}%
		\includegraphics[width=\imagesize,trim={0 0 0 0 },clip]{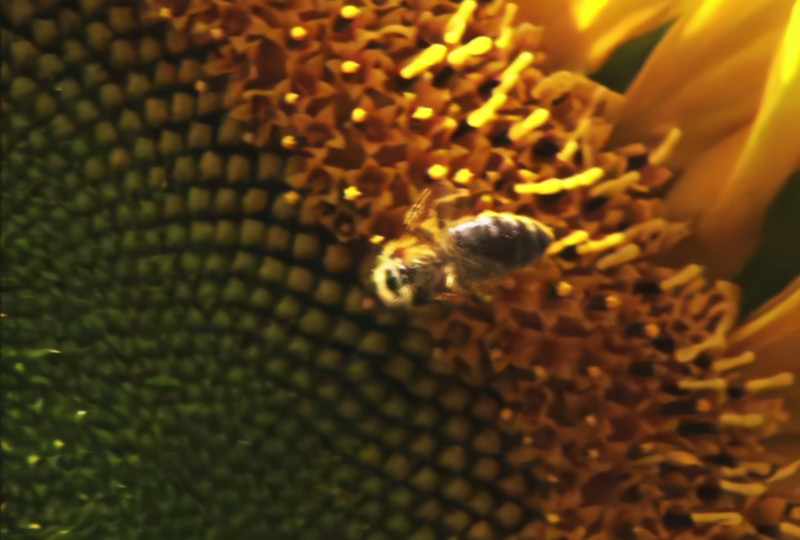}%
		\includegraphics[width=\imagesize,trim={0 0 0 0 },clip]{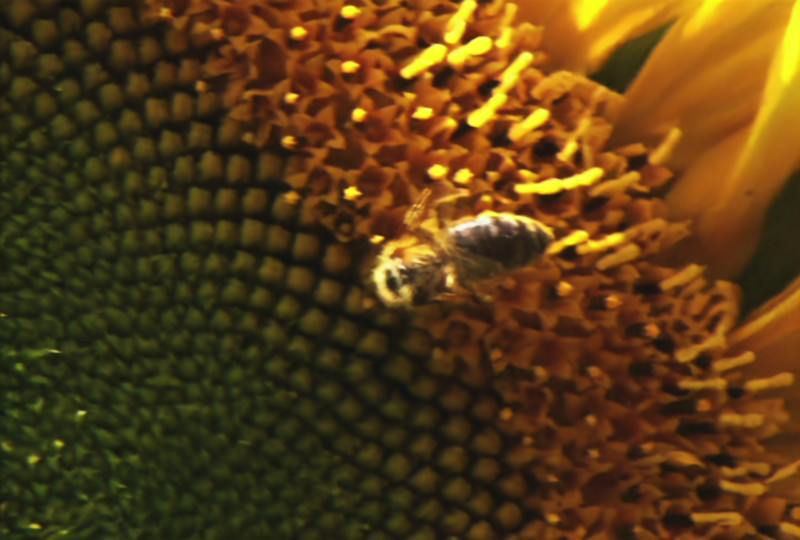}%
		\includegraphics[width=\imagesize,trim={0 0 0 0 },clip]{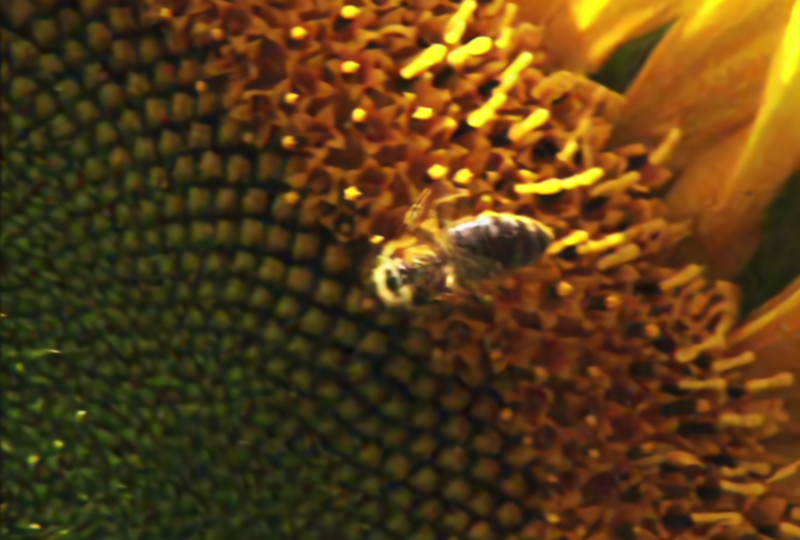}%
					
		\overimg[width=\imagesize,trim={0 0 0 0 },clip]{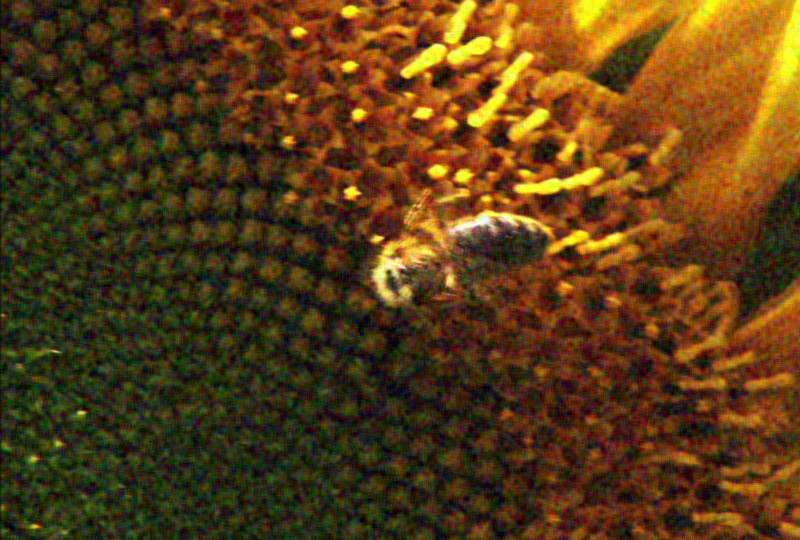}{Box 65 5}%
		\includegraphics[width=\imagesize,trim={0 0 0 0 },clip]{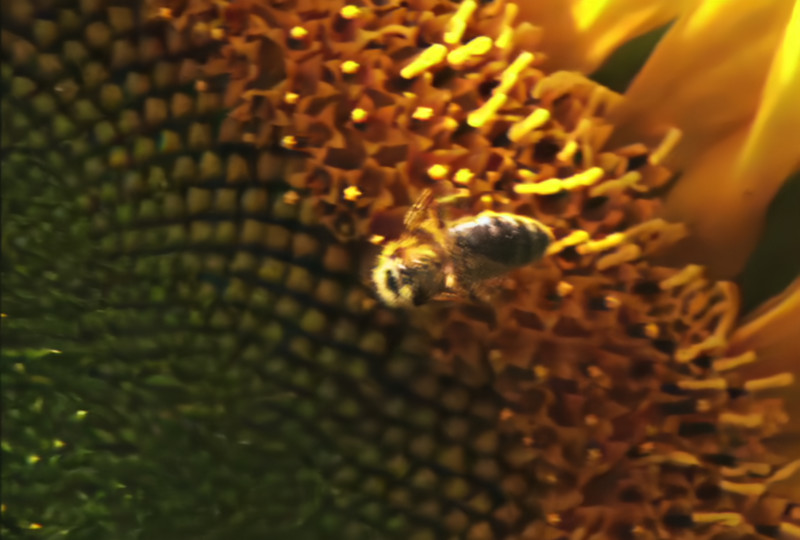}%
		\includegraphics[width=\imagesize,trim={0 0 0 0 },clip]{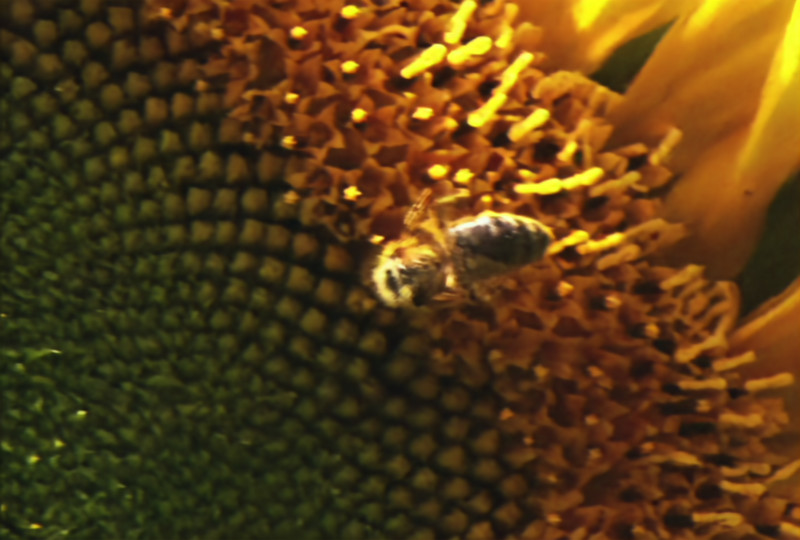}%
		\includegraphics[width=\imagesize,trim={0 0 0 0 },clip]{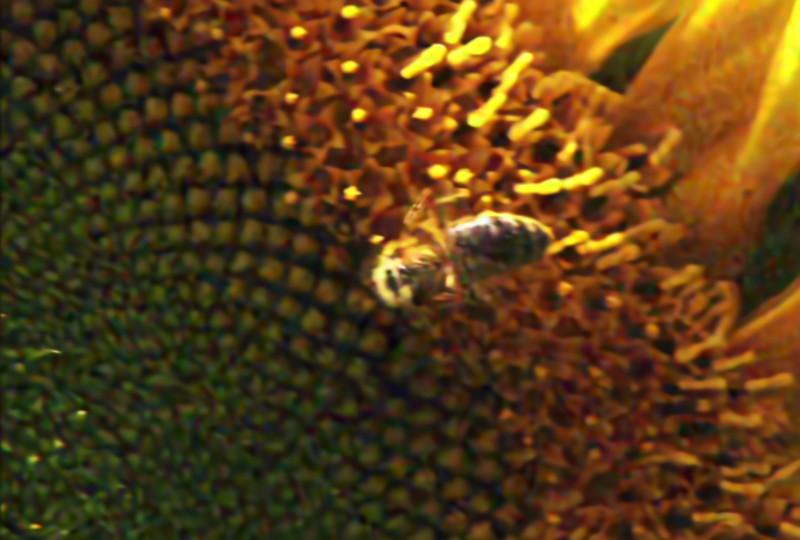}%
		
		\overimg[width=\imagesize,trim={0 0 0 0 },clip]{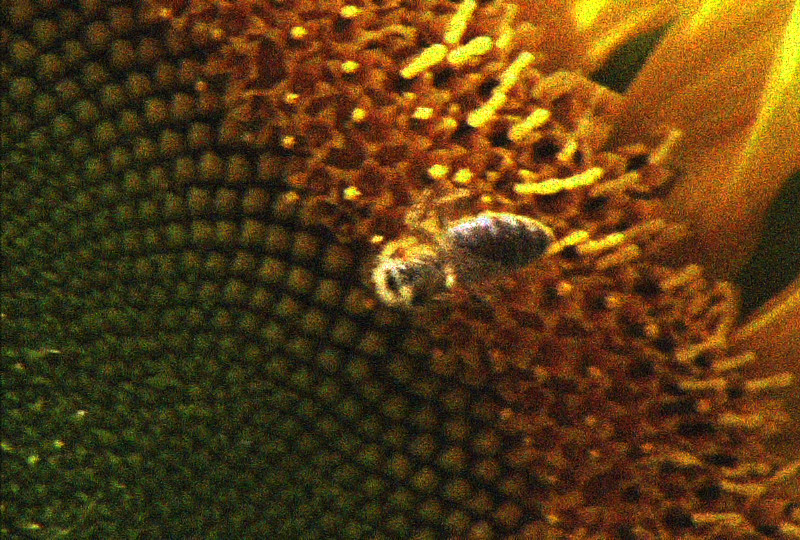}{Demosaicked 4}%
		\includegraphics[width=\imagesize,trim={0 0 0 0 },clip]{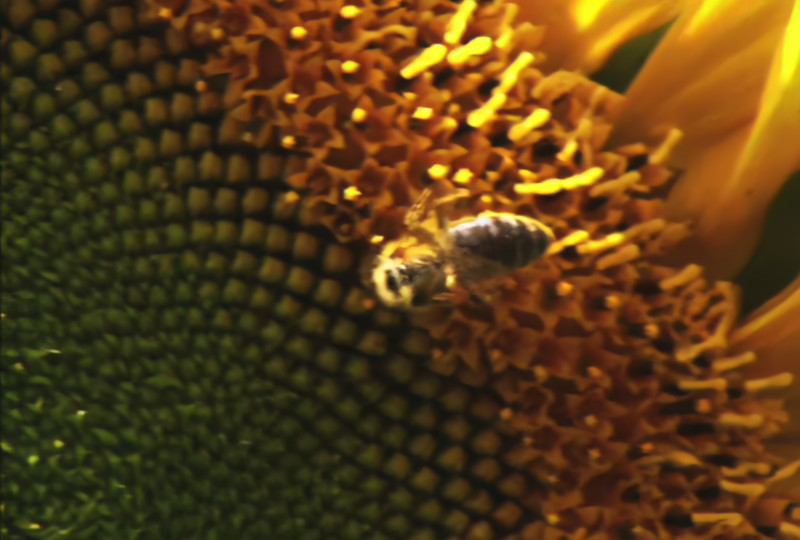}%
		\includegraphics[width=\imagesize,trim={0 0 0 0 },clip]{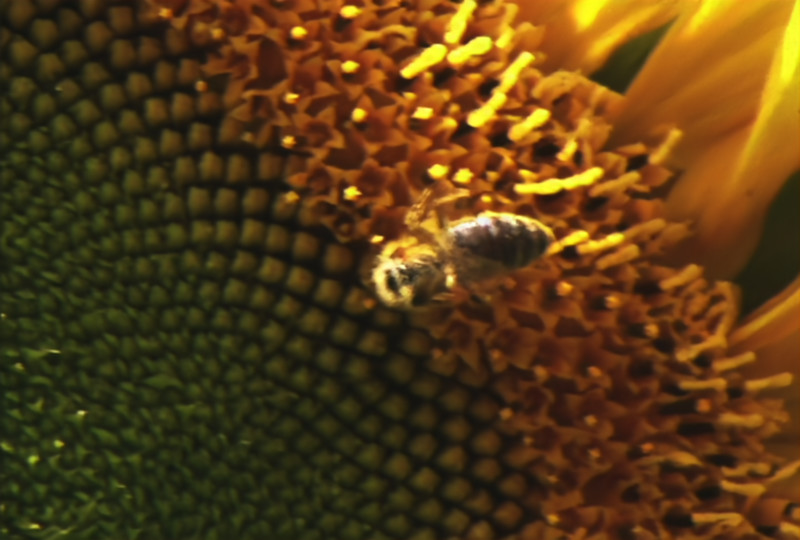}%
		\includegraphics[width=\imagesize,trim={0 0 0 0 },clip]{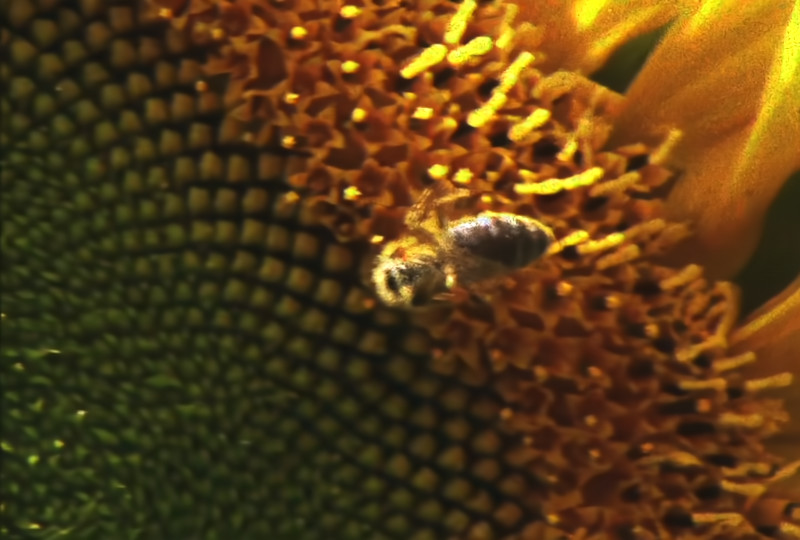}%
				
	\caption{Comparison on synthetic noise types. From left to right: the noisy image, the result of the noise-specific FastDVDnet (\textit{supervised}), the result of our offline MF2F fine-tuning (\textit{self-supervised}) and the per-level variance map MF2F (\textit{self-supervised}). From the top to the bottom: AWGN20, AWGN40, Poisson1, Poisson8, box noise $3 \times 3, \sigma=40$, box noise $5\times5, \sigma=65$ and the demosaicking noise.}
	\label{fig:big_fig_all_noises_sunflower}
	\end{center}
\end{figure*}

Figures \ref{fig:big_fig_all_noises_tractor} and \ref{fig:big_fig_all_noises_sunflower} illustrate the results for all the synthetic noise types used in the table 2 in the main paper, for two video sequences. In all the cases the starting point were the weights  pre-trained  for AWGN25.  The results of our offline MF2F attains the performance of the noise-specific FastDVDnet trained in a supervised settings. We also display the results when fine-tuning the per-level variance map. From the PSNR and SSIM tables (see the main paper) we can  see  that for the AWGN and Poisson noise this fine-tuning yields results similar to the noise-specific FastDVDnet trained in a supervised settings. For the box noise and the demosaicked noise, the  performance of  MF2F  is slightly below the result of the noise-specific FastDVDnet. Yet, we can see from this figure that qualitatively the results are comparable.

\begin{figure*}
	\begin{center}
		\def\imagesize{0.33\textwidth}
		\overimg[width=\imagesize,trim={0 0 0 0 },clip]{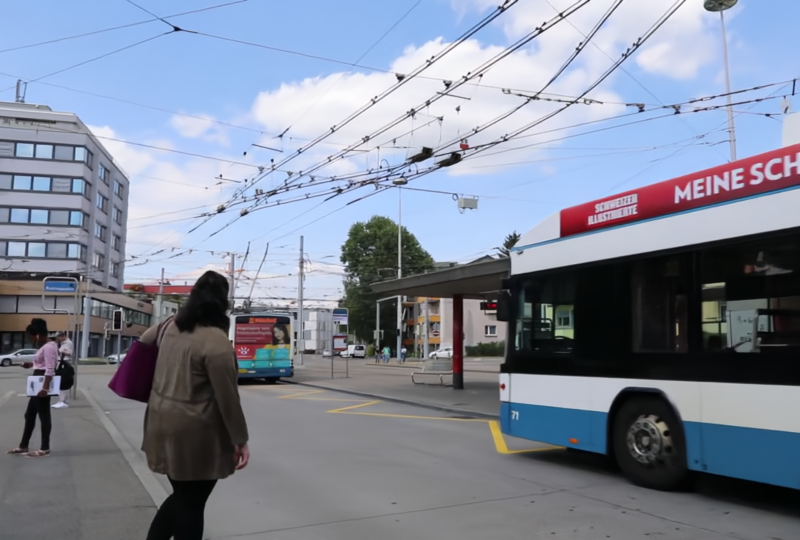}{37.69dB}%
        \overimg[width=\imagesize,trim={0 0 0 0 },clip]{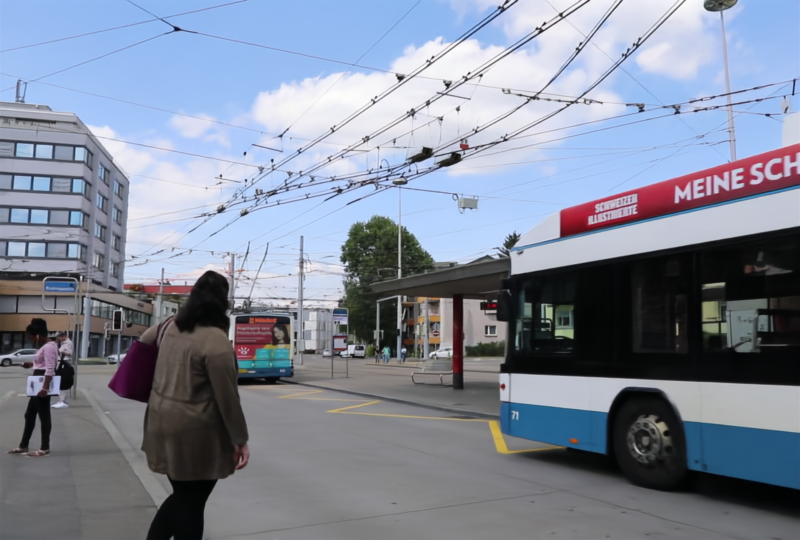}{37.91dB}%
        \overimg[width=\imagesize,trim={0 0 0 0 },clip]{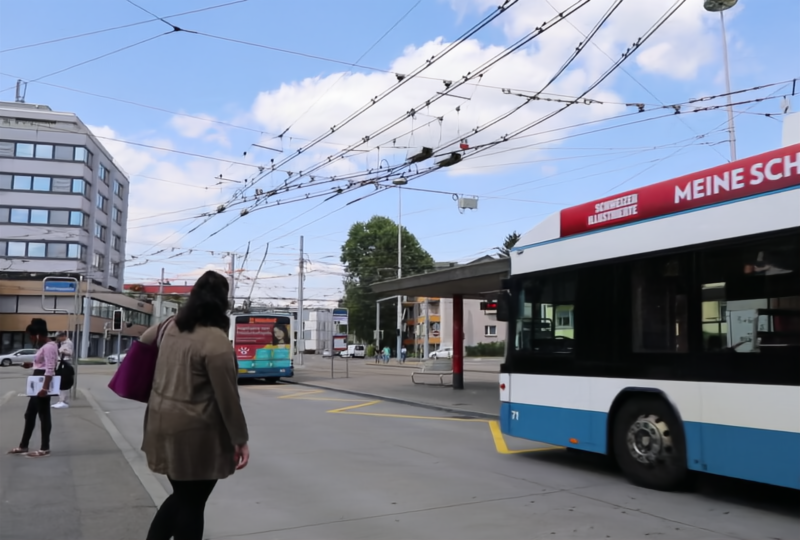}{38.02dB}\\
        \vspace{0.05cm}
        \overimg[width=\imagesize,trim={0 0 0 0 },clip]{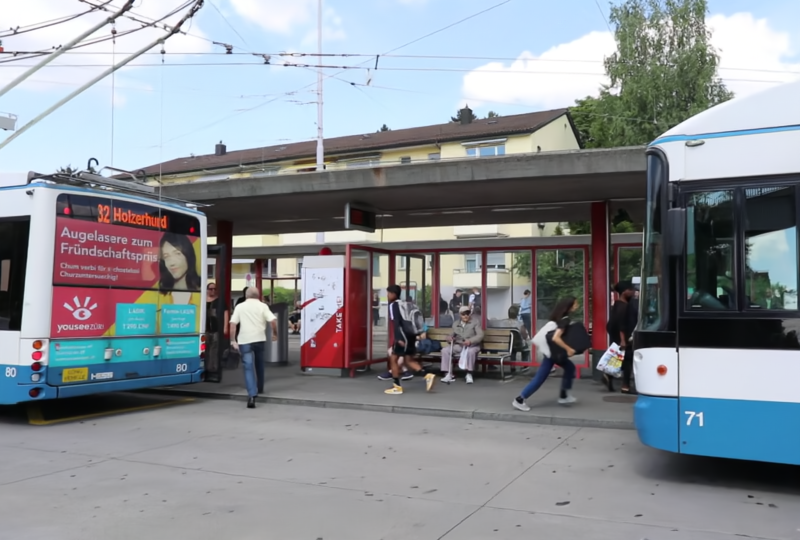}{37.63dB}%
        \overimg[width=\imagesize,trim={0 0 0 0 },clip]{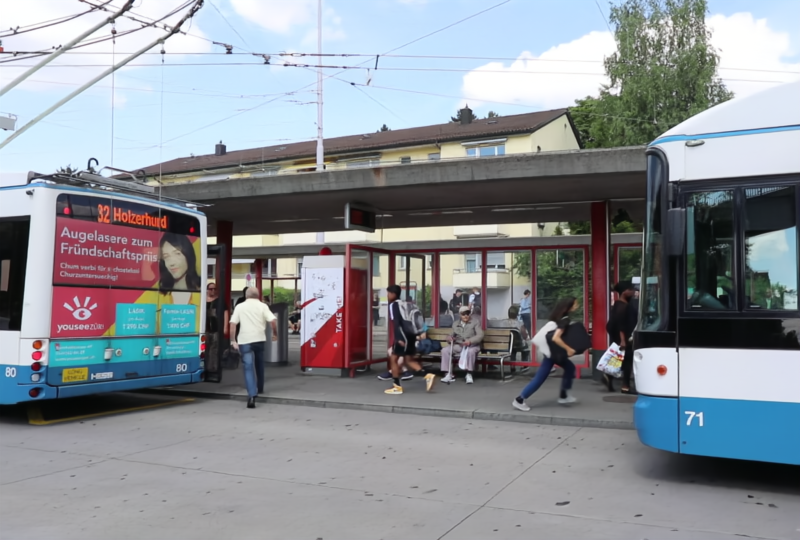}{37.92dB}%
        \overimg[width=\imagesize,trim={0 0 0 0 },clip]{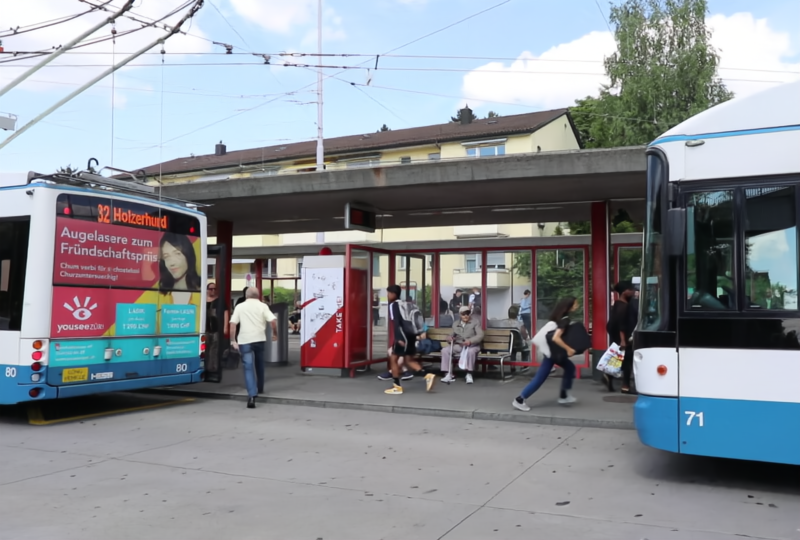}{37.92dB}\\
        \vspace{0.05cm}
        \overimg[width=\imagesize,trim={0 0 0 0 },clip]{fig/online_vs_offline/FastDVDnet-gaussian20-114.png}{36.93dB}%
        \overimg[width=\imagesize,trim={0 0 0 0 },clip]{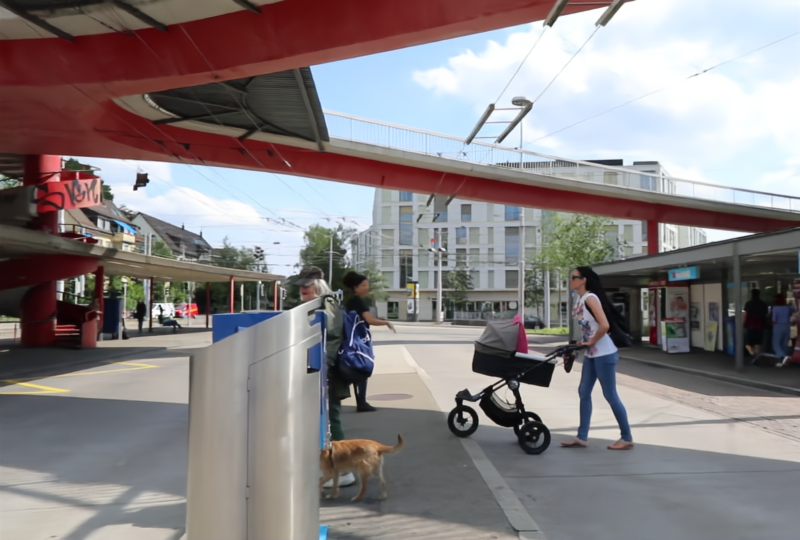}{37.22dB}%
        \overimg[width=\imagesize,trim={0 0 0 0 },clip]{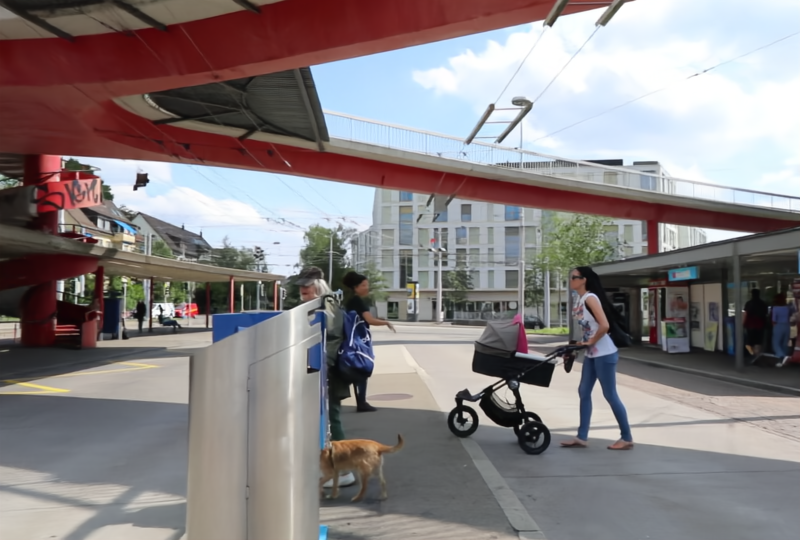}{37.18dB}%
	\caption{Comparison of results obtained with the online and offline MF2F (both self-supervised) on Gaussian 20. From left to right: noise-specific FastDVDnet (\textit{supervised}), online MF2F (\textit{self-supervised}) and offline MF2F (\textit{self-supervised})}
	\label{fig:results_MF2F}
	\end{center}
\end{figure*}

\begin{figure*}
	\begin{center}
		\def\imagesize{0.33\textwidth}
		\overimg[width=\imagesize]{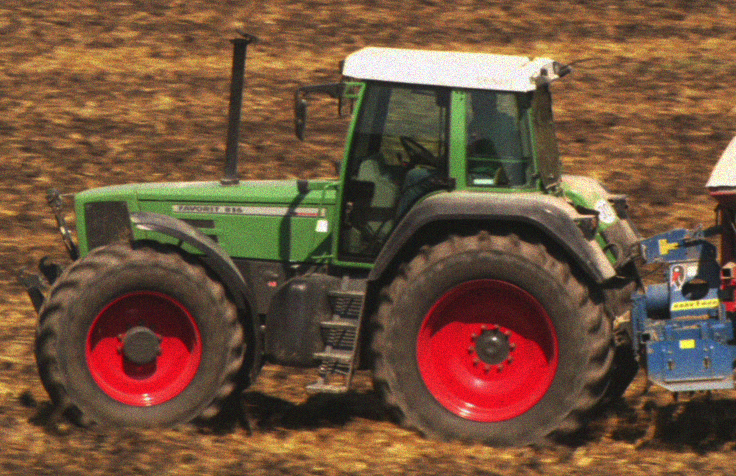}{28.67dB}%
		\overimg[width=\imagesize]{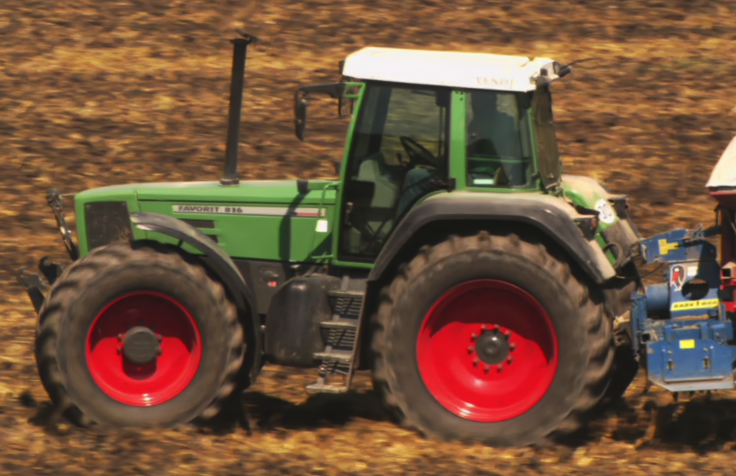}{38.37dB}%
    	\overimg[width=\imagesize]{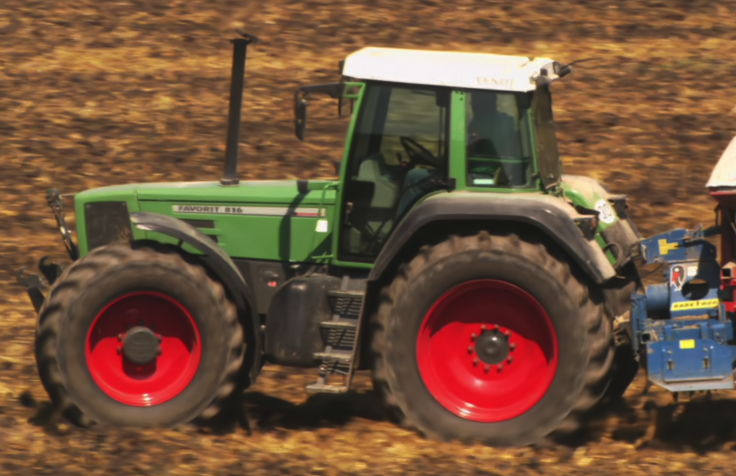}{37.71dB}\\
    	\vspace{0.05cm}
		\overimg[width=\imagesize]{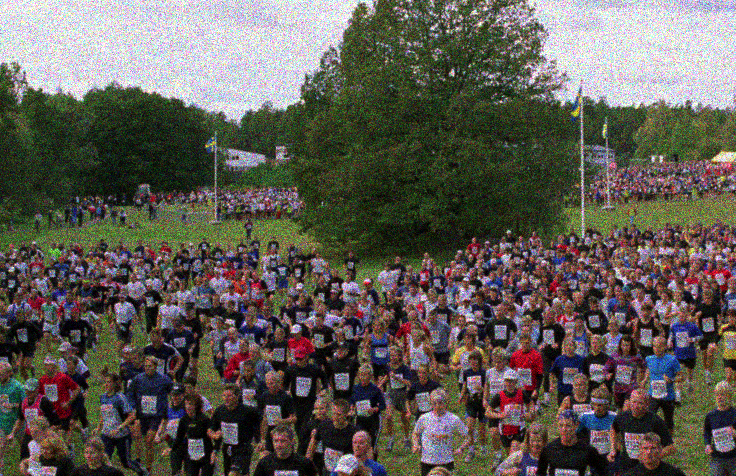}{19.13dB}%
		\overimg[width=\imagesize]{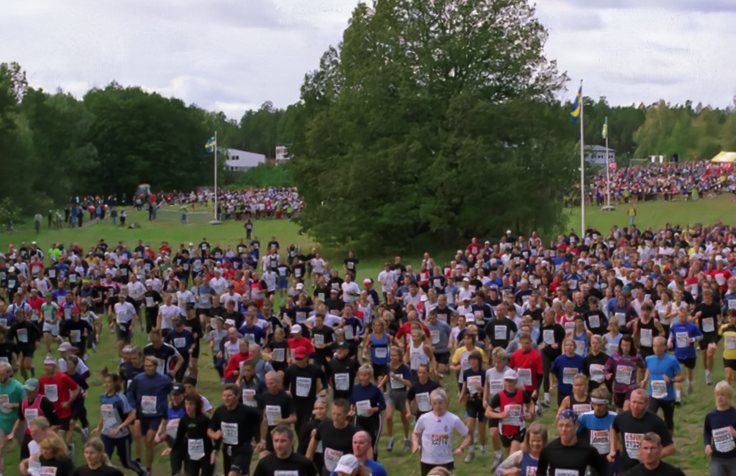}{31.91dB}%
    	\overimg[width=\imagesize]{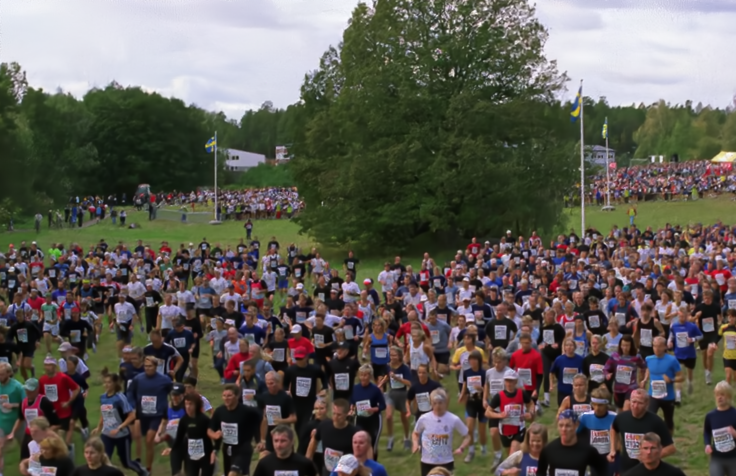}{30.96dB}%
	\caption{Comparison of results obtained with the per-level and the spatially variant variance map on poisson noise $p=1$ (first row) and $p=8$ (second row). From left to right: noisy, per-level variance map and spatially variant variance map.}
	\label{fig:comparison_spatial_variance_map}
	\end{center}
\end{figure*}

Additional results obtained with the proposed fine-tuning for both online and offline versions are shown in Figure~\ref{fig:results_MF2F}. Those results are compared with the ones obtained by evaluation of the noise-specific FastDVDnet trained in a supervised settings in case of AWGN20. It shows that both the online and offline method achieve the performance of the supervised network and surpasses it.

\begin{figure*}
	\begin{center}
		\def\imagesize{0.33\textwidth}
		\includegraphics[width=\imagesize]{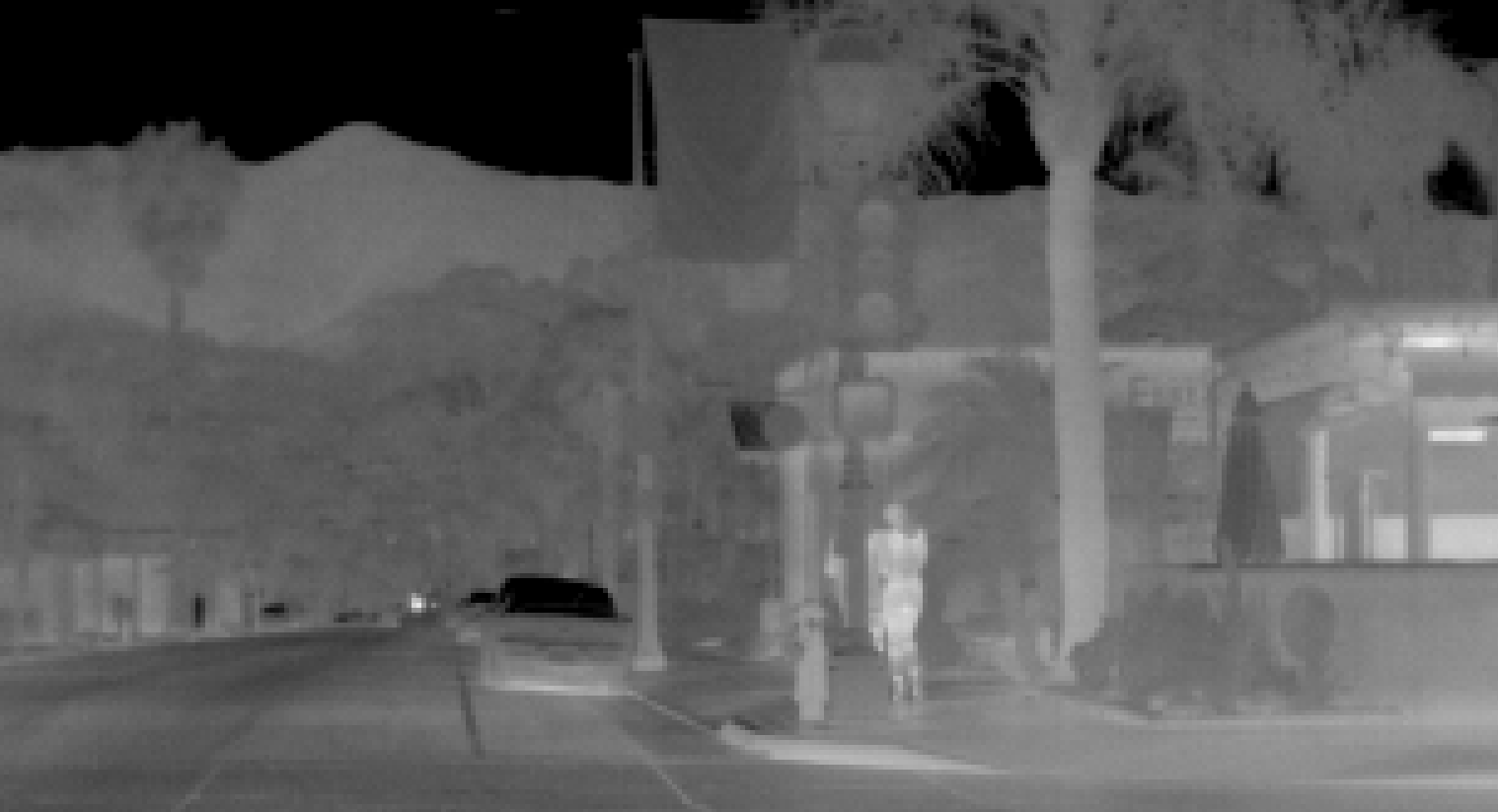}%
		\includegraphics[width=\imagesize]{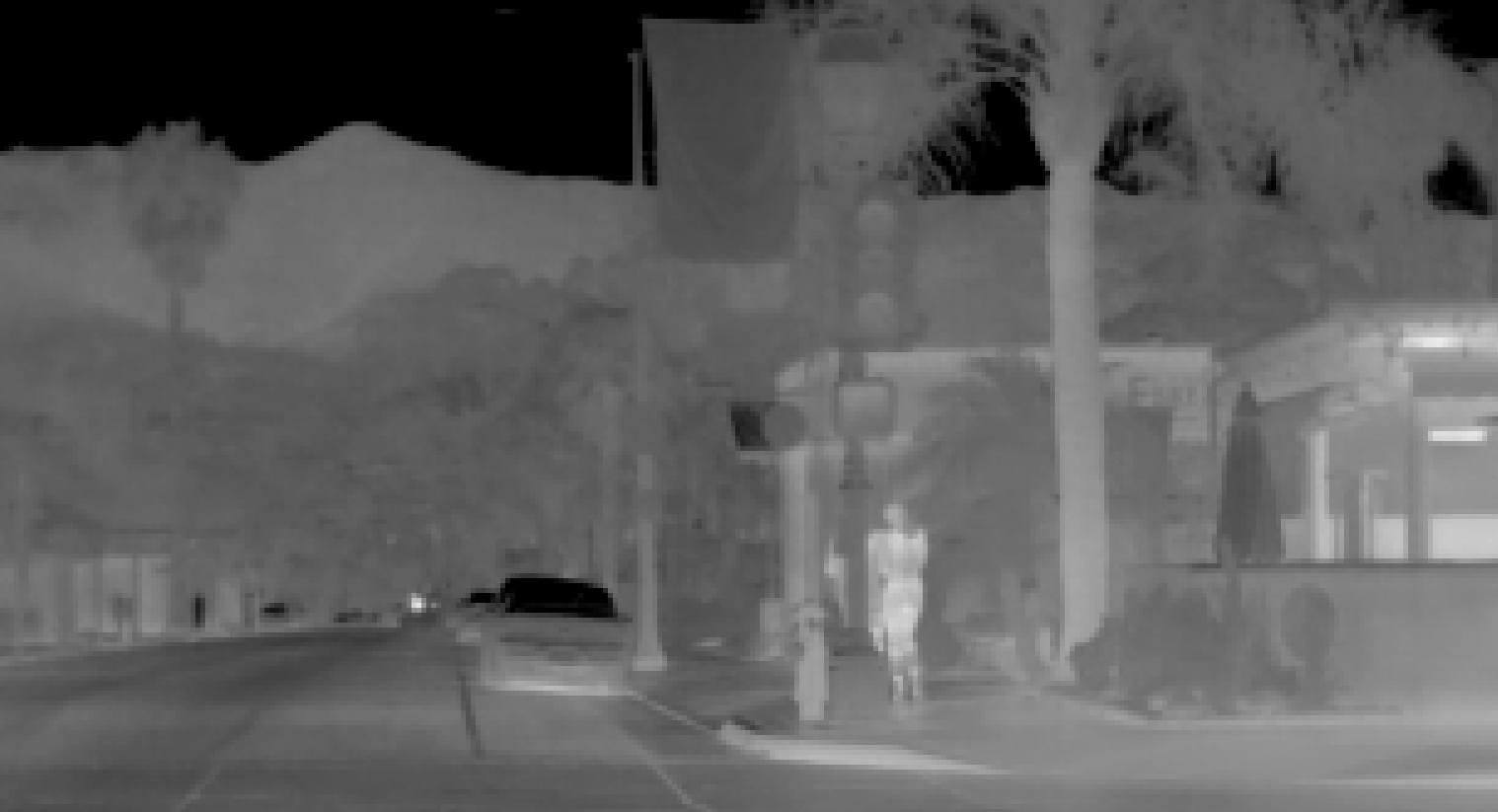}%
		\includegraphics[width=\imagesize]{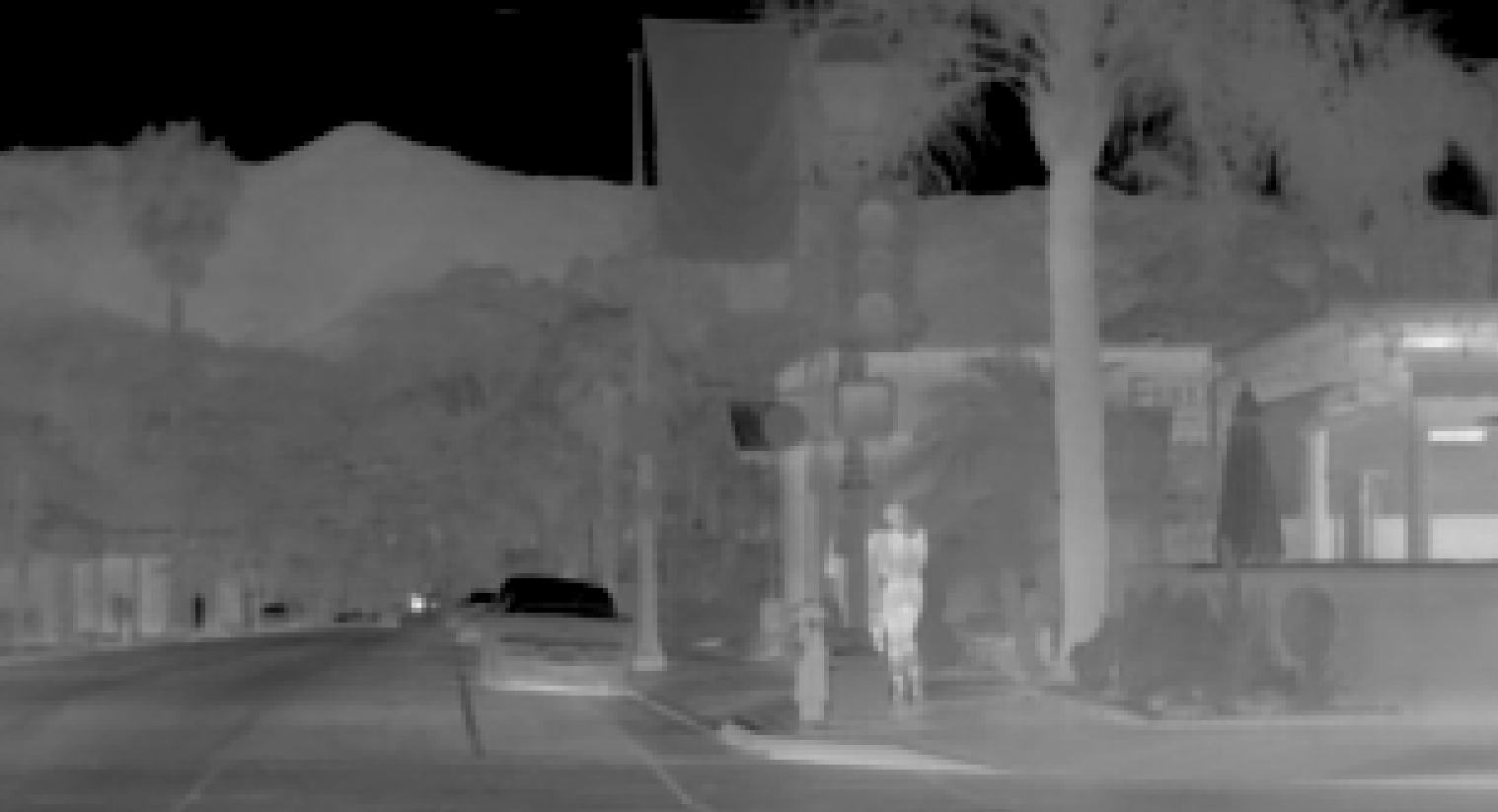}\\
		\vspace{0.07cm}
		\includegraphics[width=\imagesize]{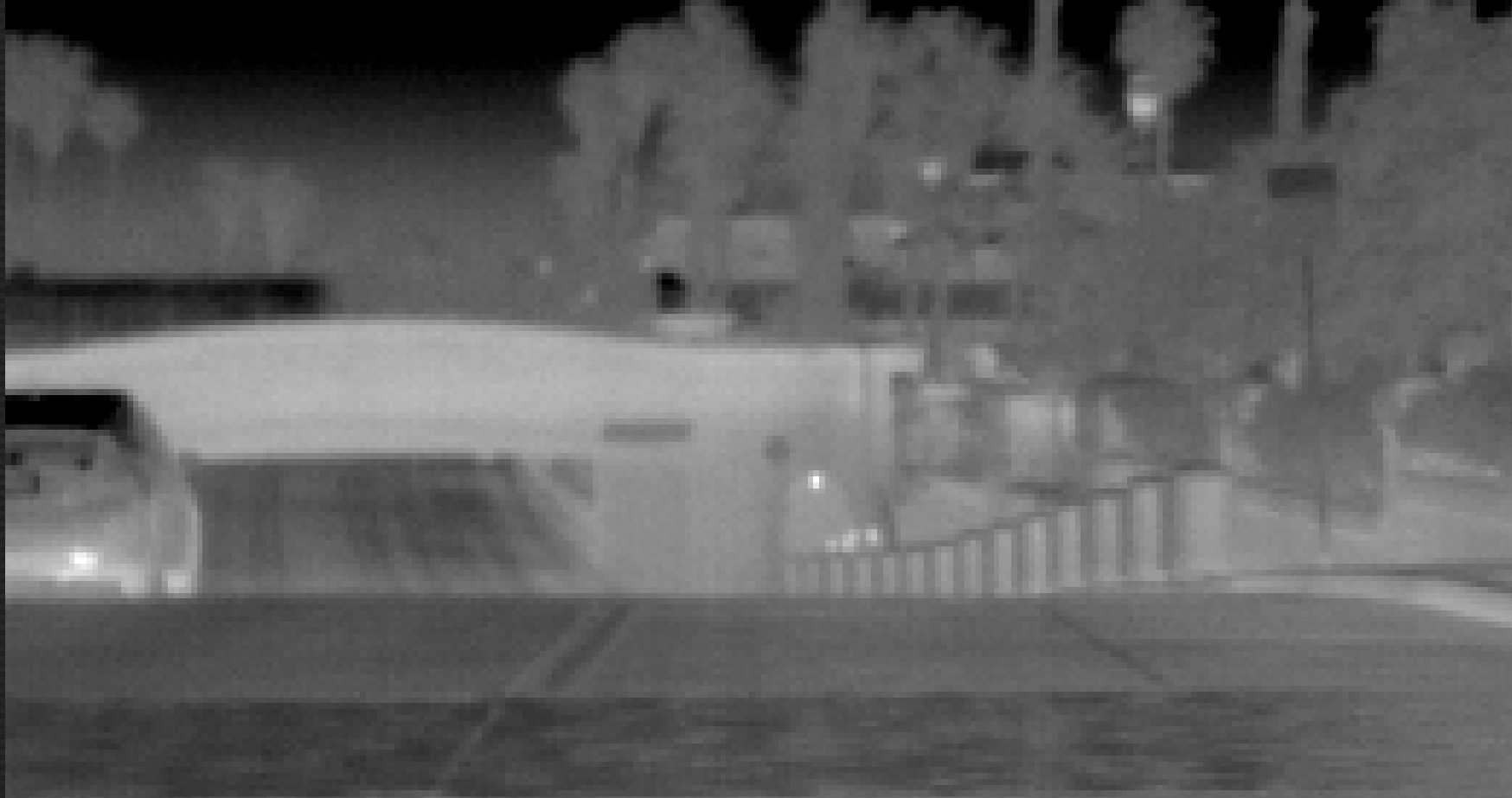}%
		\includegraphics[width=\imagesize]{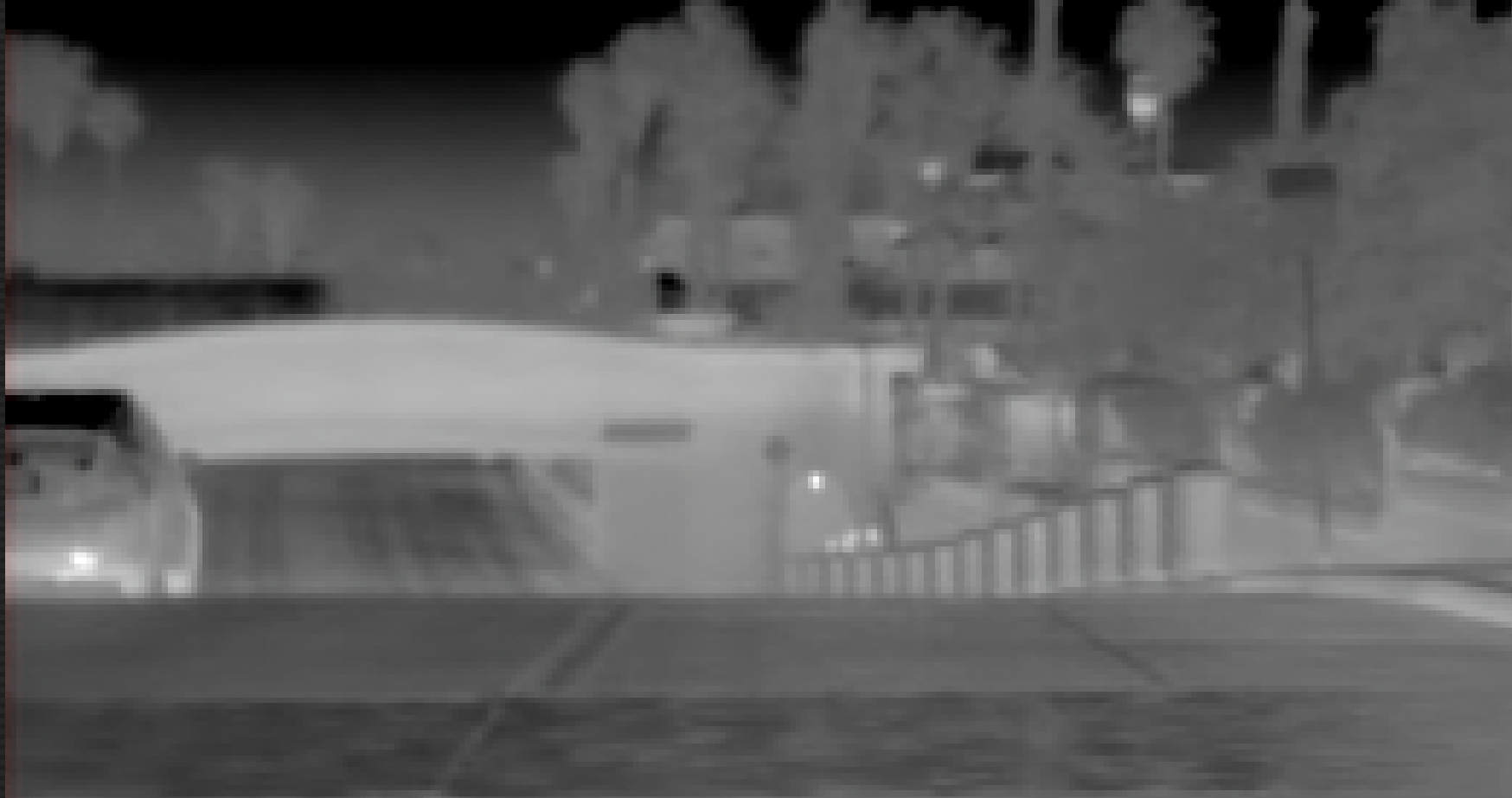}%
		\includegraphics[width=\imagesize]{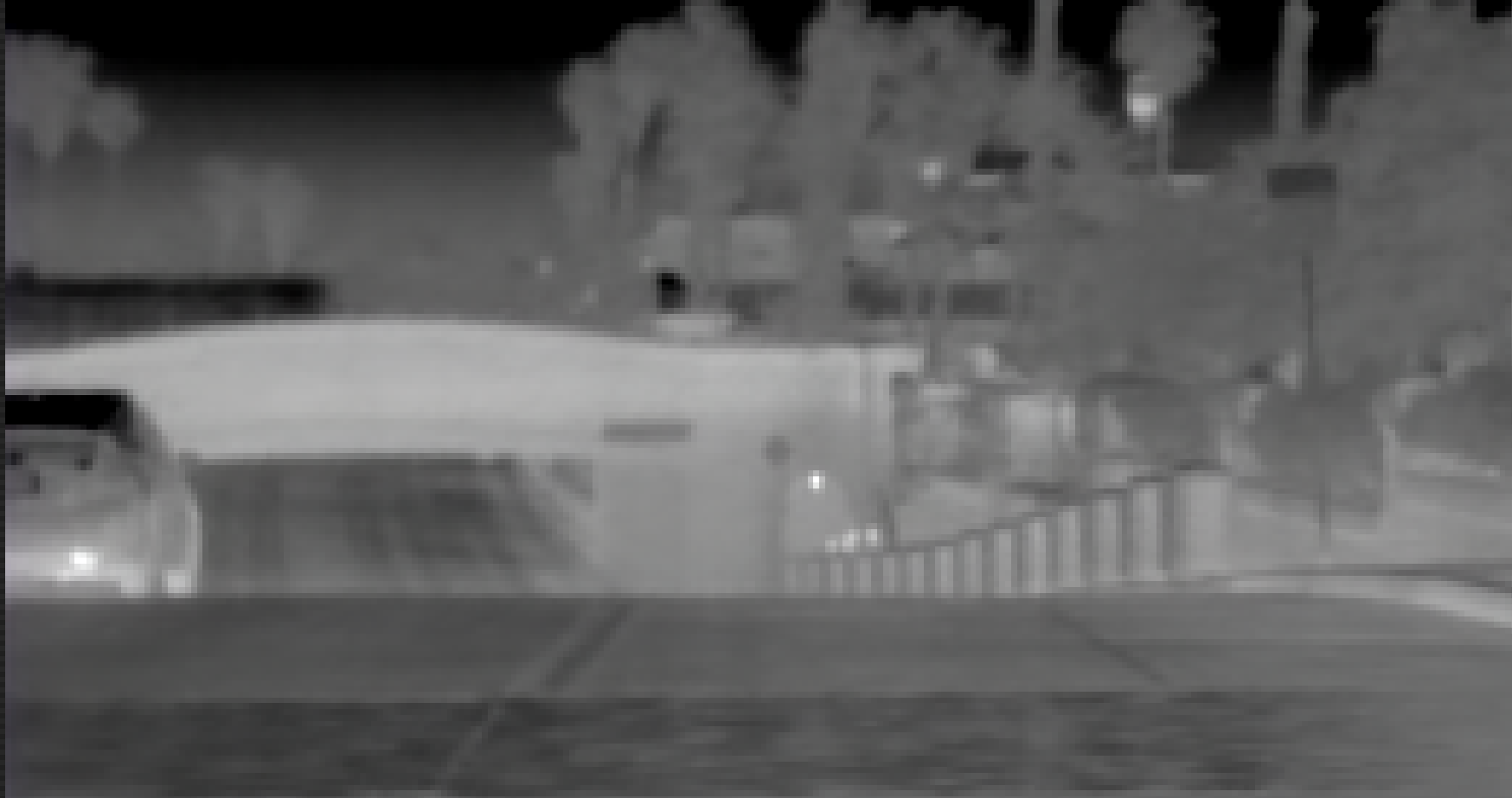}\\
		\vspace{0.07cm}
		\includegraphics[width=\imagesize]{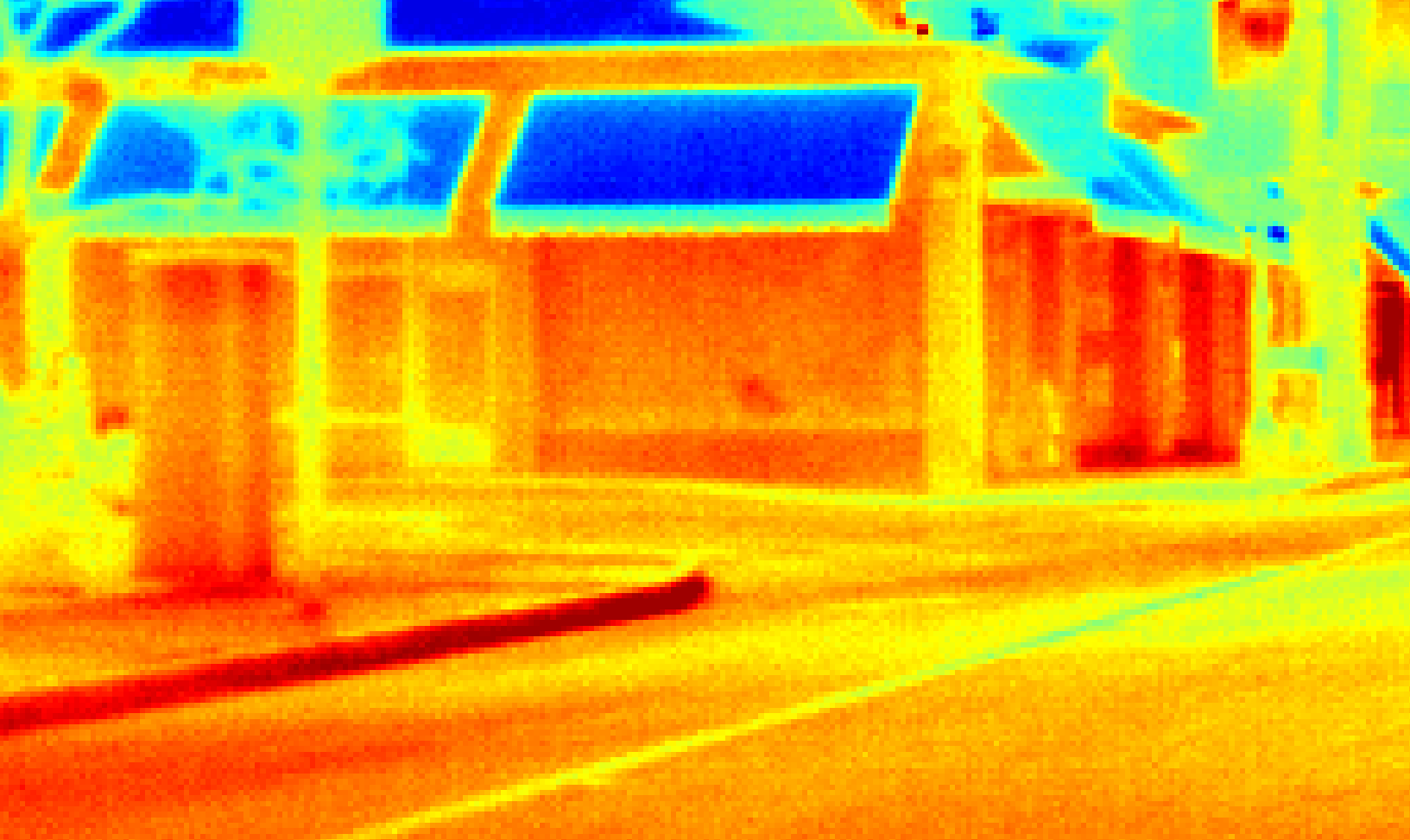}%
		\includegraphics[width=\imagesize]{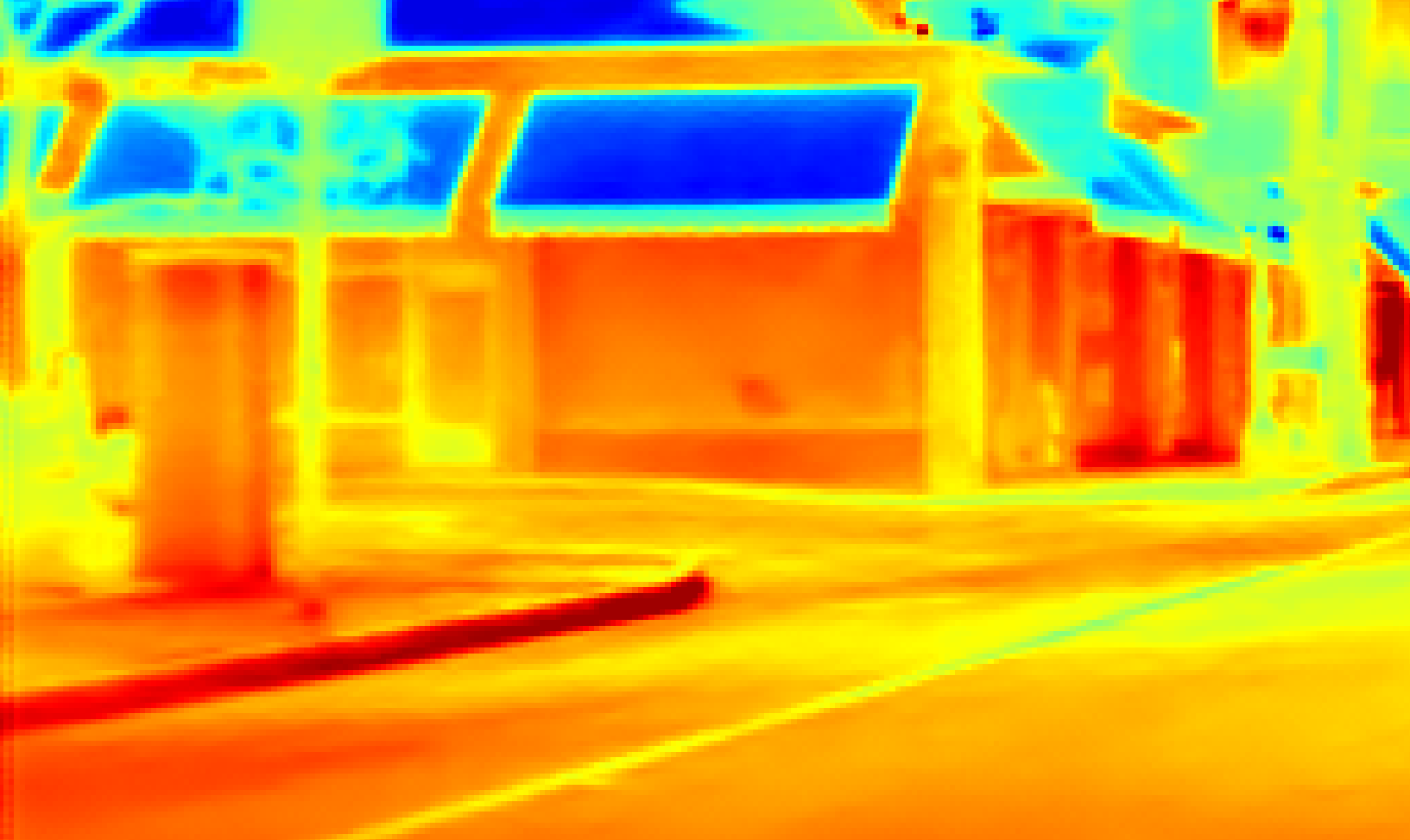}%
		\includegraphics[width=\imagesize]{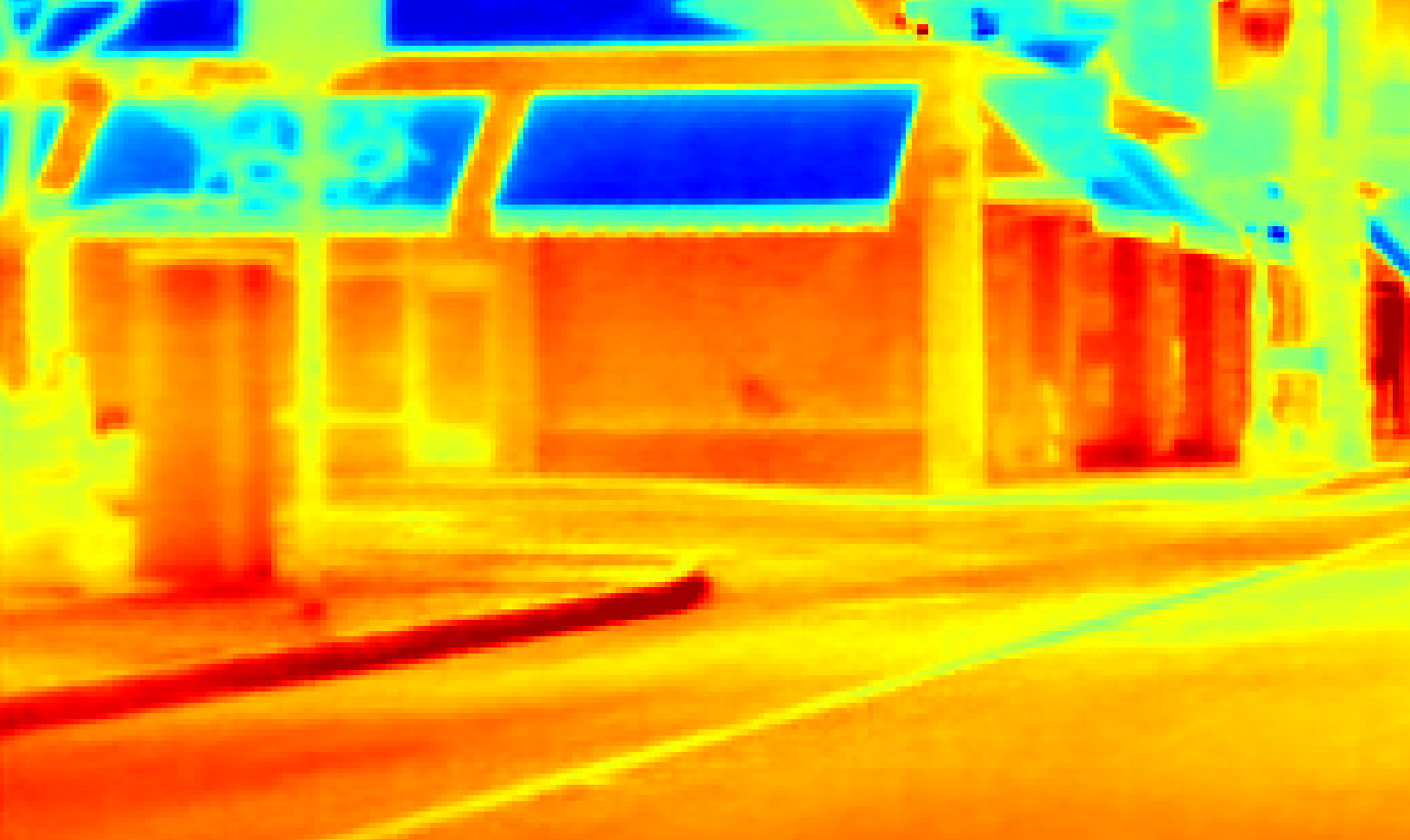}%
	\caption{Results on real noise from a thermal camera (FLIR ADAS dataset). From left to right: noisy, online MF2F and offline MF2F.}
	\label{fig:IR}
	\end{center}
\end{figure*}

Figure \ref{fig:IR} shows results on videos with real noise  from the FLIR ADAS  thermal infra-red dataset, both online and offline methods are compared. Results of the last row were displayed using a \textit{jet} color map.

\begin{figure*}
    \begin{center}
        \def\imagesize{0.2\linewidth}
        \subfloat{\includegraphics[clip,trim=0cm 0cm 0cm 0cm, width=\imagesize]{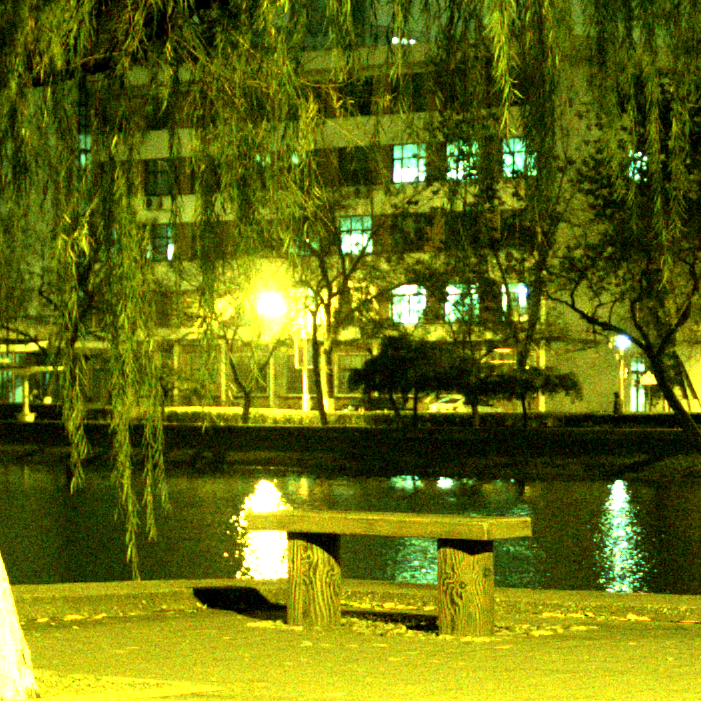}}
        \subfloat{\includegraphics[clip,trim=0cm 0cm 0cm 0cm, width=\imagesize]{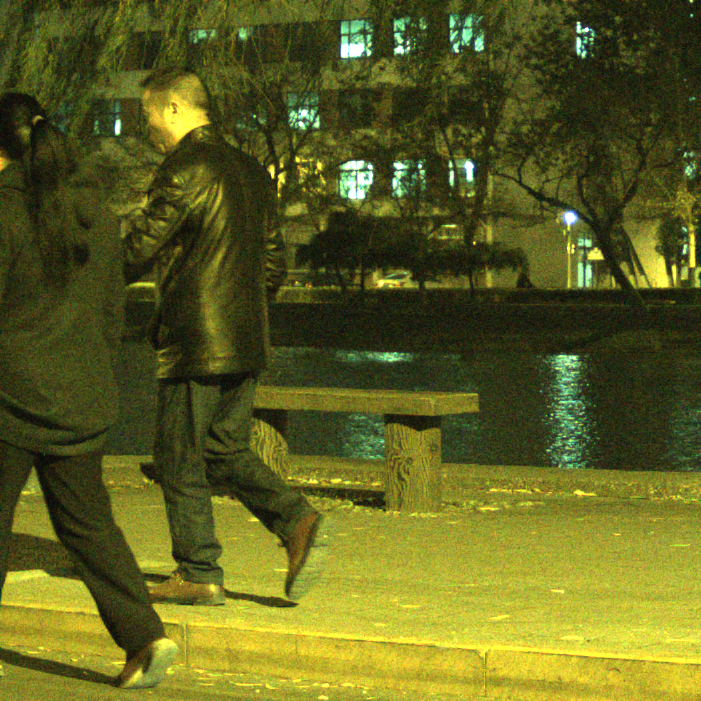}}
        \subfloat{\includegraphics[clip,trim=0cm 0cm 0cm 0cm, width=\imagesize]{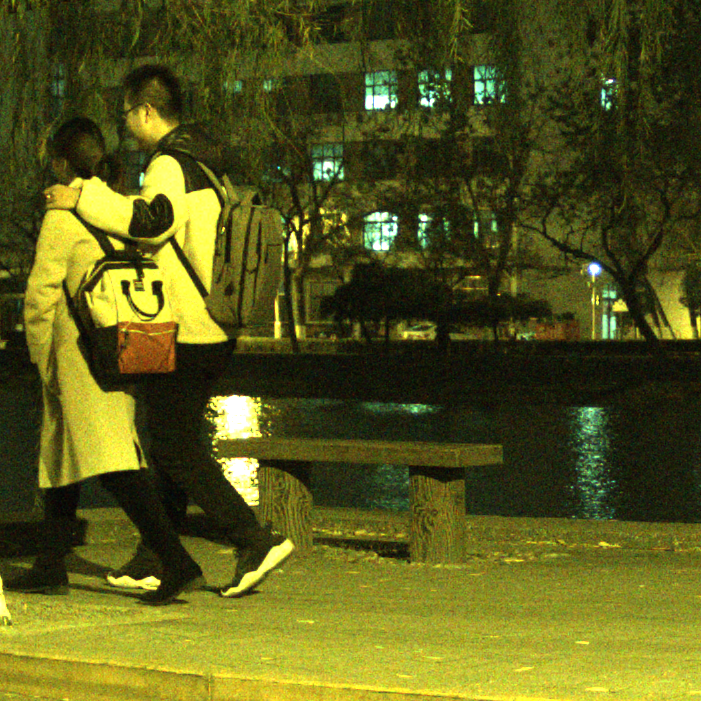}}
        \subfloat{\includegraphics[clip,trim=0cm 0cm 0cm 0cm, width=\imagesize]{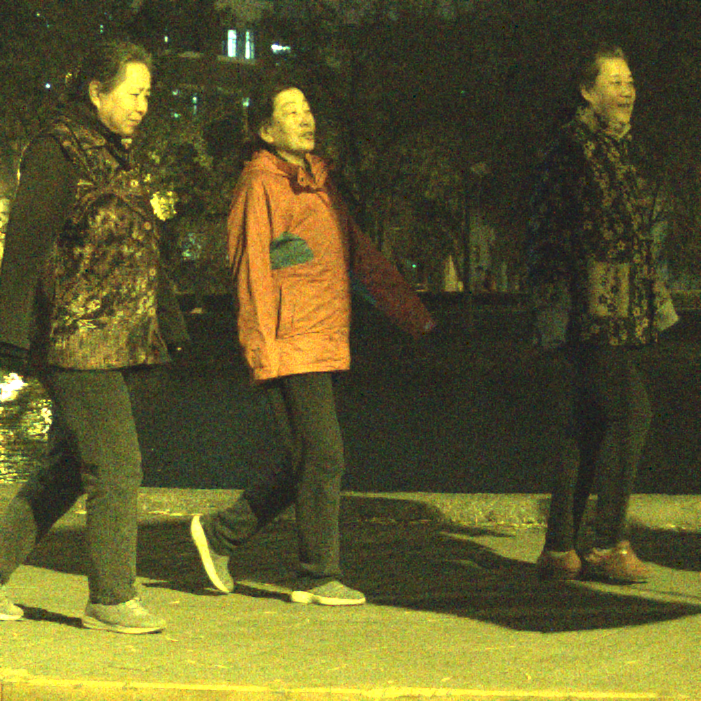}}
        \subfloat{\includegraphics[clip,trim=0cm 0cm 0cm 0cm, width=\imagesize]{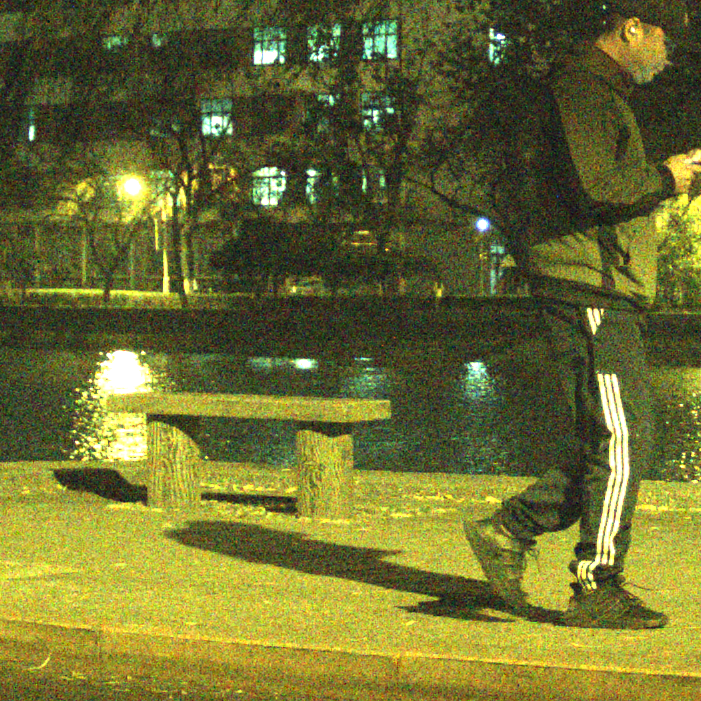}}
        
         \subfloat{\includegraphics[clip,trim=0cm 0cm 0cm 0cm, width=\imagesize]{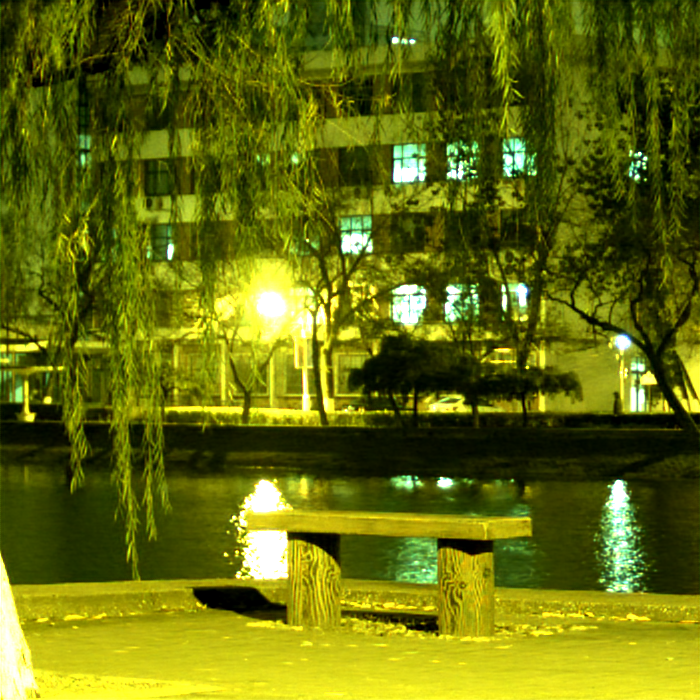}}
        \subfloat{\includegraphics[clip,trim=0cm 0cm 0cm 0cm, width=\imagesize]{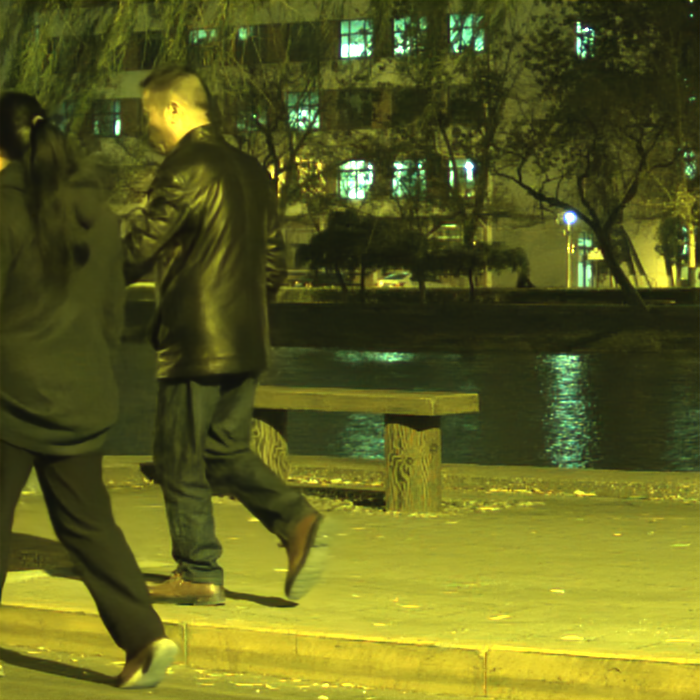}}
        \subfloat{\includegraphics[clip,trim=0cm 0cm 0cm 0cm, width=\imagesize]{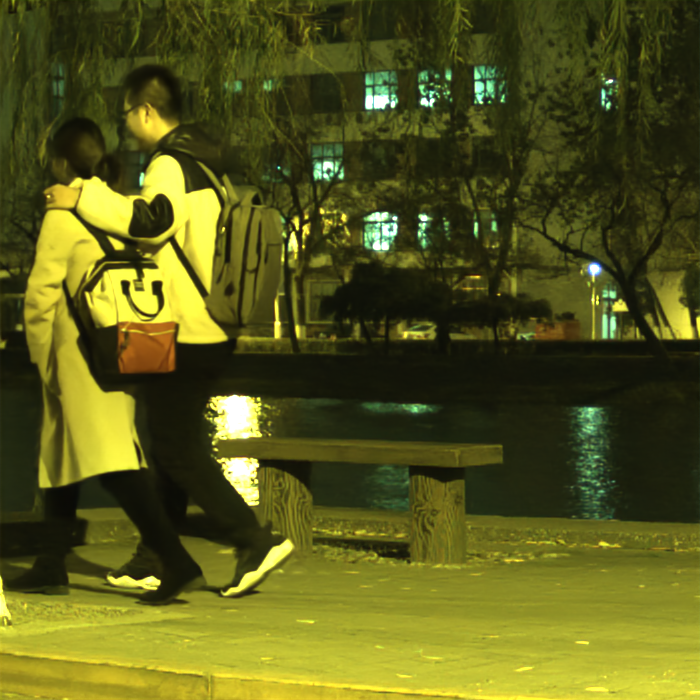}}
        \subfloat{\includegraphics[clip,trim=0cm 0cm 0cm 0cm, width=\imagesize]{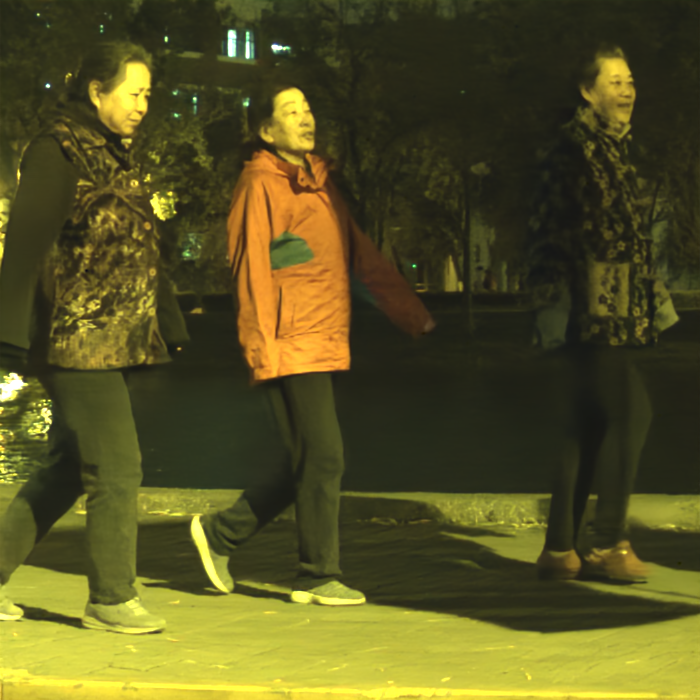}}
        \subfloat{\includegraphics[clip,trim=0cm 0cm 0cm 0cm, width=\imagesize]{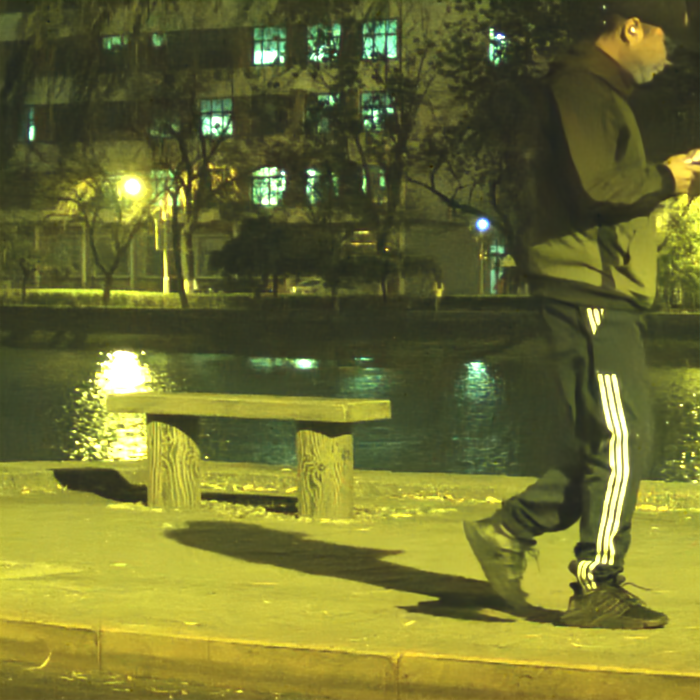}}
        
        \subfloat{\includegraphics[clip,trim=0cm 0cm 0cm 0cm, width=\imagesize]{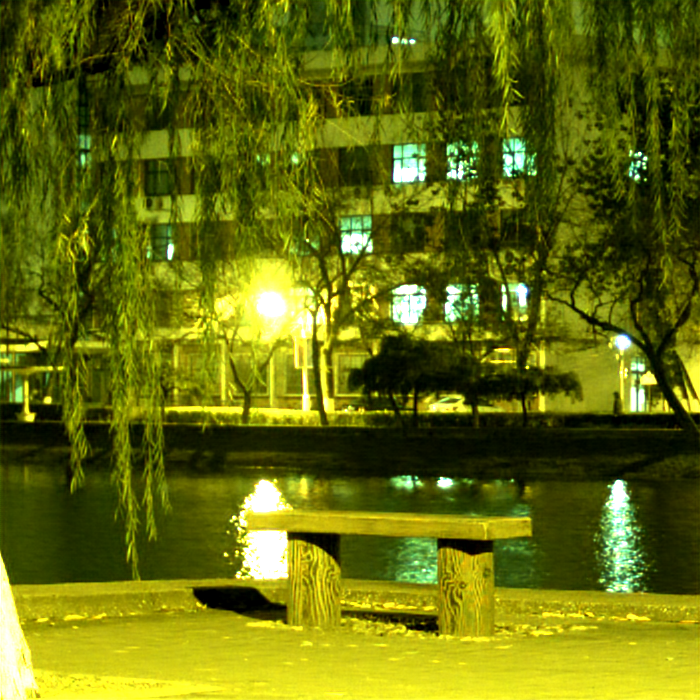}}
        \subfloat{\includegraphics[clip,trim=0cm 0cm 0cm 0cm, width=\imagesize]{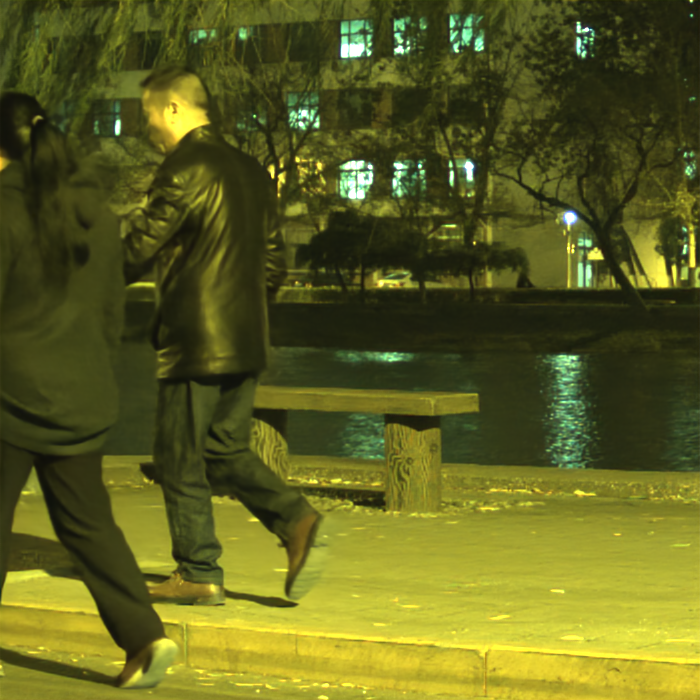}}
        \subfloat{\includegraphics[clip,trim=0cm 0cm 0cm 0cm, width=\imagesize]{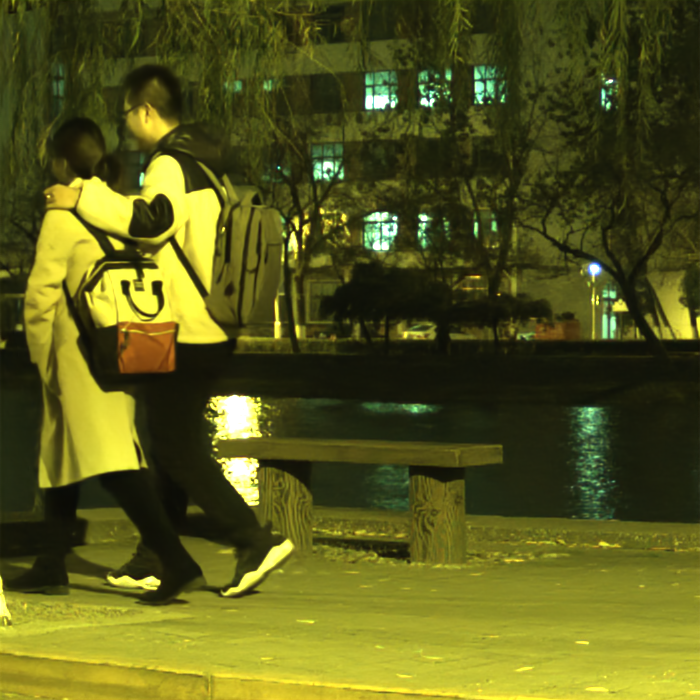}}
        \subfloat{\includegraphics[clip,trim=0cm 0cm 0cm 0cm, width=\imagesize]{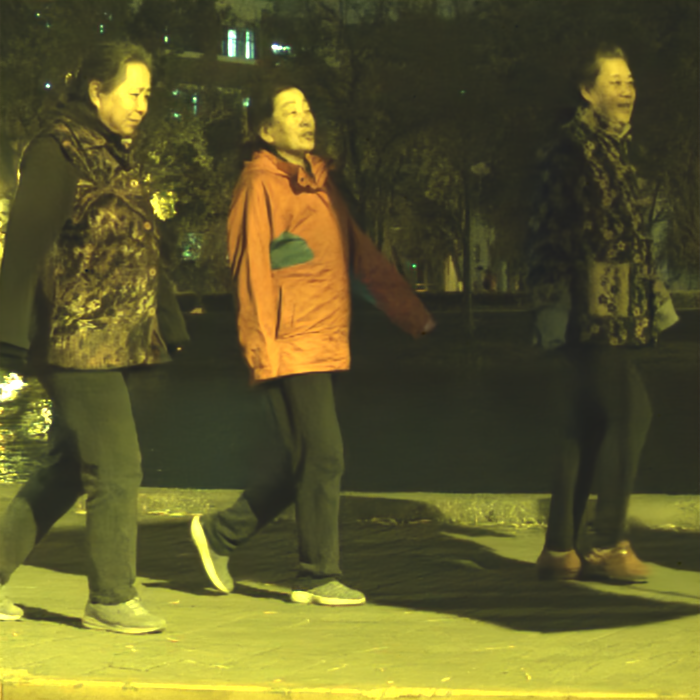}}
        \subfloat{\includegraphics[clip,trim=0cm 0cm 0cm 0cm, width=\imagesize]{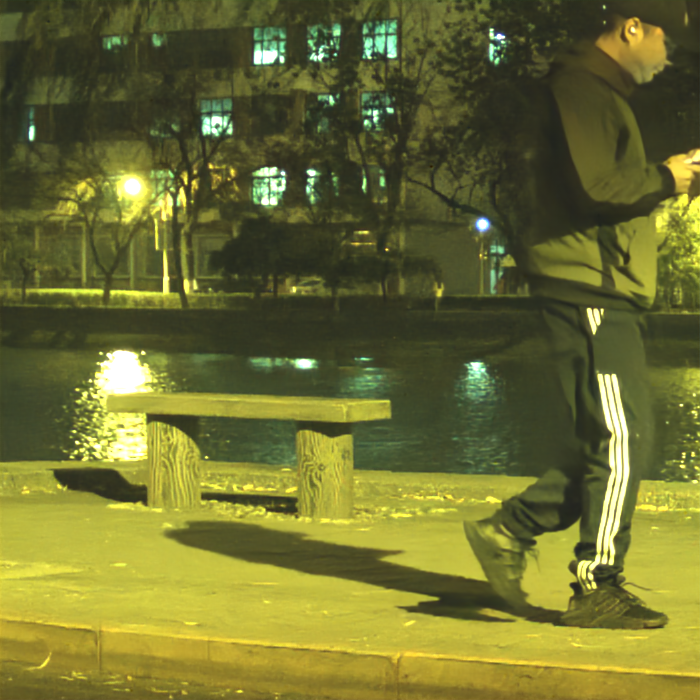}}
        
        \subfloat{\includegraphics[clip,trim=0cm 0cm 0cm 0cm, width=\imagesize]{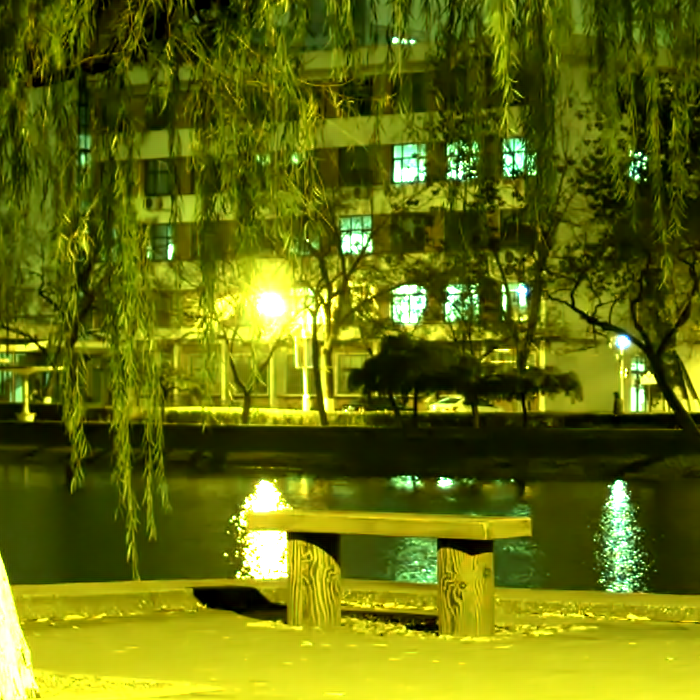}}
        \subfloat{\includegraphics[clip,trim=0cm 0cm 0cm 0cm, width=\imagesize]{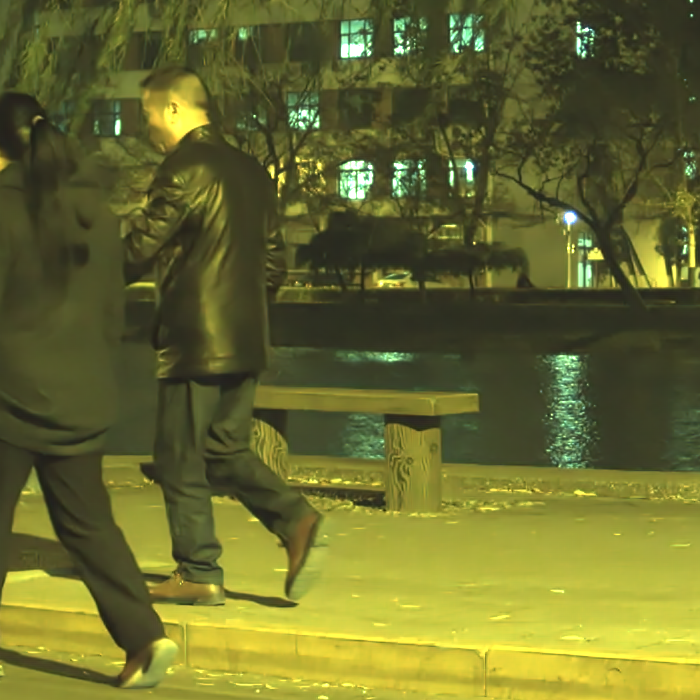}}
        \subfloat{\includegraphics[clip,trim=0cm 0cm 0cm 0cm, width=\imagesize]{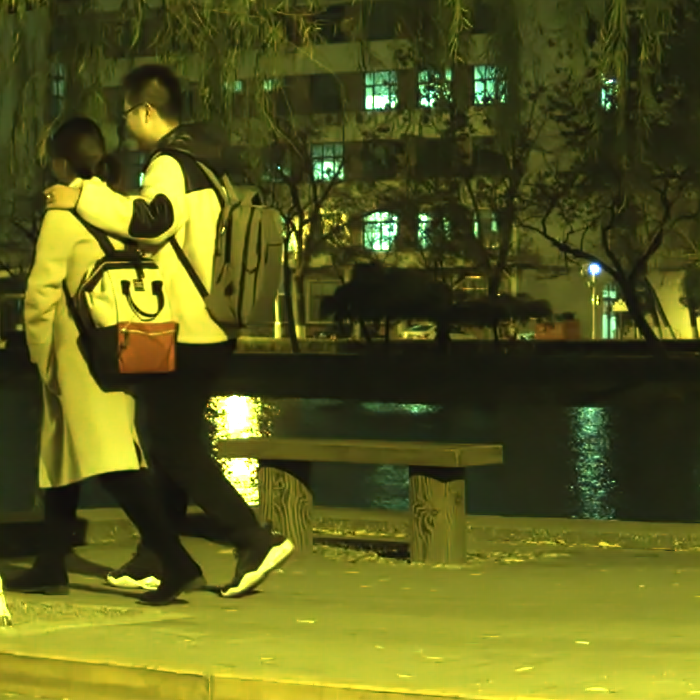}}
        \subfloat{\includegraphics[clip,trim=0cm 0cm 0cm 0cm, width=\imagesize]{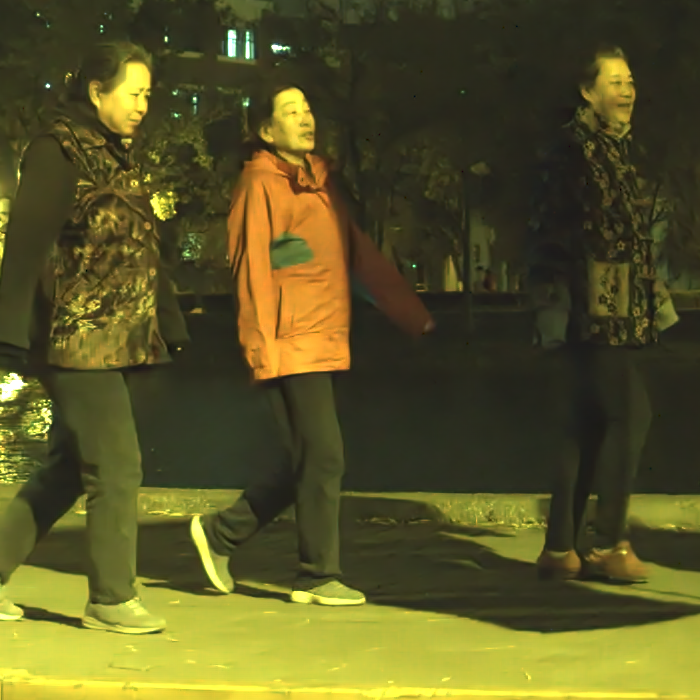}}
        \subfloat{\includegraphics[clip,trim=0cm 0cm 0cm 0cm, width=\imagesize]{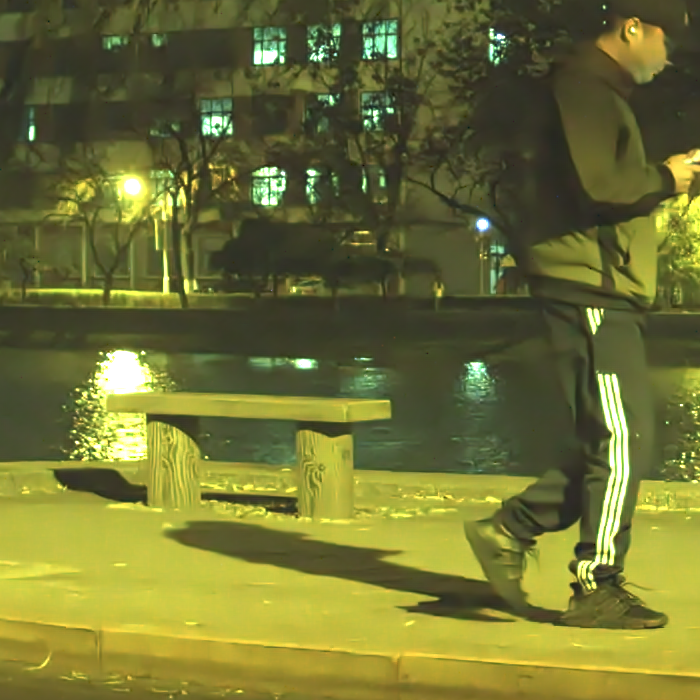}}

	   \setcounter{subfigure}{0}
        \subfloat[ISO1600]{\includegraphics[clip,trim=0cm 0cm 0cm 0cm, width=\imagesize]{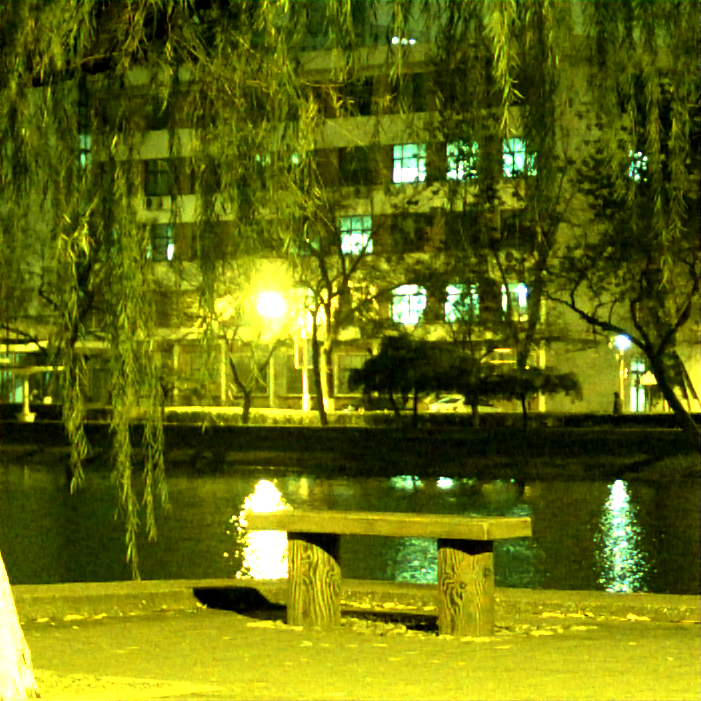}}
        \subfloat[ISO3200]{\includegraphics[clip,trim=0cm 0cm 0cm 0cm, width=\imagesize]{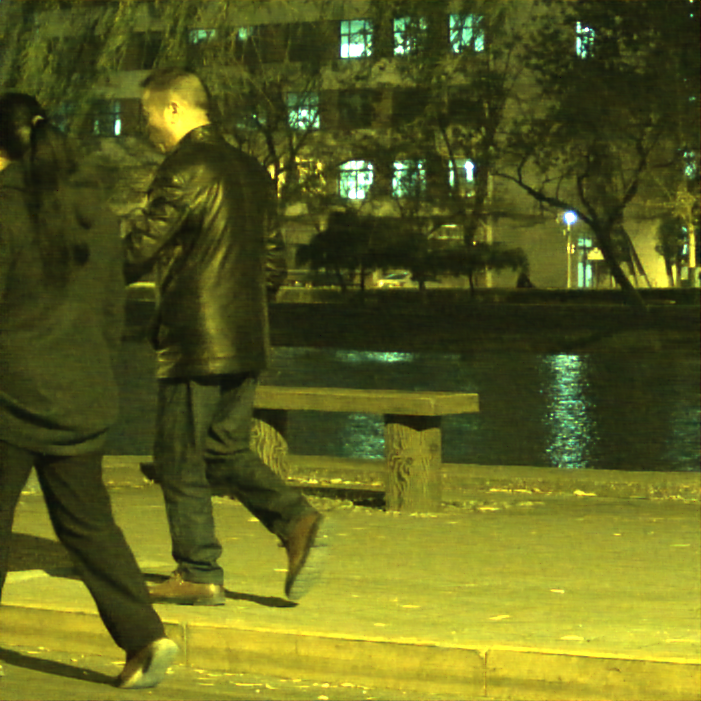}}
        \subfloat[ISO6400]{\includegraphics[clip,trim=0cm 0cm 0cm 0cm, width=\imagesize]{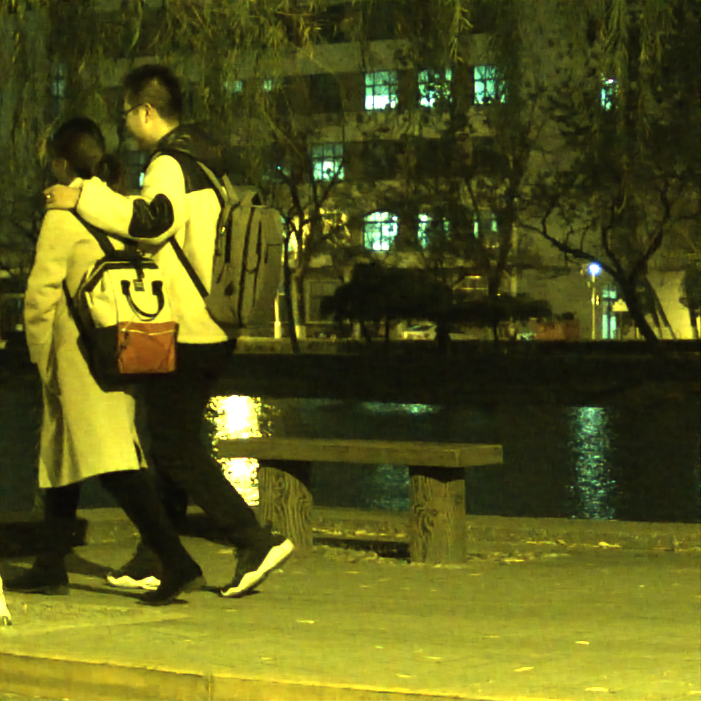}}
        \subfloat[ISO12800]{\includegraphics[clip,trim=0cm 0cm 0cm 0cm, width=\imagesize]{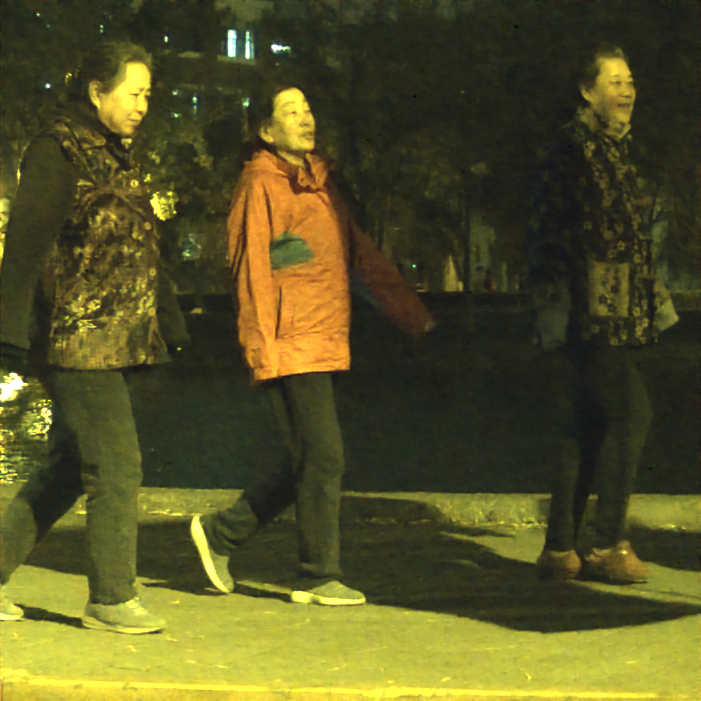}}
        \subfloat[ISO25600]{\includegraphics[clip,trim=0cm 0cm 0cm 0cm, width=\imagesize]{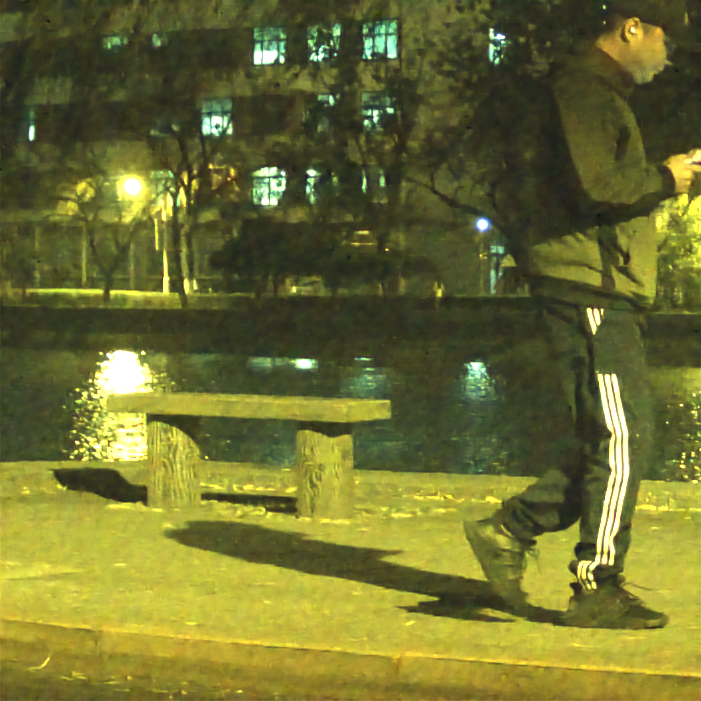}}

	    \caption{Real noise sequence: comparison of the same scene with different ISO levels. From top to bottom: noisy, online MF2F, offline MF2F, RViDeNet and online F2F.}
	    \label{fig:comparison_ISO}
	\end{center}
\end{figure*}

\begin{figure*}
    \begin{center}
        \setcounter{subfigure}{0}
        \subfloat[RViDeNet]{   \includegraphics[ width=0.49\textwidth]{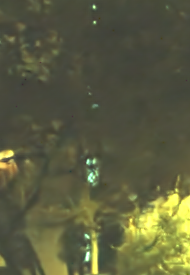}}
        \subfloat[Offline MF2F]{    \includegraphics[ width=0.49\textwidth]{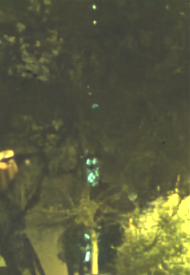}} 
        
        \setcounter{subfigure}{0}
        \subfloat[RViDeNet]{   \includegraphics[ width=0.24\textwidth]{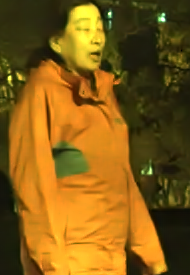}}
        \subfloat[Offline MF2F]{    \includegraphics[ width=0.24\textwidth]{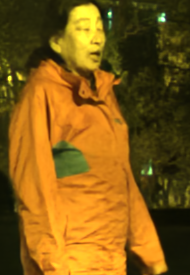}}
        \setcounter{subfigure}{0}
        \subfloat[RViDeNet]{   \includegraphics[ width=0.24\textwidth]{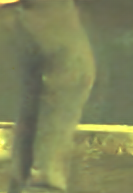}}
        \subfloat[Offline MF2F]{\includegraphics[ width=0.24\textwidth]{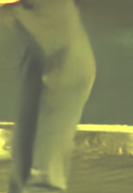}} 
    
	\caption{Comparison between RViDeNet and MF2F on real noisy images \protect\cite{kim2020-transfer-synth-to-real-noise}. The texture of trees, the coat, and the legs are poorly reconstructed by RViDeNet. On the contrary, MF2F produces  results  with more details. Furthermore, on the legs, we can see a ghosting effect on the result of  RViDenet.}
	\label{fig:comparison-rvidenet-MF2F}
	\end{center}
\end{figure*}

\begin{figure*}
	\begin{center}
		\def\imagesize{0.48\textwidth}
		\overimg[width=\imagesize]{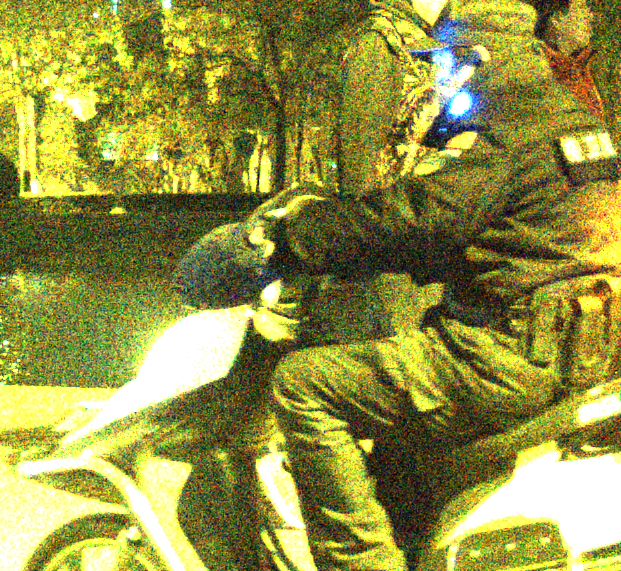}{noisy raw (demosaicked)}%
		\overimg[width=\imagesize]{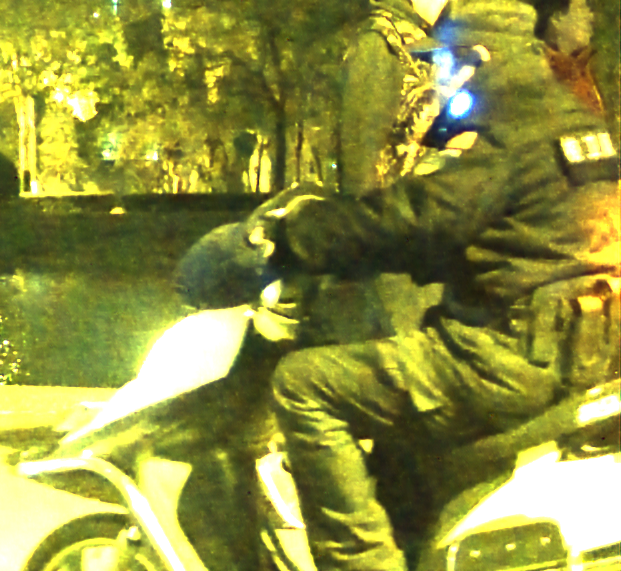}{online F2F}
	    \overimg[width=\imagesize]{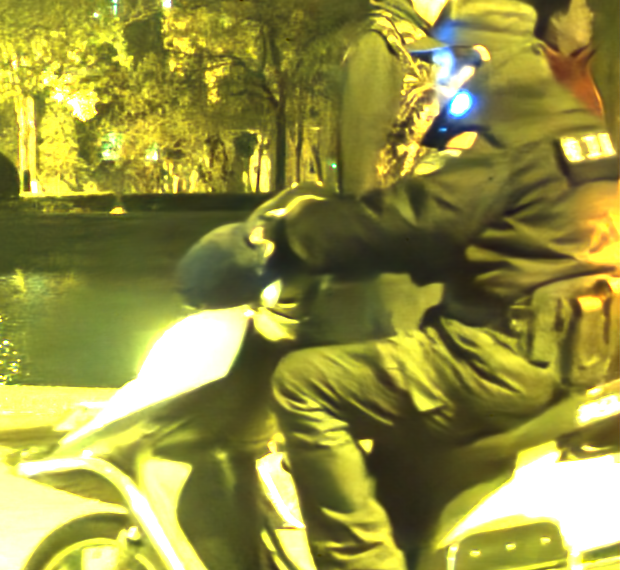}{offline MF2F}%
	    \overimg[width=\imagesize]{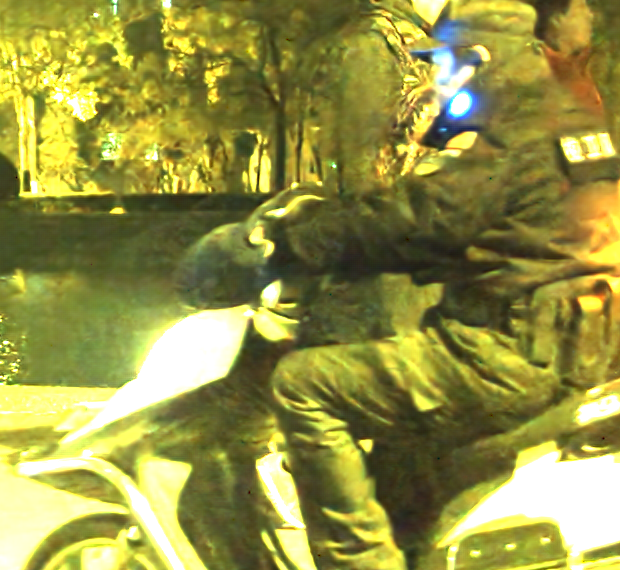}{RViDeNet}%
	\caption{A frame from a denoised raw video (ISO 12800) processed by F2F, offline MF2F, and RViDeNet. The contrast was changed for display purposes. All images are demosaicked and gamma corrected. The result of RViDeNet suffers from a strong ghosting effect and the people are poorly reconstructed. On the contrary the proposed MF2F gives better results.}
	\label{fig:ghosting_rvidenet}
	\end{center}
\end{figure*}

Figure \ref{fig:comparison_ISO} shows the results on real noise from the CRVD dataset. In this figure we compare the results on a same scene but with different ISO levels: 1600, 3200, 6400, 12800, 25600. The visual quality of denoising is not affected by the ISO level since the method quickly adapts to those different noise level. We display the results obtained both by the online MF2F and the offline MF2F. For comparison, we also added the results of online F2F and RViDeNet~\cite{yue2020supervised}. MF2F extends the performance of F2F and can adapt specifically to the noise of the video. As a results, it gives sharper results and with more details than RViDeNet. To illustrate that, more crops are shown in Fig. \ref{fig:comparison-rvidenet-MF2F}). RViDenet poorly reconstructs the texture of the trees, the sidewalk and even the folds in the coat. On the crops showing the legs, we see that RViDeNet has also \textit{ghosting effect} which is not present on the results of MF2F. An illustration of this ghosting effect is also shown in Fig. \ref{fig:ghosting_rvidenet} (see in front of the motorbike)

\section{Running time}

Table \ref{tab:running_times} reports the running time needed to process one color frame of  $800 \times 540$ pixels (including file IO) for all the proposed online methods. The times were measured on a multi-core server with a {\it NVIDIA RTX 2080 TI} GPU.
The online methods compute 20 Adam weight updates of the network (FastDVDnet) for each frame of the sequence. The offline method, on the other hand, performs a fixed number of Adam update steps regardless of the length of the video.
A comparison with the inference time of the FastDVDnet network is also provided.

Note that fine-tuning  the variance map or all the weights of the network requires roughly the same amount of time. This is because in both cases we need to back-propagate through the entire network and we perform the same number of weight update steps.

\begin{table}
\centering
\begin{tabular}{|l|c|}
  \hline
  \hspace{2.5cm} Method & time (in s) \\
  \hline
  MF2F (Online fine-tuning) & 6.78  \\
  MF2F fine-tuning the 8 levels variance map & 4.45 \\
  FastDVDnet (inference) & 0.56  \\
  \hline
  
\end{tabular}
\caption{Running time needed to process one color frame ($800 \times 540$) with the online algorithm. The fine-tunings are all on the FastDVDnet network. \label{tab:running_times}} 
\end{table}

\clearpage
\small{

}